\documentclass[a4paper,fleqn]{cas-dc}

\usepackage[numbers]{natbib}
\usepackage{amsmath,amssymb,amsfonts}
\usepackage{algorithmic}
\usepackage{algorithm}
\usepackage{booktabs}
\usepackage{multirow}
\usepackage{xcolor}
\usepackage{url}
\usepackage{xurl}
\usepackage{float}
\usepackage{placeins}
\usepackage{cuted}
\usepackage{capt-of}

\graphicspath{{../include/}}
\newcommand{\tablefontsize}{\scriptsize}
\begin{document}
\let\WriteBookmarks\relax
\def\floatpagepagefraction{1}
\def\textpagefraction{.001}

\shorttitle{TMT}
\shortauthors{Ruofan Wang et al.}

\title[mode=title]{TMT: Adapting CLIP to Infrared Imagery via Text-Derived Modality Knowledge Transfer}

\author[1]{Ruofan Wang}[orcid=0009-0007-4065-0832]
\ead{wangruofan@nudt.edu.cn}

\credit{Conceptualization, Methodology, Software, Formal analysis, Writing -- original draft}

\author[1]{Puzuo Wang}[orcid=0000-0002-0971-4013]
\ead{puzuo.wang@nudt.edu.cn}
\credit{Investigation, Validation, Data curation}

\author[1]{Xin Chen}[orcid=0009-0002-2596-7755]
\ead{cx342@nudt.edu.cn}
\credit{Investigation, Validation}

\author[1]{Zibo Yu}[orcid=0009-0006-3002-4507]
\ead{yzb_23@nudt.edu.cn}
\credit{Formal analysis, Visualization}

\author[1]{Jungang Yang}[orcid=0000-0002-3127-8705]
\ead{yangjungang@nudt.edu.cn}
\credit{Methodology, Writing -- review \& editing}

\author[1]{Yi Hou}[orcid=0009-0009-6720-1759]
\cormark[1]
\ead{yihou@nudt.edu.cn}
\credit{Supervision, Project administration, Writing -- review \& editing}

\affiliation[1]{organization={College of Electronic Science and Technology, National University of Defense Technology},
                city={Changsha},
                state={Hunan},
                country={China}}
\cortext[1]{Corresponding author}

\begin{abstract}
Vision-language models such as CLIP achieve strong performance across computer vision tasks, yet their image--text alignment is less reliable for infrared (IR) imagery because large-scale pre-training is dominated by visible-spectrum data.
Due to the high cost of the infrared--text labeling, we investigate adapting CLIP to IR through only paired infrared--visible images, which requires considering modality-specific knowledge while preserving scene content and CLIP's image--text alignment.
To achieve this, we propose \textbf{TMT} (Text-Derived Modality Knowledge Transfer), which adapts CLIP's image encoder through a text-defined infrared-to-visible direction and supervision from paired IR--visible images.
First, we found that modality discrepancy can be described by corresponding text embeddings, thus using the direction of text embedding differences to guide the knowledge transfer between visible and infrared modalities without erasing infrared-specific information.
To enhance the representation of semantics and details of infrared images, we introduce Generative--Perceptual Co-Supervision that combines semantic alignment from CLIP visible image encoder with diffusion noise-prediction feedbacks.
On four source-disjoint IR classification benchmarks, TMT achieves the highest mean zero-shot accuracy among the compared methods (48.92\%).
It also achieves the best IR--text retrieval results among the compared methods without infrared--text supervision during adaptation and the highest open-vocabulary segmentation mIoU on both CART and SemanticRT among the CLIP-like methods.
\end{abstract}

\begin{highlights}
\item Pseudo-visible features train an IR encoder for direct CLIP text matching.
\item Generic modality descriptions define an IR-to-visible transfer direction.
\item Paired semantic alignment and diffusion feedback jointly adapt the IR encoder.
\end{highlights}

\begin{keywords}
Infrared vision \sep Vision--language models \sep CLIP adaptation \sep Text-derived modality transfer \sep Infrared--visible image pairs \sep Diffusion noise-prediction supervision
\end{keywords}

\maketitle

\section{Introduction}

\par 
Infrared (IR) imaging captures emitted radiation rather than reflected visible light~\cite{mao2026pid} and supports perception when visible appearance is degraded, including multispectral object detection~\cite{chen2024amfd,xie2026madf}, aerial RGB--thermal perception~\cite{lee2024cart}, and multispectral semantic segmentation~\cite{ji2023semanticrt}. 
In the visible domain, vision--language pre-training has turned image--text alignment into a general-purpose interface for a wide range of visual tasks. 
Extending this interface to infrared imagery is particularly attractive and motivates the development of infrared representation models, including CLIP-style models that align images with natural language in the infrared domain. 
However, the native CLIP-style models meet challenges when applied to infrared imagery: they are pre-trained predominantly on visible-spectrum data, so their performance degrades on infrared data. 

\begin{figure}[!t]
\centering
\includegraphics[width=\columnwidth]{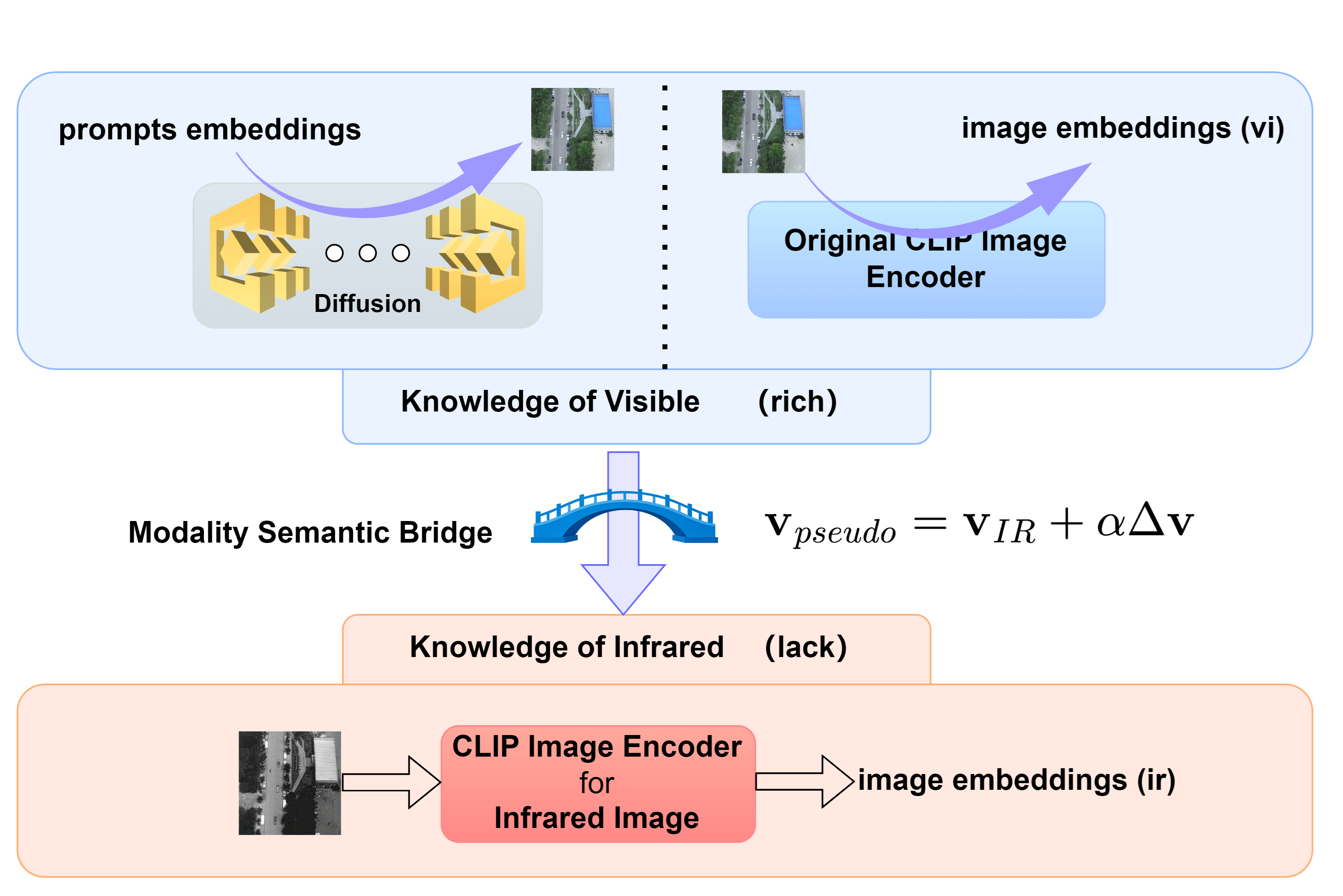}
\caption{Illustration of the core idea. In order to guide the knowledge transfer between visible (rich) and infrared (lack) modalities and increase the performance of infrared--text alignment,
a text-derived infrared-to-visible direction moves the infrared feature toward the visible-image region of CLIP's embedding space, and paired visible images provide a semantic reference, and a frozen diffusion model provides noise-prediction feedback during adaptation.}
\label{fig:framework_intro}
\end{figure}

\par To mitigate this degradation, existing methods connect infrared imagery to language in different ways, drawing on resources that are already available, such as paired infrared--visible images and pre-trained vision--language models in the visible domain.
A direct method fine-tunes image--text models with paired infrared--text supervision. However, manually annotating fine-grained infrared captions is expensive, which makes this supervision difficult to scale for large-scale infrared training.
Synthesizing such captions with large vision--language models offers one way to reduce this cost~\cite{qazi2026tclip}, while the thermal fidelity of the synthesized captions remains bounded by the generator's predominantly visible-spectrum knowledge.
A second route embeds infrared into a multimodal alignment framework. LanguageBind~\cite{zhu2023languagebind} aligns infrared features, together with features from multiple other modalities, toward the same text. ImageBind~\cite{girdhar2023imagebind} connects infrared features to a shared multimodal space through paired infrared--visible images. However, by forcing infrared features into alignment with other modalities---particularly other image modalities---these methods tend to lose modality-specific information of infrared, which in turn degrades performance on downstream tasks.

\par  
The limitations of the above routes arise from how they use visible-domain resources: one depends on infrared captions that are either costly to annotate or bounded in fidelity when synthesized, and the other forces infrared features into alignment with other modalities without preserving their modality-specific characteristics.
We instead consider moving the infrared feature itself toward the visible-image region of CLIP's embedding space, forming a pseudo-visible feature that existing visible-domain models can directly process. 
Figure~\ref{fig:framework_intro} illustrates the reasoning that leads to our core idea. The feature moving module serves as a bridge between the two modalities: it turns the forced alignment between infrared and visible features into the optimization of the pseudo-visible feature; the knowledge accumulated in visible-domain pre-trained models thus passes through this bridge into the infrared encoder, transferring visible-domain knowledge without infrared--text annotations and without erasing infrared-specific information. Then, the question is how to construct this bridge and what supervision can make the knowledge of visible-domain models pass through it.

\par To answer this question, we examine what the paired images do and do not provide: they preserve scene and object correspondence, but the spectral difference still needs to be expressed in CLIP's image--text embedding space. CLIP provides a shared space for image and text representations~\cite{radford2021clip}. Text-embedding differences have been used to steer CLIP-conditioned image generation~\cite{ramesh2022unclip} and to guide semantic augmentation across visual domains~\cite{vidit2023clipgap}. These precedents motivate a candidate infrared-to-visible direction constructed from generic descriptions of the two modalities; whether this direction helps infrared alignment is tested in our experiments. 

Once an infrared feature is transferred along this direction, CLIP's global image--text alignment can guide it toward the correct semantic content, but it does not directly specify which local visual details the feature should retain~\cite{wang2024diffusion,li2025un2clip}. A frozen CLIP image encoder can anchor the feature to its paired visible image, yet this semantic anchoring is not sensitive to local image detail. A complementary detail-sensitive signal is therefore needed, and a frozen diffusion model---which, following unCLIP, generates visible images conditioned on CLIP image embeddings~\cite{ramesh2022unclip}---provides it through noise-prediction feedback. 


\par We therefore propose \textbf{TMT} (\textbf{T}ext-Derived \textbf{M}odality Knowledge \textbf{T}ransfer), which adapts CLIP's image encoder to infrared inputs through a text-defined transfer direction and paired semantic and diffusion supervision.
The key insight is that infrared and visible images may differ substantially in radiometric appearance and local detail because they arise from different sensing processes~\cite{mao2026pid}, while paired observations still preserve object identity and coarse scene structure across spectra~\cite{jia2021llvip,m3fd}. This separation allows modality-specific changes to be represented and transferred without treating the two images as pixel-identical observations.

\par TMT first encodes a matched pair of generic visible and infrared descriptions with the frozen CLIP text encoder. It takes their visible-minus-infrared embedding difference, normalizes it, and adds the resulting direction to an infrared image feature to form a pseudo-visible feature. This construction is inspired by embedding arithmetic~\cite{mikolov2013efficient,mikolov2013distributed} and text-difference-based image manipulation~\cite{ramesh2022unclip}; it does not require every cross-spectral difference to be linear. The direction depends on the descriptions used to define it, so Section~5 compares it with a no-transfer variant and an image-derived direction. During training, the frozen visible-image CLIP encoder and the frozen diffusion model supervise only the pseudo-visible feature. Because the added direction remains fixed, satisfying their objectives requires updating the infrared image encoder that produced it: optimizing the pseudo-visible feature is thus the route by which visible-domain knowledge adapts the infrared encoder, and at inference the adapted encoder is used alone, without the shift or the supervisors. 


\par The main contributions of this paper are summarized as follows:
\begin{itemize}

    \item We introduce TMT, a novel CLIP adaptation framework that uses a pseudo-visible feature as a training-time interface for visible-domain supervision while retaining direct infrared--text matching at inference.
    \item Text-Derived Modality Transfer computes a modality direction from generic visible and infrared descriptions encoded by the frozen CLIP text encoder and applies it to infrared features through vector addition.
    \item Generative--Perceptual Co-Supervision combines noise-prediction feedback from a frozen diffusion model with paired semantic alignment to a frozen visible-image CLIP encoder; only the infrared image encoder is updated. 

    \item On four source-disjoint infrared classification benchmarks, TMT achieves 48.92\% mean zero-shot accuracy versus 43.52\% for an ImageBind-style CLIP control trained on the same IR500K image pairs. It also leads the compared methods without infrared--text supervision during adaptation in IR--text retrieval and achieves the highest mIoU among the compared CLIP-like methods on CART and SemanticRT.
\end{itemize}

\par The remainder of this paper is organized as follows. Section 2 reviews related work. Section 3 presents the proposed TMT framework in detail. Section 4 reports the experimental setup and results, followed by ablation studies and analysis in Section 5. Finally, Section 6 concludes the paper and discusses future work.

\section{Related Work}
\subsection{Vision--Language Alignment for Infrared Adaptation}

Vision--language pre-training establishes an embedding space for comparing images with text. CLIP~\cite{radford2021clip} learns this interface through image--text contrastive training, while ALIGN~\cite{jia2021align} scales alignment using noisy web supervision. Subsequent adaptation methods modify different parts of the interface: CoOp~\cite{zhou2022coop} learns continuous prompt contexts, CoCoOp~\cite{zhou2022cocoop} conditions them on the input image, and CLIP-Adapter~\cite{gao2023clipadapter} introduces residual feature adaptation. These approaches provide a basis for reusing pre-trained representations, but infrared adaptation also requires a source of supervision connecting the new sensing modality to language.

LanguageBind~\cite{zhu2023languagebind} establishes this connection directly by aligning infrared and other modalities with a frozen language encoder using paired text. ImageBind~\cite{girdhar2023imagebind} instead builds an image-centered shared space. Its thermal encoder learns from paired image--thermal data from LLVIP, gaining access to language alignment through the image modality. The distinction concerns the correspondence available during training: one route supplies infrared--text pairs, whereas the other uses images as an intermediate reference. ImageBind thus already demonstrates that explicit infrared--text pairing is not required for this form of cross-modal alignment.

IRGPT~\cite{cao2025irgpt} and Thermal-Det~\cite{ranasinghe2026thermaldet} further illustrate how supervision is organized around different infrared--language tasks. IRGPT uses the IR-TD corpus, comprising more than 260K infrared--text samples across pre-training, instruction, and benchmark subsets. Its bi-cross-modal curriculum and instruction tuning support generative question answering, including recognition and grounding. Thermal-Det combines synthetic thermal images with grounding text and captions for open-vocabulary detection, and distills geometric and semantic cues from a frozen RGB teacher using paired images. The comparisons therefore involve both the availability of language-linked infrared data and the form of the prediction. Their generative-answer and detection interfaces differ from the image--text similarity interface retained by an adapted CLIP encoder.

A concurrent line of work reduces the annotation cost itself rather than avoiding infrared text. T-CLIP~\cite{qazi2026tclip} conditions a large vision--language model on the paired visible image to synthesize physics-aware thermal captions at two granularities, and adapts both CLIP encoders with two independently trained LoRA branches that are ensembled at inference for cross-modal retrieval. This route supplies instance-level infrared--text supervision through generation. TMT instead uses text only to define a shared modality direction and adapts the image encoder alone, retaining the standard single-encoder matching interface at inference and serving classification, retrieval, and segmentation.

TMT shares ImageBind's use of infrared--visible pairs without instance-level infrared--text annotations. It introduces generic modality descriptions to specify a shared infrared-to-visible change, while paired visible images provide the content reference for each training instance. This separates the language-defined modality change from the image-defined scene correspondence. The paired observations then supervise a transferred representation through both semantic alignment and generative feedback, extending the role of image pairing in encoder adaptation. Caption synthesis and this caption-free use of paired images are complementary answers to the scarcity of infrared--text supervision.

\subsection{Text-Guided Semantic Transfer}

Semantic relationships in word embeddings can appear as approximately consistent vector offsets~\cite{mikolov2013efficient,mikolov2013distributed}. This observation motivates representing a specified semantic change by a direction in feature space. CLIP~\cite{radford2021clip} supplies a joint image--text space in which such directions can be investigated across modalities. Alignment alone does not guarantee that arbitrary text differences produce the intended visual change; the descriptions, representation, and transformation used remain relevant to the outcome.

In unCLIP, Ramesh et al.~\cite{ramesh2022unclip} use a normalized difference between original and target text descriptions to manipulate an image embedding through spherical interpolation. A decoder then maps the modified embedding to an image, demonstrating how a text-specified change can guide visual synthesis. CLIPGap~\cite{vidit2023clipgap} uses textual domain concepts for single-domain-generalization object detection. Its semantic augmentation operates on detector-backbone features and is combined with a text-based classification loss. Both approaches use language to specify a desired visual variation, but they differ in the representation being modified and the objective used to evaluate that modification. Together, these examples suggest that text guidance must be considered alongside the feature space and downstream objective. Successful visual editing or detector augmentation motivates further study of such directions, but provides no direct evidence of their effectiveness for infrared encoder adaptation.

TMT uses text-guided transformation to construct a training interface for infrared encoder adaptation. Generic descriptions of visible and infrared imaging define a fixed, shared direction that maps infrared features to pseudo-visible representations. These representations receive supervision from visible-domain models, with gradients passing through the transfer operation to update the infrared encoder. Whereas the preceding applications target image editing or detector augmentation, this operation supports learning an infrared representation for image--text matching; the text-derived shift is removed at inference.

\subsection{Generative Supervision for Visual Representation Learning}

Generative models connect visual representations to the image information needed for synthesis. DDPM~\cite{ho2020ddpm} learns through denoising, while latent diffusion models~\cite{rombach2022ldm} perform diffusion in an autoencoded image space. unCLIP~\cite{ramesh2022unclip} conditions generation on CLIP image embeddings. This conditioning interface allows a generative objective to evaluate information carried by a visual representation, providing a route for transferring knowledge from a pre-trained generator to an encoder.

DIVA~\cite{wang2024diffusion} uses feedback from a frozen diffusion model to optimize CLIP representations with images alone, targeting fine-grained visual distinctions while retaining zero-shot capabilities. un$^2$CLIP~\cite{li2025un2clip} improves CLIP's visual-detail sensitivity by inverting the unCLIP generator. In both cases, generation provides an additional criterion for learning visual features beyond their similarity to text. Their connection to TMT lies in using a pre-trained generative model as a source of encoder supervision, with the learned representation subsequently serving perception tasks. DIVA also illustrates that visual knowledge acquired during generative pre-training can supply an image-dependent adaptation signal without a newly annotated caption for each training image.

VLV~\cite{zhang2025vlv} and De-Diffusion~\cite{wei2024dediffusion} explore related feedback through language representations. VLV distills diffusion knowledge through continuous embeddings in a vision--language--vision auto-encoding path. De-Diffusion trains an image-to-text encoder with a frozen text-to-image diffusion decoder, encouraging the textual representation to retain visual information. These approaches broaden the role of generative supervision, while differing in whether the information is retained in an image embedding, a continuous language representation, or generated text.

The above studies primarily examine visible-image representations. Infrared adaptation additionally requires generative feedback to remain tied to the scene observed in another sensing modality. TMT couples two constraints on the same pseudo-visible representation: a frozen CLIP image encoder supplies the paired visible-image embedding as a semantic reference, while a frozen diffusion model supplies noise-prediction feedback that encourages retention of visual details useful for generation. Their joint use connects generative representation learning to paired cross-modal content alignment, allowing visible-domain supervision to train an infrared encoder for subsequent image--text matching. 

\section{Method}

This section describes the framework for adapting CLIP to infrared imagery.
We first formalize how infrared inputs affect CLIP's zero-shot image--text matching.
The framework then comprises two parts: Text-Derived Modality Transfer and Generative--Perceptual Co-Supervision.

\subsection{Problem Formulation}

Given a pre-trained CLIP model with an image encoder $f_{\text{img}}$ and a text encoder $f_{\text{txt}}$, the task is to adapt the image encoder for infrared--text alignment. For an infrared image $\mathbf{I}_{\text{IR}}$, its visual representation $\mathbf{v}_{\text{IR}} = f_{\text{img}}(\mathbf{I}_{\text{IR}})$ should be close to the text prototype that describes its content. For a class-specific prompt, this prototype is $\mathbf{t}_c = f_{\text{txt}}(\langle\text{prompt}_c\rangle)$, and their alignment can be evaluated through the following normalized similarity distribution:
\begin{equation}
    p(y = c \mid \mathbf{I}_{\text{IR}}) = \frac{\exp(\text{sim}(\mathbf{v}_{\text{IR}}, \mathbf{t}_c) / \tau)}{\sum_{c'=1}^{C} \exp(\text{sim}(\mathbf{v}_{\text{IR}}, \mathbf{t}_{c'}) / \tau)}
\end{equation}
where $\text{sim}(\cdot, \cdot)$ denotes cosine similarity and $\tau$ is the temperature parameter. This distribution is the inference interface of CLIP rather than the training problem itself: the central objective is to learn an adapted image encoder whose infrared features remain content-consistent with their visible counterparts and are better aligned with text prototypes.

For infrared inputs, the original CLIP image encoder may produce features that are poorly positioned relative to the text-aligned geometry learned from visible images. TMT therefore learns an adapted encoder $f_{\text{img}}^{\text{TMT}}$ such that, for an infrared--visible image pair $(\mathbf{I}_{\text{IR}},\mathbf{I}_{\text{vis}})$, the representation $f_{\text{img}}^{\text{TMT}}(\mathbf{I}_{\text{IR}})$ preserves the content correspondence with $\mathbf{I}_{\text{vis}}$ while moving toward a text-compatible visible-image representation. The adaptation uses paired infrared--visible images and generic modality descriptions, but does not require paired infrared image--text annotations. The resulting encoder is evaluated through zero-shot classification, infrared--text retrieval, and open-vocabulary semantic segmentation.

\subsection{Framework Overview}

\begin{figure*}[t]
\centering
\includegraphics[width=\textwidth]{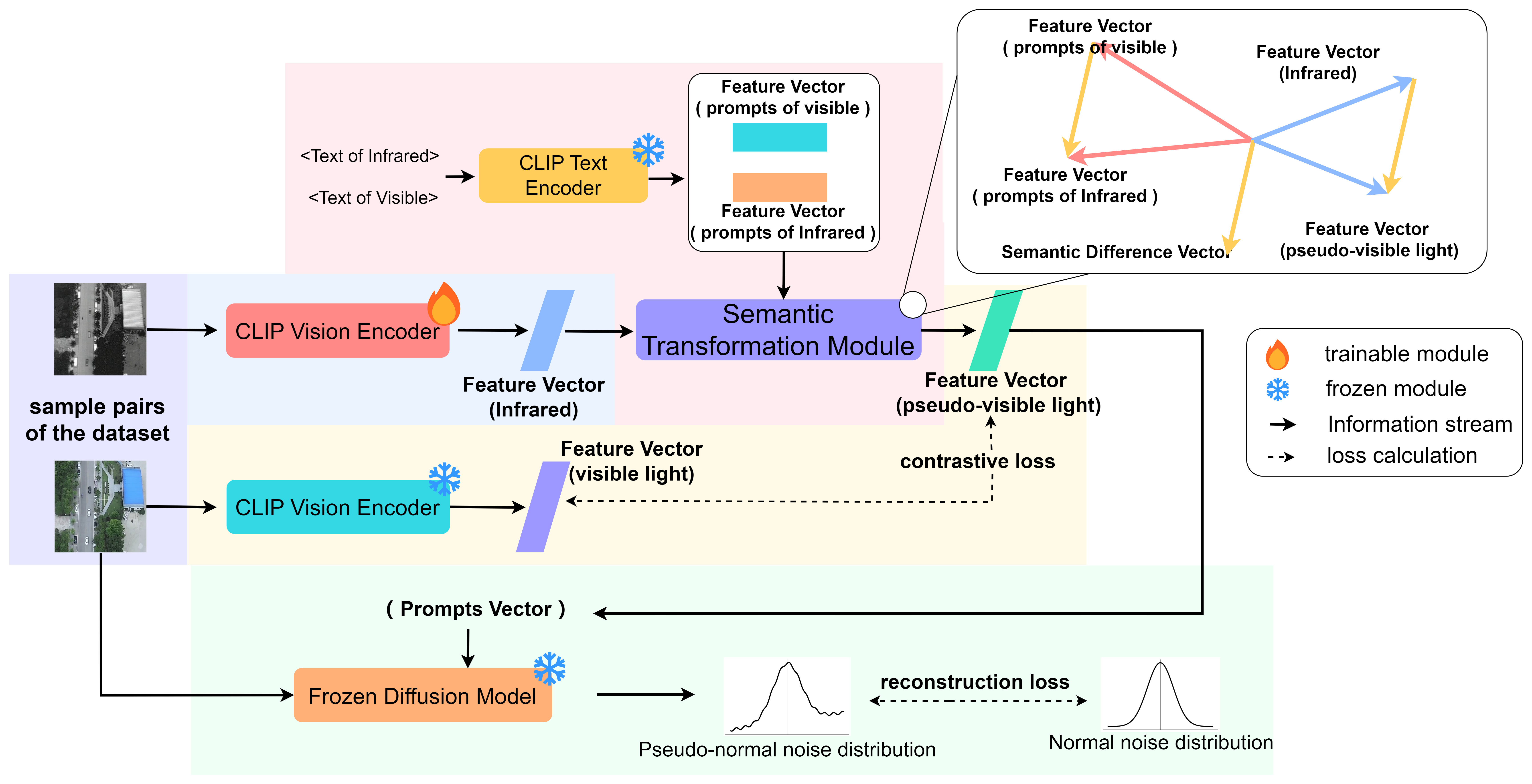}
\caption{Overview of TMT. A paired infrared--visible sample (purple zone) is processed by a trainable CLIP image encoder for the infrared image and a frozen CLIP image encoder for the visible counterpart. Generic visible--infrared modality descriptions are encoded by the frozen CLIP text encoder; their normalized difference produces the modality direction $\Delta \mathbf{v}$, which is applied by Text-Derived Modality Transfer to the infrared feature and yields a pseudo-visible representation (red zone). Generative--Perceptual Co-Supervision (yellow and green zones) constrains this representation through two complementary paths: the frozen CLIP image encoder provides paired semantic alignment with the visible feature, while the frozen diffusion model uses the visible image and the pseudo-visible representation to compute a latent-space noise-prediction loss against a sampled standard-Gaussian target. Only the trainable infrared image encoder (blue zone) is updated during training. (best viewed in color)}
\label{fig:framework}
\end{figure*}

Figure~\ref{fig:framework} summarizes the data flow of TMT. 
Given a paired infrared--visible sample $(\mathbf{I}_{\text{IR}},\mathbf{I}_{\text{vis}})$, the trainable CLIP image encoder first extracts the infrared feature, which Text-Derived Modality Transfer then shifts along a text-derived modality direction to obtain a pseudo-visible representation. 
To construct this direction, a frozen CLIP text encoder embeds generic visible--infrared modality descriptions, whose normalized visible-minus-infrared difference provides $\Delta\mathbf{v}$; the module then adds $\alpha\Delta\mathbf{v}$ to the infrared feature to form this representation $\mathbf{v}_{\text{pseudo vis}}$.
Generative--Perceptual Co-Supervision then constrains $\mathbf{v}_{\text{pseudo vis}}$ through two complementary paths. First, a frozen CLIP image encoder extracts the feature of the paired visible image and constrains $\mathbf{v}_{\text{pseudo vis}}$ through paired semantic alignment with it. 
Second, the frozen diffusion model supplies latent-space noise-prediction feedback conditioned on $\mathbf{v}_{\text{pseudo vis}}$, with the paired visible-image latent as the diffusion input. 
Their weighted sum updates only the trainable CLIP image encoder; the text encoder, both frozen supervision models, and $\Delta\mathbf{v}$ remain fixed. 
At inference time, the adapted image encoder alone produces infrared features for the standard CLIP text-matching pipeline, without the text-derived shift or diffusion decoding.

\subsection{Text-Derived Modality Transfer}


Text-Derived Modality Transfer serves as a bridge between infrared representations and pre-trained visible-image models by constructing pseudo-visible features through which visible-image supervision can adapt the infrared encoder. Prior work has shown that text-embedding differences can steer CLIP-conditioned image generation~\cite{ramesh2022unclip} and that text-described domain concepts can guide visual feature augmentation across visual domains~\cite{vidit2023clipgap}. 
For infrared adaptation, the modality gap arises mainly from the different imaging mechanisms of the two spectra, so the differences that the transfer must express are largely shared across the images of each modality rather than tied to individual scenes. We therefore derive the transfer direction from language descriptions of visible and infrared imaging properties. This direction guides infrared features toward the visible-image semantic region; the shifted feature, denoted by $\mathbf{v}_{\text{pseudo vis}}$, is termed a pseudo-visible feature. Paired visible images supply the instance-specific reference that constrains content consistency during the transfer. 


To get the text-derived modality direction, 
we use a fixed pair of generic visible and infrared descriptions, denoted by $(\mathbf{d}^{\text{vis}},\mathbf{d}^{\text{IR}})$, generated with Kimi-K2.5~\cite{kimi2026k25} and then manually selected and refined for the infrared-to-visible transfer. One description specifies the source modality and the other target modality, allowing the desired change to be expressed through their semantic difference. This construction separates the description of the imaging modality from the identity of an individual training sample: the text supplies a shared reference, while the image encoder supplies the sample-specific representation. The descriptions therefore require no captions for individual infrared images and are separate from the class-specific prompts used for downstream evaluation.

The two descriptions are designed around modality-level properties, which are shared by the images of a modality rather than tied to any particular scene. Table~\ref{tab:modality-properties} summarizes the properties that connect the imaging mechanism to the appearance change we seek to express. The two modalities differ in the radiation that forms the image---reflected visible light versus emitted thermal radiation~\cite{mao2026pid}---and consequently in their contrast, color, texture, and noise characteristics, while object shape and spatial arrangement are preserved across spectra. The descriptions verbalize exactly these properties: the modality-dependent rows distinguish the source description from the target description, and the shared structural properties appear in both descriptions as content to be preserved. Because these properties characterize each modality as a whole rather than any individual sample, a single fixed description pair is sufficient to specify the transfer for all training data.

\begin{table*}[!t]
\centering
\tablefontsize
\caption{Modality-level properties underlying the description pair. The source and target descriptions verbalize the modality-dependent rows; the shared structural properties are stated in both descriptions as content to be preserved.}
\label{tab:modality-properties}
\begin{tabular*}{\textwidth}{@{\extracolsep{\fill}}lll@{}}
\toprule
Property & Visible & Infrared \\
\midrule
Radiation source & Reflected scene illumination & Emitted thermal radiation (temperature, emissivity) \\
Contrast and color & Albedo- and shading-based; rich color & Thermal contrast; weak or absent color \\
Texture and noise & Strong high-frequency detail; near-uniform noise & Smoother texture; fixed-pattern noise \\
Shared structure & \multicolumn{2}{c}{Object shape and spatial arrangement (preserved)} \\
\bottomrule
\end{tabular*}
\end{table*}

We encode the two descriptions with the frozen CLIP text encoder:
\begin{equation}
    \mathbf{e}^{\text{vis}} = f_{\text{txt}}(\mathbf{d}^{\text{vis}}),
    \qquad
    \mathbf{e}^{\text{IR}} = f_{\text{txt}}(\mathbf{d}^{\text{IR}}).
\end{equation}
Using the same encoder places the descriptions in a common embedding space, while freezing it preserves the text representation learned during CLIP pre-training. The paired construction provides two semantic reference points for the transformation. Their difference encodes the specified source-to-target change, and the subtraction order determines its orientation. We subtract the infrared embedding from the visible embedding and normalize the difference:
\begin{equation}
    \Delta\mathbf{v} = \frac{\mathbf{e}^{\text{vis}}-\mathbf{e}^{\text{IR}}}
    {\|\mathbf{e}^{\text{vis}}-\mathbf{e}^{\text{IR}}\|_2}.
\end{equation}
Because the pair verbalizes modality-level properties (Table~\ref{tab:modality-properties}) rather than sample-specific content, the resulting direction can be shared by every training sample and every experiment, and its usefulness depends on the two descriptions and their encoding rather than on any per-sample quantity. L2 normalization then separates the orientation of this difference from its magnitude. Consequently, the transfer coefficient controls the length of the added vector without also inheriting the norm of the text difference. 
The resulting direction is a language-defined reference for adaptation: a fixed, content-free offset whose role is to move infrared features into the visible domain. It is not intended to remove or rewrite sample semantics---the direction contains no content description by construction, and the additive form leaves the encoder's output untouched beneath the shift.

For an infrared image $\mathbf{I}_{\text{IR}}$, the trainable image encoder produces $\mathbf{v}_{\text{IR}}=f_{\text{img}}(\mathbf{I}_{\text{IR}})$. We apply the shared modality direction through an additive transformation:
\begin{equation}
    \mathbf{v}_{\text{pseudo vis}}=\mathbf{v}_{\text{IR}}+\alpha\cdot\Delta\mathbf{v},
    \label{eq:semantic_transfer}
\end{equation}
where $\alpha$ controls the transfer strength. The same offset is applied to every infrared feature, so sample-specific information continues to enter through $f_{\text{img}}(\mathbf{I}_{\text{IR}})$. We compute $\Delta\mathbf{v}$ once and keep both it and $\alpha$ fixed during training. This provides a constant modality reference while the encoder learns from the paired observations. 
Since the offset $\alpha\Delta\mathbf{v}$ is constant with respect to the encoder parameters, the derivative of $\mathbf{v}_{\text{pseudo vis}}$ with respect to $\mathbf{v}_{\text{IR}}$ is the identity, allowing gradients from the supervision losses to pass directly to the encoder. 


The resulting $\mathbf{v}_{\text{pseudo vis}}$ serves as the shared input to the Generative--Perceptual Co-Supervision module. Using the paired visible image as a reference, the perceptual branch constrains semantic correspondence, while the generative branch provides feedback on visual detail, as described in the following subsection.

\subsection{Generative--Perceptual Co-Supervision}

To align infrared images with text prototypes more effectively and transfer visible-light knowledge to the infrared domain, we require the pseudo-visible embedding produced by TMT to satisfy two related but distinct requirements. It should preserve the content correspondence between a paired infrared image and its visible counterpart, while also retaining visual details that are weakly represented in the infrared input, such as texture and edges. These requirements are not equivalent in the CLIP representation. Image--text contrastive pre-training gives CLIP strong global semantic alignment, but prior work has reported that this training paradigm can underrepresent fine-grained visual differences~\cite{wang2024diffusion,li2025un2clip}. A supervision signal based only on semantic proximity could therefore encourage content consistency without sufficiently constraining the detail information required by the visible-image branch. Conversely, a signal based only on visual generation could favor a plausible visible appearance without ensuring that the generated content remains the same as the paired infrared observation. This motivates two complementary constraints rather than an unsupported claim that either objective is universally insufficient on its own. 

To implement these complementary constraints, TMT uses a frozen CLIP image encoder to provide the paired visible-image embedding as a semantic anchor, together with a frozen diffusion generator for visual-detail supervision. Following unCLIP~\cite{ramesh2022unclip}, this generator synthesizes visible images conditioned on CLIP image embeddings through diffusion noise prediction. This learned relationship between the conditioning embedding and image appearance provides the basis for its detail feedback. In TMT, the pseudo-visible embedding conditions the generator's noise prediction on the paired visible-image latent, allowing the resulting loss to encourage the infrared encoder to retain visual information useful to the generator.

The CLIP branch constrains $\mathbf{v}_{\text{pseudo vis}}$ to remain semantically aligned with the paired visible-image embedding, preserving object and scene correspondence. The diffusion branch uses noise-prediction feedback to encourage the same representation to retain visual details useful for conditional image generation. Together, the two branches couple semantic alignment with generative detail supervision: the paired semantic reference constrains what content is represented, while the generative objective constrains how that representation supports the visible-image noise-prediction process.

\paragraph{Diffusion-based Noise-Prediction Supervision.}
We use the frozen diffusion model to impose this generative constraint. Let $\mathbf{x}_{\text{vis}}$ denote the clean latent corresponding to the visible image $\mathbf{I}_{\text{vis}}$. In the standard DDPM objective~\cite{ho2020ddpm}, the image latent is first perturbed with a timestep-dependent Gaussian noise and the diffusion model predicts that same noise. For TMT, we do not add forward diffusion noise to the clean visible latent: the latent is supplied directly to the frozen diffusion model, $\mathbf{v}_{\text{pseudo vis}}$ is used as the condition, and an independently sampled Gaussian vector remains the prediction target. This choice separates the latent used as the visual reference from the random target used for noise prediction. Appendix~\ref{app:noise-free} provides the corresponding gradient derivation and the empirical comparison of the no-noising and DDPM-style strategies, including their training stability. For a paired sample $(\mathbf{I}_{\text{IR}},\mathbf{I}_{\text{vis}})$, with $\boldsymbol{\epsilon} \sim \mathcal{N}(\mathbf{0},\mathbf{I})$, we define:
\begin{equation}
    \mathcal{L}_{\text{diff}} =
    \mathbb{E}_{(\mathbf{I}_{\text{IR}},\mathbf{I}_{\text{vis}}), \boldsymbol{\epsilon}, t}
    \left[
    \left\|
    \boldsymbol{\epsilon}
    -
    \epsilon_\theta
    \left(\mathbf{x}_{\text{vis}}, t, \mathbf{v}_{\text{pseudo vis}}\right)
    \right\|_2^2
    \right]
    \label{eq:diff_loss}
\end{equation}
where $\epsilon_\theta(\cdot)$ denotes the frozen diffusion model and $t$ is a sampled timestep. $\mathcal{L}_{\text{diff}}$ compares predicted and sampled noise in the diffusion latent space; it does not compare a generated RGB image with the paired visible image pixel by pixel. Gradients through the frozen diffusion model update the trainable infrared image encoder so that $\mathbf{v}_{\text{pseudo vis}}$ carries information useful to the visible-image generator.

\paragraph{Paired Semantic Alignment Supervision.} 
The generative constraint alone does not identify which plausible visible content belongs to the observed infrared instance. We therefore retain the paired visible image as a semantic anchor. The frozen CLIP image encoder $\phi_{\text{CLIP}}^{\text{vis}}$ maps $\mathbf{I}_{\text{vis}}$ to an embedding in the CLIP joint embedding space:
\begin{equation}
    \mathbf{v}_{\text{vis}} = \phi_{\text{CLIP}}^{\text{vis}}(\mathbf{I}_{\text{vis}}).
\end{equation}
We then align the pseudo-visible embedding with this paired embedding using the cosine distance:
\begin{equation}
    \mathcal{L}_{\text{semantic}} =
    1 -
    \frac{\mathbf{v}_{\text{pseudo vis}} \cdot \mathbf{v}_{\text{vis}}}
    {\|\mathbf{v}_{\text{pseudo vis}}\|_2 \cdot \|\mathbf{v}_{\text{vis}}\|_2}
    \label{eq:semantic_loss}
\end{equation}
This is a paired semantic alignment loss rather than a contrastive objective: it uses the visible counterpart as a positive reference but does not introduce negative samples or class labels. Minimizing it preserves the content correspondence of the infrared--visible pair and limits semantic drift, while the frozen teacher prevents the limited paired training data from redefining the visible CLIP embedding space. Together with the diffusion-based constraint, it gives the trainable image encoder a target that is both content-consistent and sensitive to visible-image detail.

The two supervision signals are combined as:
\begin{equation}
    \mathcal{L}_{\text{total}} = \lambda_{\text{diff}} \cdot \mathcal{L}_{\text{diff}} + \lambda_{\text{semantic}} \cdot \mathcal{L}_{\text{semantic}}
    \label{eq:total_loss}
\end{equation}
where $\lambda_{\text{diff}}$ and $\lambda_{\text{semantic}}$ weight the diffusion-based noise-prediction supervision and the paired semantic alignment supervision, respectively. The first term encourages $\mathbf{v}_{\text{pseudo vis}}$ to retain information that supports visible-image detail, whereas the second limits changes that would break the content correspondence with $\mathbf{I}_{\text{vis}}$. Because $\Delta\mathbf{v}$ and both supervisory networks are fixed, the dependence of $\mathbf{v}_{\text{pseudo vis}}$ on the trainable parameters is inherited from $\mathbf{v}_{\text{IR}}$. Gradients from both terms can therefore pass through the transfer operation and update the same image encoder. The objective does not train a separate transfer module to be retained at test time; it uses that module to expose a visible-oriented training signal to the encoder that will ultimately be deployed.

\subsection{Training and Inference}
\textbf{Training.}
Only the trainable CLIP image encoder $f_{\text{img}}$ is optimized through $\mathcal{L}_{\text{total}}$. The transfer coefficient $\alpha$ and the text-derived direction $\Delta\mathbf{v}$ remain fixed, as do the text encoder, the visible-image CLIP teacher, and the diffusion model. These frozen components still participate in the forward computation and transmit gradients through their outputs, but their parameters do not absorb the infrared-domain correction. Keeping the reference models fixed also prevents the supervision targets from moving with the adapted encoder, making the optimization target stable throughout training.

\textbf{Inference.}
Given an infrared image without a visible counterpart, the TMT-adapted image encoder directly produces its image embedding. This embedding is compared with the text embeddings of target-class prompts through the standard CLIP similarity interface. The generic modality descriptions used to construct the training-time direction are distinct from these dataset-specific evaluation prompts. The text-derived shift, the visible-image teacher, and the diffusion model are not invoked at inference. The deployed system consequently has the same input and matching interface as CLIP, while its image encoder has been adapted using the two complementary training signals described above.

\section{Experiments}

We evaluate TMT from the perspective of the tasks for which an adapted CLIP encoder is intended to be used. The experiments cover zero-shot classification, infrared--text retrieval, and open-vocabulary semantic segmentation. These tasks test progressively different properties of the adapted representation: category transfer, image--language alignment, and spatially resolved prediction. Section~5 then examines the components and mechanisms responsible for the observed behavior.

\subsection{Experimental Setup}

\paragraph{Training Data.}
TMT is trained on approximately 70K aligned infrared--visible image pairs from IR500K~\cite{ran2025diffv2ir}. The adaptation uses only the paired images. In particular, no paired infrared image--text annotation, evaluation class label, or evaluation caption is used to train TMT. These conditions define the supervision setting studied in this work. This setting is practically motivated: infrared--text corpora require dedicated caption collection or synthesis at scale, as illustrated by the construction of IR-TD~\cite{cao2025irgpt} and by VLM-based caption synthesis for thermal benchmarks~\cite{qazi2026tclip}, whereas paired infrared--visible images are directly available from dual-sensor benchmarks and require no per-image language annotation~\cite{jia2021llvip,m3fd,lee2024cart}.

\paragraph{Evaluation Datasets.}
For infrared--text retrieval, we use IR-TD~\cite{cao2025irgpt}. For this study, we randomly divide it into 80\% training and 20\% reserved data; from the reserved portion, we randomly sample 1{,}000 images with their annotations as the test set and use the remainder as a validation set. All reported retrieval results are evaluated on this test set. The training split is used only to fine-tune the CLIP-FT reference, whereas TMT, OpenAI CLIP, un$^2$CLIP, and DIVA do not use IR-TD training pairs.
We use Kimi-K2.5 to rewrite the source descriptions into concise infrared-oriented captions, and evaluate every method with the same test captions. LanguageBind is evaluated as an externally trained infrared--language reference. For zero-shot classification, we evaluate eight infrared benchmarks: M3FD~\cite{m3fd}, SMOD~\cite{chen2024amfd},
FLIRv1~\cite{flirv1}, FLIRv2~\cite{flirv2}, LLVIP~\cite{jia2021llvip}, JU-VNT-T~\cite{ghosh2021juvnt}, MFNet~\cite{ha2017mfnet}, and PST900~\cite{shivakumar2020pst900}. The primary cross-source group comprises SMOD, JU-VNT-T, MFNet, and PST900, which are absent from the IR500K adaptation subset. M3FD, FLIRv1, FLIRv2, and LLVIP share source-dataset families with IR500K and are reported as supplementary benchmarks. Evaluation images are excluded from TMT adaptation, including for the source-overlapping benchmarks. Detection-style datasets are converted into classification benchmarks by cropping the annotated infrared bounding boxes and using the crop labels as class labels. For MFNet and PST900, labeled infrared regions are used to form classification examples from the available annotations. TMT does not use the evaluation labels or class-specific text during adaptation.

For dense prediction, we evaluate open-vocabulary semantic segmentation on CART~\cite{lee2024cart} and SemanticRT~\cite{ji2023semanticrt}. Neither dataset is included in the IR500K training subset, so these experiments test transfer across datasets rather than direct reuse of segmentation annotations. The four source-disjoint classification benchmarks provide the primary zero-shot evaluation, while the four source-overlapping benchmarks provide supplementary results.
The original images and, where applicable, captions and annotations in the training and evaluation datasets listed above were released by their respective dataset creators and should be obtained from those original sources.

\paragraph{Prompts and Metrics.}
For classification and segmentation, Kimi-K2.5 is used to generate three distinct groups of class prompts for each dataset. These prompts pair dataset-specific class names with infrared appearance cues, such as thermal contrast and silhouette, that a generic visible-photo template would leave unspecified. They are evaluation inputs only: the generic visible--infrared descriptions used to construct TMT's training-time modality direction do not contain the evaluation class names. Every method is evaluated with the same prompt groups, and we report each group's result and the average to show sensitivity to wording. This design reduces dependence on a single prompt set, but it does not establish that generated prompts outperform fixed templates or eliminate all prompt bias. Infrared--text retrieval uses the infrared-oriented IR-TD captions described above rather than these generated class prompts.
The three prompt-group JSON files for each classification and segmentation dataset and the rewritten IR-TD retrieval captions are evaluation materials produced in this study.

Classification is reported as Top-1 accuracy. Retrieval is reported as Recall@K in both the infrared-to-text and text-to-infrared directions. Segmentation is reported as mean Intersection-over-Union (mIoU). The main text reports the averaged results; the prompt-level classification and segmentation values and the detailed ablation values are provided in Sections S1.1--S1.3 of the Supplementary Material.

\paragraph{Baselines.}
We compare TMT with seven baselines. OpenAI CLIP~\cite{radford2021clip} is the original ViT-L/14 model without infrared adaptation. LanguageBind~\cite{zhu2023languagebind} provides an infrared branch trained within a language-centered multimodal framework. un$^2$CLIP~\cite{li2025un2clip} uses an unCLIP generator to recover visual information while retaining the CLIP embedding space. DIVA~\cite{wang2024diffusion} uses diffusion feedback to improve CLIP visual representations from image-only data. ImageBind (official)~\cite{girdhar2023imagebind} denotes the released model, whose thermal branch was trained on LLVIP image--thermal pairs. ImageBind (trained) denotes our paired CLIP control: starting from OpenAI CLIP, we train its infrared image encoder to align directly with visible-image embeddings using the same IR500K infrared--visible training pairs as TMT. The two methods differ in their adaptation objectives; this control follows ImageBind's image-paired alignment principle but does not fine-tune the official ImageBind checkpoint. CLIP-FT fine-tunes OpenAI CLIP with LoRA on the IR-TD training split and is included as an infrared--language supervised reference where the result is available.

LanguageBind and CLIP-FT are marked with $\dagger$ in the tables because they use infrared--language supervision that is unavailable to TMT. They are therefore useful reference points, especially for retrieval, but are not strictly matched competitors in terms of supervision. TMT and ImageBind (trained) start from OpenAI CLIP, use the same IR500K infrared--visible training pairs, and require no paired infrared--text annotations; comparing them isolates the effect of the adaptation method. 
OpenAI CLIP, un$^2$CLIP, and DIVA are additional references without paired infrared--text supervision in this comparison. The released ImageBind model is an external reference with a different backbone and thermal training source.
We exclude IRGPT~\cite{cao2025irgpt} from the numerical comparison because its generative question-answering interface differs from the CLIP-style image--text alignment interface evaluated here. 

We also do not numerically compare with the concurrent T-CLIP~\cite{qazi2026tclip}: it is evaluated on different benchmarks with its own synthesized captions, and no public implementation was available at the time of submission. Its use of synthesized infrared--text supervision would also place it with the $\dagger$ reference methods rather than the matched group.


\paragraph{Implementation Details.}
All experiments use OpenAI CLIP ViT-L/14 as the base image encoder. The generative branch uses Stable Diffusion 2.1 unCLIP small with the Karlo v1-alpha prior. Only the CLIP image encoder is trainable; the VAE, diffusion model, text encoder, prior model, and visible-image CLIP teacher are frozen. Training images are resized to $256 \times 256$, and CLIP inputs are resized to $224 \times 224$. We optimize the model with SGD for 1000 steps using a learning rate of $1 \times 10^{-4}$, momentum 0.9, and zero weight decay. Training uses eight RTX 4090 GPUs 
with per-device batch size 8, gradient accumulation 10, and effective batch size 640. BF16 and TF32 acceleration are enabled, and the diffusion noise-prediction branch uses two inner diffusion steps during training.
The loss weights are $\lambda_{\text{diff}}=\lambda_{\text{semantic}}=1$ in all reported experiments.

\subsection{Zero-Shot Classification}

\begin{center}
\begin{minipage}{\columnwidth}
\centering
\tablefontsize
\captionof{table}{Zero-shot Top-1 accuracy (\%) on four source-disjoint benchmarks, averaged over three shared prompt groups. ImageBind (trained) is the ImageBind-style paired CLIP control, distinct from the official checkpoint. $\dagger$ marks infrared--language-supervised references. Best values under our protocol are bold.}
\label{tab:classification}
\resizebox{\columnwidth}{!}{\begin{tabular}{lccccc}
\toprule
Method & SMOD & JU-VNT-T & MFNet & PST900 & Avg. \\
\midrule
LanguageBind$^\dagger$ & 56.30 & 47.90 & 38.59 & 36.98 & 44.94 \\
un$^2$CLIP & 57.87 & 32.63 & 35.66 & 34.19 & 40.09 \\
DIVA & 60.07 & 43.73 & 34.08 & 32.27 & 42.54 \\
ImageBind (official) & 25.73 & 4.17 & 6.07 & 27.04 & 15.75 \\
ImageBind (trained) & \textbf{63.00} & 36.28 & 36.28 & 38.50 & 43.52 \\
CLIP-FT$^\dagger$ & 62.90 & \textbf{51.56} & 31.89 & 38.24 & 46.15 \\
\midrule
\textbf{TMT} & \textbf{63.00} & 50.17 & \textbf{42.04} & \textbf{40.49} & \textbf{48.92} \\
\bottomrule
\end{tabular}}
\end{minipage}
\end{center}

Table~\ref{tab:classification} prioritizes the four source-disjoint benchmarks. 
TMT averages 48.92\% on SMOD, JU-VNT-T, MFNet, and PST900, 2.78 points above the strongest reference average (CLIP-FT, 46.15\%) and 3.98 points above LanguageBind (44.94\%); differences are computed from unrounded values. It leads outright on two of these four datasets: MFNet (42.04\%) and PST900 (40.49\%). 
On MFNet and PST900, the gains over the strongest alternative are 3.45 points over LanguageBind and 1.99 points over ImageBind (trained), respectively. On SMOD, TMT and ImageBind (trained) both report 63.00\% at two-decimal precision; CLIP-FT leads on JU-VNT-T (51.56\% versus 50.17\%). The four source-overlapping benchmarks are shown in Table~\ref{tab:classification-overlap}; over all eight, TMT averages 60.40\% and leads outright on five, with the displayed tie on SMOD. The original LanguageBind paper reports 82.90\% on FLIRv1, 48.00\% on FLIRv2, and 87.20\% on LLVIP; these values are contextual and are excluded from our matched-protocol averages.

\begin{center}
\begin{minipage}{\columnwidth}
\centering
\tablefontsize
\captionof{table}{Supplementary Top-1 accuracy (\%) on four benchmarks sharing source families with IR500K. ImageBind (trained) is the ImageBind-style paired CLIP control. $\ddagger$ gives originally reported LanguageBind values under a different protocol and is excluded from the eight-dataset average. Best values under our protocol are bold.}
\label{tab:classification-overlap}
\resizebox{\columnwidth}{!}{\begin{tabular}{lccccc}
\toprule
Method & M3FD & FLIRv1 & FLIRv2 & LLVIP & All-8 avg. \\
\midrule
LanguageBind$^\dagger$ & 67.10 & 77.27 & \textbf{45.20} & 83.23 & 56.57 \\
LanguageBind$^\ddagger$ & -- & 82.90 & 48.00 & 87.20 & -- \\
un$^2$CLIP & 54.87 & 80.20 & 35.70 & 79.83 & 51.37 \\
DIVA & 61.77 & 71.77 & 30.33 & 85.20 & 52.40 \\
ImageBind (official) & 5.74 & 6.27 & 1.10 & 72.86 & 18.62 \\
ImageBind (trained) & 57.54 & 73.94 & 31.22 & 72.09 & 51.11 \\
CLIP-FT$^\dagger$ & 57.46 & 67.05 & 30.01 & 87.16 & 53.28 \\
\midrule
\textbf{TMT} & \textbf{71.53} & \textbf{84.87} & 42.57 & \textbf{88.57} & \textbf{60.40} \\
\bottomrule
\end{tabular}}
\end{minipage}
\end{center}

The primary group spans road scenes (SMOD and MFNet), indoor objects (JU-VNT-T), and the subterranean robotic imagery of PST900. MFNet and PST900 were originally introduced for pixel-labeled scene understanding~\cite{ha2017mfnet,shivakumar2020pst900}; here their labeled regions are used for infrared classification. This mixture probes whether category transfer persists outside the IR500K source families. TMT's lead on both new datasets is consistent with improved transfer, although their moderate absolute accuracies (42.04\% and 40.49\%) show that region-level recognition remains difficult. Prompt-level results are given in Section S1.1 of the Supplementary Material.

TMT does not lead outright on every benchmark. On SMOD, TMT and ImageBind (trained) both reach 63.00\% at the reported precision; the trained control scores higher with Prompts 1 and 2, whereas TMT scores higher with Prompt 3. On JU-VNT-T, CLIP-FT has the advantage of infrared--text training on IR-TD, of which JU-VNT-T is a source, so this comparison has a different supervision and source relationship. On FLIRv2, LanguageBind reaches 45.20\% under our protocol and 48.00\% in its original report, while TMT reaches 42.57\%. The finer vehicle and roadside taxonomy may demand distinctions that a global modality shift does not fully supply. These exceptions bound the claim to average category transfer under the stated protocols.

ImageBind (official) reaches 72.86\% on LLVIP but 1.10--27.04\% on the other seven classification datasets. This pattern is consistent with its thermal branch being trained on LLVIP image--thermal pairs~\cite{girdhar2023imagebind}; it does not establish that training-source coverage alone explains the gap, because the official model also differs in backbone and preprocessing. ImageBind (trained) scores higher than the official checkpoint on seven of eight datasets, but slightly lower on LLVIP (72.09\% versus 72.86\%). Under our shared prompt protocol, TMT exceeds the official checkpoint on LLVIP by 15.71 points and the trained control by 16.48 points. LLVIP is a training source family for the official checkpoint and for the shared IR500K subset used by TMT and ImageBind (trained), so this comparison is not a test of source-disjoint transfer.

\subsection{Infrared--Text Retrieval}

\begin{table*}[t]
\centering
\tablefontsize
\caption{Infrared--text retrieval on IR-TD. Recall@K is reported in both retrieval directions. ImageBind (official) uses the released checkpoint; ImageBind (trained) is our paired CLIP alignment control. The upper block contains methods without infrared--text supervision during adaptation; the lower block contains supervised references. Methods marked with $\dagger$ use infrared--language supervision during pre-training or fine-tuning. Best results among the methods without infrared--text supervision are in \textbf{bold}.}
\label{tab:retrieval}
\begin{tabular*}{\textwidth}{@{\extracolsep{\fill}}lcccccccc@{}}
\toprule
\multirow{2}{*}{Method} & \multicolumn{4}{c}{IR $\rightarrow$ Text} & \multicolumn{4}{c}{Text $\rightarrow$ IR} \\
\cmidrule(lr){2-5}\cmidrule(lr){6-9}
 & R@1 & R@5 & R@10 & R@20 & R@1 & R@5 & R@10 & R@20 \\
\midrule
\multicolumn{9}{l}{\textit{Without infrared--text supervision}} \\
OpenAI CLIP~\cite{radford2021clip} & 5.00 & 11.72 & 16.72 & 25.47 & 10.47 & 20.31 & 30.31 & 43.44 \\
un$^2$CLIP~\cite{li2025un2clip} & 3.59 & 10.00 & 15.00 & 22.03 & 10.31 & 21.41 & 29.22 & 39.38 \\
DIVA~\cite{wang2024diffusion} & 4.84 & 11.41 & 16.09 & 25.00 & 9.06 & 19.22 & 29.84 & 42.81 \\
ImageBind (official)~\cite{girdhar2023imagebind} & 1.09 & 4.06 & 6.25 & 9.53 & 1.41 & 4.37 & 7.66 & 11.25 \\
ImageBind (trained) & 4.39 & 13.57 & 19.24 & 25.93 & 6.35 & 15.87 & 22.17 & 31.01 \\
\textbf{TMT} & \textbf{6.72} & \textbf{16.41} & \textbf{23.59} & \textbf{32.34} & \textbf{10.94} & \textbf{24.22} & \textbf{32.81} & \textbf{44.22} \\
\midrule
\multicolumn{9}{l}{\textit{Infrared--language-supervised references}} \\
LanguageBind$^\dagger$~\cite{zhu2023languagebind} & 14.69 & 31.41 & 41.09 & 51.56 & 13.59 & 30.31 & 39.06 & 50.00 \\
CLIP-FT$^\dagger$ & 18.12 & 41.72 & 53.28 & 68.28 & 19.37 & 42.03 & 56.41 & 72.03 \\
\bottomrule
\end{tabular*}
\end{table*}

Retrieval tests whether an infrared image embedding enters the appropriate linguistic neighborhood when the text encoder remains fixed. The table is therefore read in two groups. Within the group without infrared--text supervision, TMT is the strongest method in every reported Recall@K column. Relative to OpenAI CLIP, TMT raises IR-to-Text R@10 from 16.72\% to 23.59\% and Text-to-IR R@5 from 20.31\% to 24.22\%; at R@20, the gains are 6.87 and 0.78 percentage points in the two directions, respectively. ImageBind (trained) reaches 25.93\% and 31.01\% at R@20, compared with 9.53\% and 11.25\% for the official checkpoint; TMT reaches 32.34\% and 44.22\%. TMT and ImageBind (trained) use the same IR500K training pairs but different adaptation methods. The official checkpoint differs from both in architecture and thermal training source, so its scores serve as an external comparison under the shared evaluation protocol.

The lower block provides a supervised reference rather than a comparison under matched supervision conditions. LanguageBind and CLIP-FT use detailed, caption-level infrared--text alignment, which directly supplies the type of correspondence measured by retrieval. CLIP-FT additionally fine-tunes on the IR-TD training split and thus benefits from similarity between its training and test distributions, including caption style and dataset-specific language conventions. This difference is visible at R@20: CLIP-FT reaches 68.28\% for IR-to-Text retrieval, compared with 32.34\% for TMT, while LanguageBind reaches 51.56\%. The gap should not be interpreted as evidence that TMT fails to improve image--text compatibility; it quantifies the additional information provided by instance-level infrared captions and in-domain supervised adaptation. TMT is intended for the complementary setting in which aligned infrared--visible images are available but infrared text annotation is absent or costly. Concurrent work that synthesizes thermal captions likewise reports retrieval gains on its own benchmarks~\cite{qazi2026tclip}, consistent with the supervised-reference pattern described above.

\subsection{Zero-Shot Semantic Segmentation}

All segmentation results are produced with NACLIP~\cite{hajimiri2025naclip}, a training-free open-vocabulary segmentation method that adapts a CLIP image encoder to dense prediction by enforcing neighbourhood-aware patch localization in its self-attention. Because NACLIP requires no task-specific training, the same configuration is applied to every compared encoder, with each dataset's class prompts serving as the text inputs; differences in the resulting masks can therefore be attributed to the image representation rather than to a segmentation head.

\begin{table}[H]
\centering
\tablefontsize
\caption{Zero-shot open-vocabulary semantic segmentation mIoU~(\%) on infrared images. Both datasets are absent from TMT training. ImageBind (trained) denotes the paired CLIP alignment control. Methods marked with $\dagger$ use infrared--language supervision.}
\label{tab:segmentation}
\begin{tabular}{lcc}
\toprule
Method & CART & SemanticRT \\
\midrule
OpenAI CLIP~\cite{radford2021clip} & 16.06 & 15.16 \\
LanguageBind$^\dagger$~\cite{zhu2023languagebind} & 18.02 & 16.87 \\
un$^2$CLIP~\cite{li2025un2clip} & 9.22 & 17.29 \\
DIVA~\cite{wang2024diffusion} & 16.40 & 15.49 \\
ImageBind (official)~\cite{girdhar2023imagebind} & 1.59 & 5.63 \\
ImageBind (trained) & 16.94 & 15.80 \\
CLIP-FT$^\dagger$ & 16.56 & 16.19 \\
\midrule
\textbf{TMT} & \textbf{18.23} & \textbf{17.38} \\
\bottomrule
\end{tabular}
\end{table}

Table~\ref{tab:segmentation} evaluates whether the adaptation remains useful when the output must be spatially resolved. TMT obtains the best mIoU on both CART and SemanticRT, which are absent from the IR500K training subset. ImageBind (official) reaches 1.59\% and 5.63\%, respectively; ImageBind (trained) reaches 16.94\% and 15.80\%. The result indicates that the adapted image representation transfers beyond cropped-object classification to two cross-dataset dense-prediction settings. The improvement is nevertheless modest in absolute terms. Under the same NACLIP pipeline, visible-image inputs reach only 25.66\% on CART and 20.60\% on SemanticRT. The infrared results should therefore be read as relative evidence that TMT preserves useful spatial information, not as a substitute for supervised segmentation.

The segmentation results also expose a boundary of the current method. Classification and retrieval mainly require global identity and image--text compatibility, whereas segmentation depends on boundary placement, part structure, and small-object localization. Thermal diffusion, sensor blur, saturation, and weak texture can remove precisely these local cues. TMT improves the placement of infrared representations in the CLIP embedding space, but the paired semantic alignment and diffusion noise-prediction objectives do not directly supervise segmentation boundaries. The component ablation in Section~5 and the linear-probe analysis in Appendix~\ref{app:noise-free} isolate the contribution of the diffusion branch; the remaining spatial limitation motivates future work on infrared imaging mechanisms and image structure. 

\section{Analysis and Discussion}

Section~4 established TMT's downstream performance.
We now make further examinations of the proposed method TMT, including the ablation studies, the source of the modality transfer direction together with qualitative evidence of its effect, the sensitivity to the transfer weight, and feature geometry. 
These experiments separate the effect of paired-image supervision from text-derived modality transfer and identify the tasks on which the choices matter.

\subsection{Ablation Studies}

\begin{table}[H]
\centering
\tablefontsize
\caption{Component ablation averaged across tasks. Each row switches the text-derived shift $\alpha\Delta\mathbf{v}$ and the two supervision branches $\mathcal{L}_{\text{sem}}$ ($\mathcal{L}_{\text{semantic}}$) and $\mathcal{L}_{\text{diff}}$ on (\checkmark) or off ($\times$); the first row is the unadapted OpenAI CLIP encoder without any further training. Retrieval averages both directions on IR-TD; primary classification averages the four source-disjoint datasets (SMOD, JU-VNT-T, MFNet, and PST900); all-dataset classification averages all eight datasets; segmentation averages CART and SemanticRT. All values are percentages.}
\label{tab:ablation}
\resizebox{\columnwidth}{!}{\begin{tabular}{cccccccc}
\toprule
Shift & $\mathcal{L}_{\text{sem}}$ & $\mathcal{L}_{\text{diff}}$ & R@1 $\uparrow$ & R@20 $\uparrow$ & Primary cls. $\uparrow$ & All cls. $\uparrow$ & Seg. mIoU $\uparrow$ \\
\midrule
$\times$ & $\times$ & $\times$ & 7.73 & 34.45 & 42.48 & 52.33 & 15.61 \\
\checkmark & \checkmark & $\times$ & \textbf{8.83} & 37.81 & 48.23 & 59.78 & 17.69 \\
\checkmark & $\times$ & \checkmark & 7.58 & 34.45 & 44.55 & 54.98 & 16.73 \\
$\times$ & \checkmark & \checkmark & 7.88 & 34.14 & 46.18 & 55.79 & 16.27 \\
\checkmark & \checkmark & \checkmark & \textbf{8.83} & \textbf{38.28} & \textbf{48.92} & \textbf{60.40} & \textbf{17.81} \\
\bottomrule
\end{tabular}}
\end{table} 

Table~\ref{tab:ablation} separates the contribution of paired semantic alignment and diffusion-based noise prediction. Removing both branches reduces all-dataset classification from 60.40\% to 52.33\% and average segmentation from 17.81\% to 15.61\%. Semantic supervision alone raises all-dataset classification by 7.45 points and gives the same mean R@1 as the full model. It directly anchors the transferred infrared feature to the paired visible-image teacher, which is consistent with its stronger effect on category recognition.

The diffusion branch has a different profile. Used alone, it does not improve mean retrieval R@1 over the no-supervision variant, but it raises average segmentation mIoU from 15.61\% to 16.73\%. The result is consistent with a branch that supplies visual-detail feedback rather than a direct category target. The full model combines the two roles: semantic supervision provides the strongest category-level correction, while diffusion feedback contributes local visual information and increases the full-model R@20 and segmentation scores. 
The fourth row of Table~\ref{tab:ablation} removes the shift while keeping both supervision branches. Relative to the full model, this lowers primary classification by 2.74 points and bidirectional R@20 by 4.14 points, isolating the contribution of the text-derived shift under fixed supervision.

\subsection{Analysis of Text-Derived Modality Transfer}

This subsection examines the Text-Derived Modality Transfer module in two complementary parts. Section~5.2.1 asks where the transfer direction itself comes from and compares control variants that replace its source or its content with the full model under the same downstream evaluation. Section~5.2.2 then provides qualitative evidence that the transfer produces visible-oriented representations, through t-SNE embeddings and generated pseudo-visible images. 

\subsubsection{Controls on the Transfer Direction}

To test whether the benefit of the shift comes from its textual origin, we compare TMT with controls that replace the text-derived direction under the same downstream evaluation protocol. The first, ImageBind (trained), removes the shift and the generative branch and instead directly aligns infrared and visible CLIP features; it is an ImageBind-style training control initialized from OpenAI CLIP and trained on the same IR500K infrared--visible pairs as TMT, rather than on the released ImageBind checkpoint. The second, image-mean transfer, keeps the shift mechanism but replaces the text-derived direction with the difference between the mean visible and mean infrared image embeddings of the paired training set. These controls test, respectively, direct paired-image alignment without any explicit transfer direction and a direction estimated from training images rather than from text. A further group of controls probes the content of the direction rather than its source: three random unit vectors used as the shift direction with different seeds, a minimal modality-name pair (``an image of visible'' versus ``an image of infrared''), and the reversed direction $-\Delta\mathbf{v}$. These ask whether an arbitrary shift, a bare naming of the two modalities, or the opposite orientation can substitute for the intended infrared-to-visible direction. The official ImageBind results are reported separately in Section~4. 

\begin{table}[t]
\centering
\tablefontsize
\caption{Transfer-direction controls. ImageBind (trained) is the ImageBind-style CLIP alignment control, not the official checkpoint. The variant without the shift is reported with the component ablation in Table~\ref{tab:ablation}. Random direction averages three seeds; the modality-name pair uses ``an image of visible'' versus ``an image of infrared'' in place of the modality descriptions; the reversed direction negates $\Delta\mathbf{v}$. Retrieval is bidirectional mean R@K; classification is Top-1 accuracy; segmentation is mean mIoU. All values are percentages. Primary classification averages SMOD, JU-VNT-T, MFNet, and PST900.}
\label{tab:transfer-controls}
\resizebox{\columnwidth}{!}{\begin{tabular}{lccccc}
\toprule
Variant & R@1 $\uparrow$ & R@20 $\uparrow$ & Primary cls. $\uparrow$ & All cls. $\uparrow$ & Seg. mIoU $\uparrow$ \\
\midrule
ImageBind (trained) & 5.37 & 28.47 & 43.52 & 51.11 & 16.37 \\
Image-mean transfer & \textbf{9.14} & 35.31 & 46.40 & 56.38 & 16.24 \\
Random direction (3 seeds) & 8.90 & 36.85 & 46.06 & 55.53 & 16.27 \\
Text pair (modality names) & 8.52 & 37.26 & 46.57 & 56.51 & 16.12 \\
Reversed direction & 7.26 & 33.13 & 40.22 & 51.12 & 15.28 \\
\textbf{TMT} & 8.83 & \textbf{38.28} & \textbf{48.92} & \textbf{60.40} & \textbf{17.81} \\
\bottomrule
\end{tabular}}
\end{table} 

The released ImageBind checkpoint provides an external comparison rather than another transfer-direction control. Its means are 15.75\% on the four primary classification datasets, 18.62\% across all eight, 10.39\% for bidirectional retrieval R@20, and 3.61\% segmentation mIoU. ImageBind (trained) reaches 43.52\%, 51.11\%, 28.47\%, and 16.37\%, respectively, while TMT reaches 48.92\%, 60.40\%, 38.28\%, and 17.81\%. The official checkpoint differs from ImageBind (trained) in backbone and thermal training source, so their gap is descriptive. ImageBind (trained) shares TMT's CLIP starting point and IR500K training pairs, making it the paired-image-alignment comparison for TMT's different adaptation method. 

As analyzed in Section~5.1, removing the shift lowers primary classification by 2.74 points and bidirectional R@20 by 4.14 points. A paired-image mean direction recovers part of this loss, but trails the text direction by 2.52 primary-classification points and 2.97 retrieval points. ImageBind (trained) is weaker on average, showing that direct paired-image alignment alone does not reproduce the combination of transfer and generative--perceptual supervision. These are comparisons of complete training variants, not an isolated estimate of the causal effect of text alone: ImageBind (trained) also changes the supervision architecture. 

The image-mean control in Table~\ref{tab:transfer-controls} shows that paired-image differences provide a useful transfer signal for classification and retrieval, but are less effective than the text-derived direction in this setting. 
One possible explanation is that the image-derived direction depends on both the sampled scenes and the original encoder's representations of the two modalities. Pairing establishes scene correspondence at the input level, but does not guarantee that the feature difference isolates modality alone. Averaging across pairs can reduce instance-specific variation, while systematic effects associated with the training distribution or the encoder's infrared representations may remain. Generic modality descriptions instead specify the imaging properties to be contrasted without deriving the direction from those sampled image features. Paired visible images remain the content reference for supervision in both variants.

A final group of controls asks whether the content of the direction matters at all. Averaged over three seeds, random directions reach 46.06\% primary classification, 55.53\% all-dataset classification, 36.85\% bidirectional R@20, and 16.27\% segmentation mIoU---close to the no-shift variant on classification and segmentation and slightly higher on retrieval, but below TMT on most metrics. The minimal modality-name pair does better than the random directions: it improves over them on the two classification averages and retrieval R@20 (46.57\% primary, 56.51\% all-dataset, 37.26\% R@20), showing that even a bare naming of the two modalities carries useful modality-level signal; it nevertheless trails TMT's designed description pair on every metric. The reversed direction is the weakest control and falls below the no-shift variant on all five metrics (40.22\% primary, 33.13\% R@20, 15.28\% mIoU). Arbitrary vectors and the opposite orientation therefore fail to reproduce the improvement, and bare modality names recover only part of it: the gain depends on the modality-level imaging properties verbalized in the designed pair rather than on the mere presence of an additive or text-derived shift. 

\subsubsection{Qualitative Evidence of the Transfer}

The controls above isolate the contribution of the direction's source to downstream performance. As a qualitative complement, we next inspect whether images generated from the transferred features exhibit a visible-oriented shift. We compare the original infrared images, the pseudo-visible images generated from their transferred features, and their paired visible images. For each selected source subset, all three image types are re-encoded by the same frozen CLIP ViT-L/14 encoder for visualization, so encoder choice is held fixed within the plot. Because the generator and this second encoding step intervene, this experiment provides indirect, qualitative evidence rather than a direct measurement of the training-time pseudo-visible feature vectors.

\begin{figure}[t]
\centering
\includegraphics[width=\columnwidth]{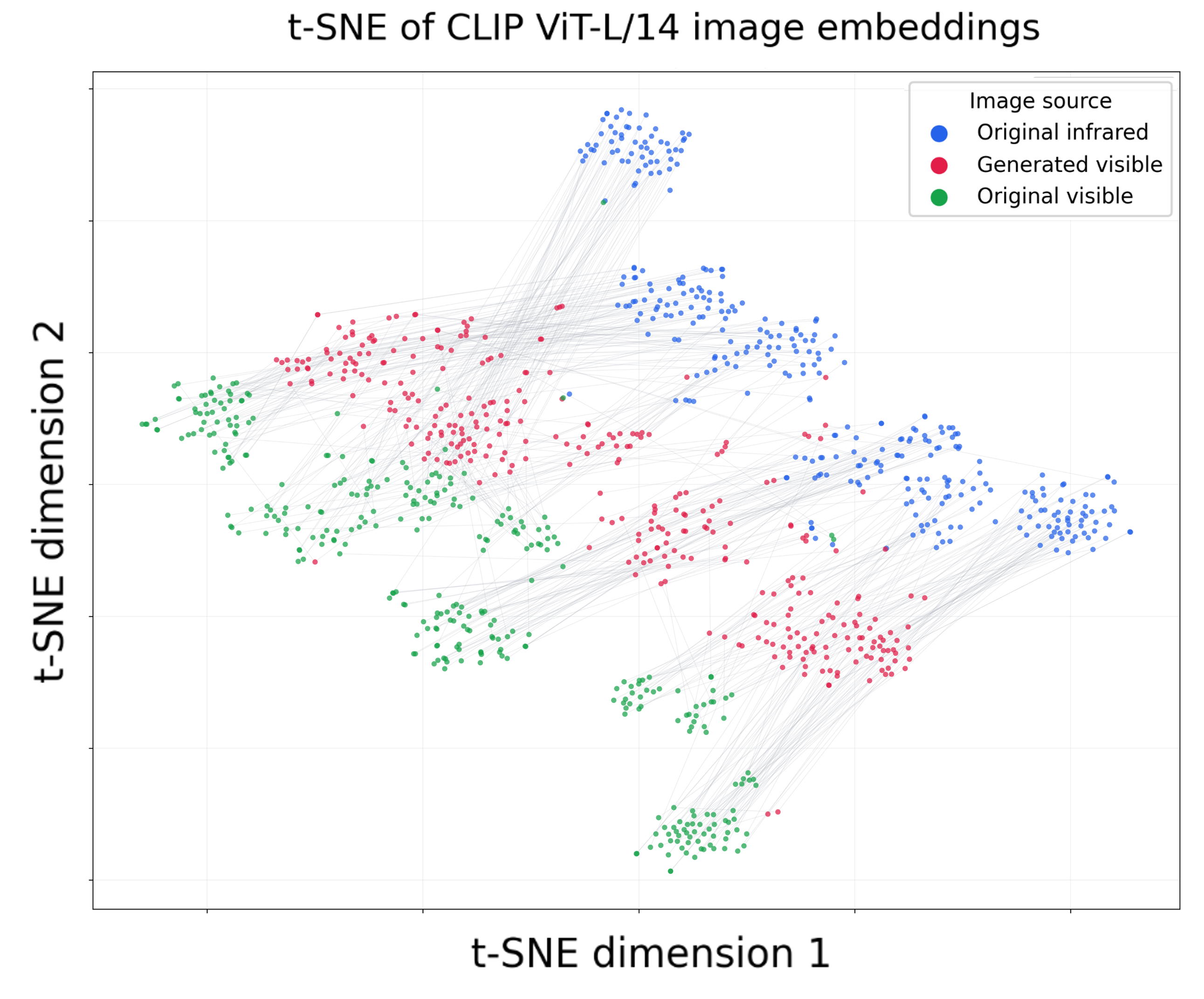}
\caption{t-SNE visualization of CLIP ViT-L/14 embeddings for original infrared images, generated pseudo-visible images, and original visible images. Matched subsets are sampled from the source datasets for readability, and all images are encoded with the same CLIP backbone. Gray lines connect matched triplets, while colors indicate image source. The embeddings of generated pseudo-visible images appear closer to the visible-image distribution than those of the original infrared images; this is an indirect visualization of the transfer.}
\label{fig:clip-tsne-transfer}
\end{figure}

Figure~\ref{fig:clip-tsne-transfer} shows that embeddings of the generated pseudo-visible images generally lie farther from the infrared-image distribution and closer to the paired visible-image distribution. This pattern is consistent with the intended visible-oriented transfer, but it does not directly establish the displacement of $\mathbf{v}_{\text{pseudo vis}}$: image generation and re-encoding intervene between that feature and the plotted points. The figure is therefore an indirect, qualitative check; t-SNE can also distort global distances.

\begin{figure}[t]
\centering
\setlength{\tabcolsep}{1pt}
\renewcommand{\arraystretch}{0.92}
\tablefontsize
\begin{tabular}{ccc}
{\scriptsize Original infrared} & {\scriptsize Pseudo-visible} & {\scriptsize Original visible} \\
\includegraphics[width=0.31\columnwidth]{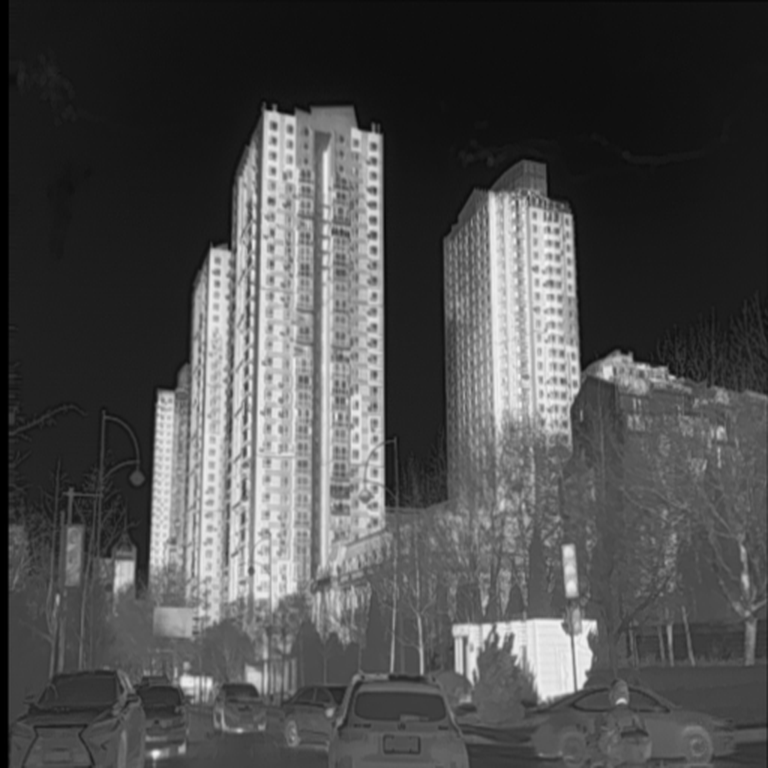} &
\includegraphics[width=0.31\columnwidth]{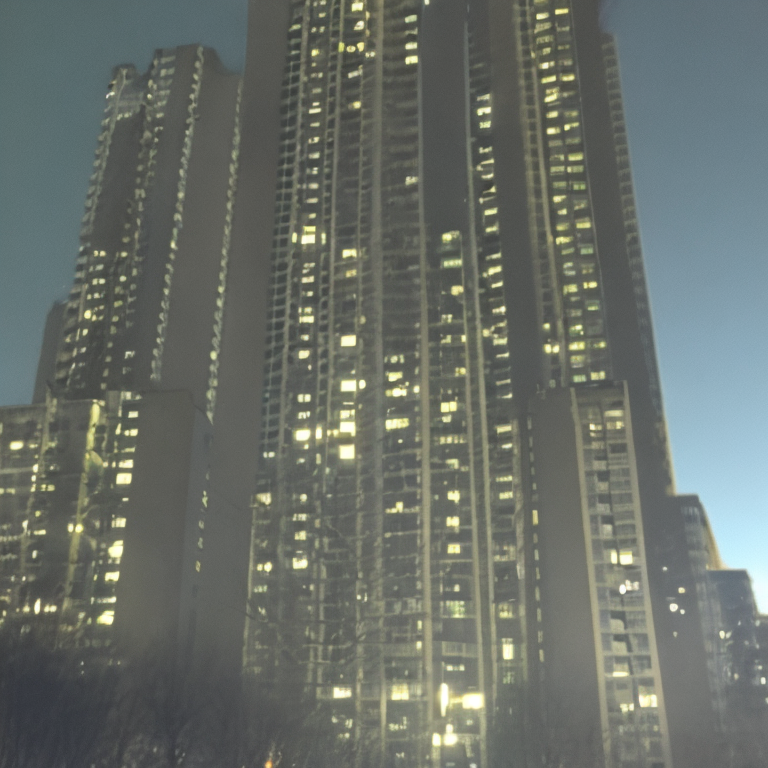} &
\includegraphics[width=0.31\columnwidth]{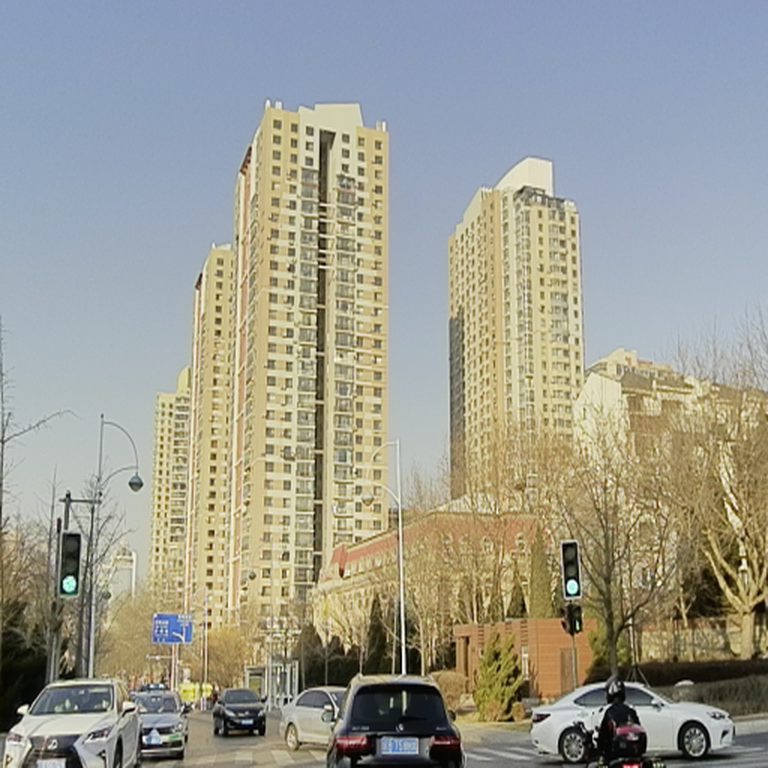} \\

\includegraphics[width=0.31\columnwidth]{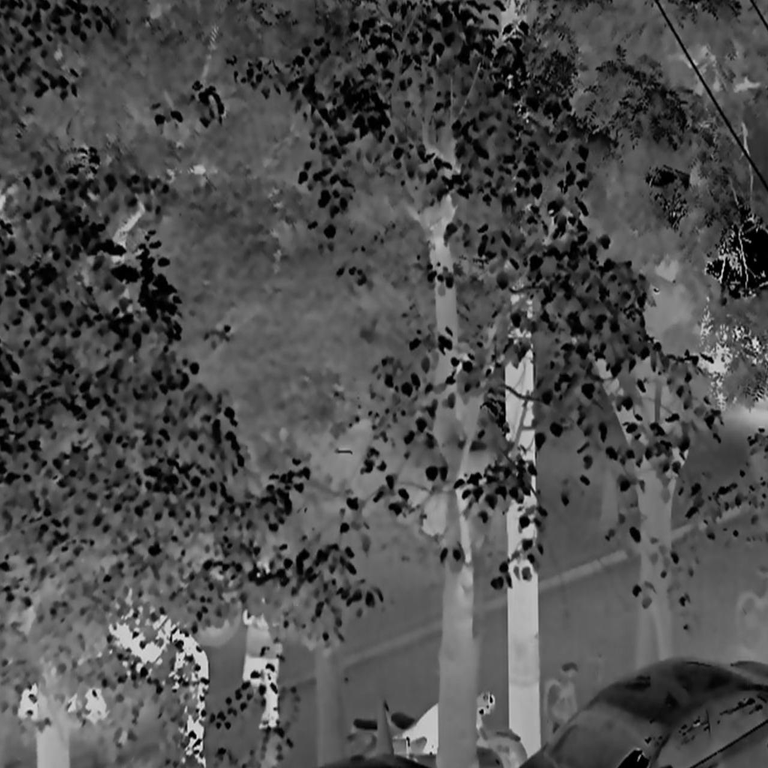} &
\includegraphics[width=0.31\columnwidth]{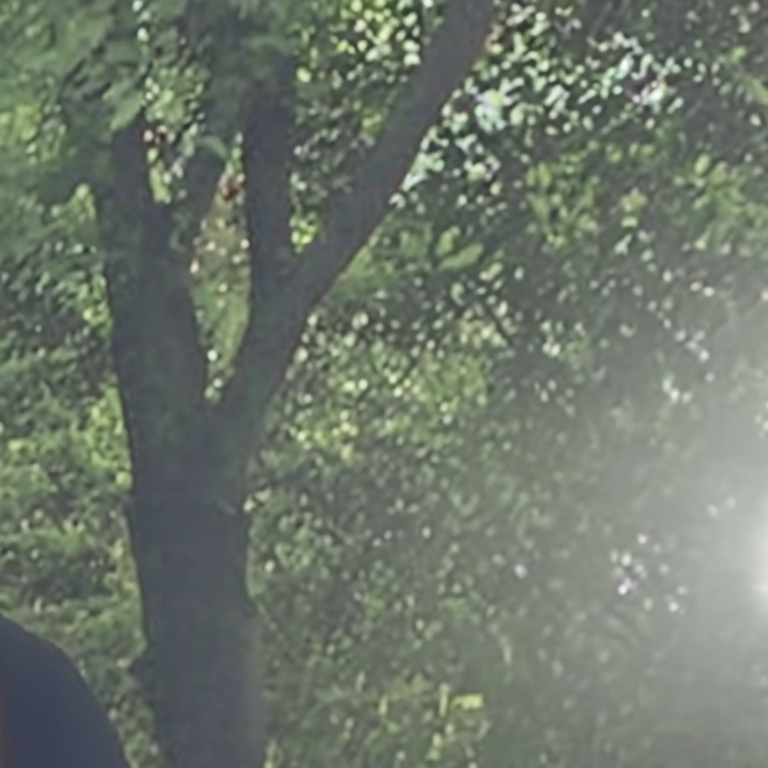} &
\includegraphics[width=0.31\columnwidth]{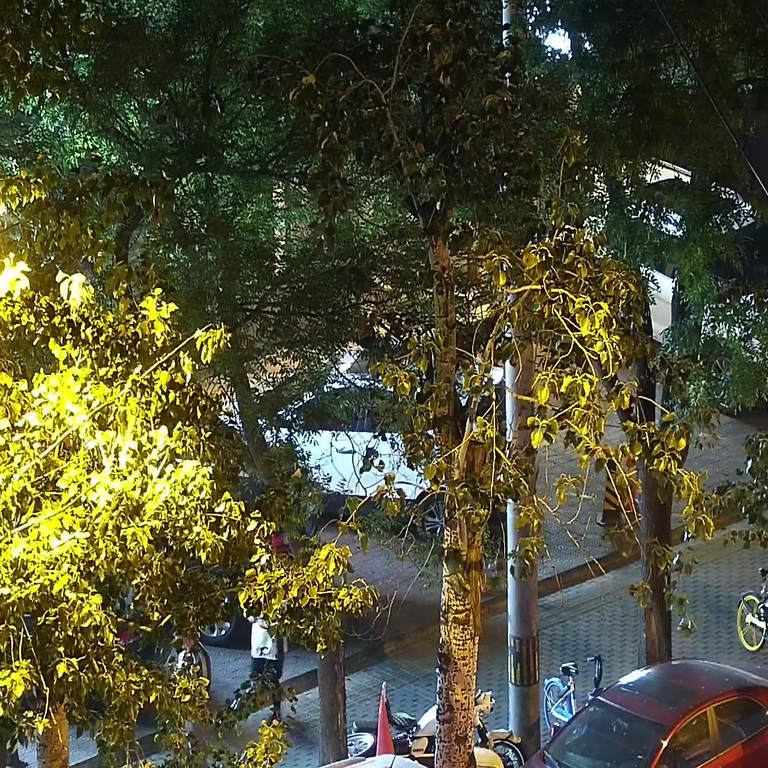} \\

\includegraphics[width=0.31\columnwidth]{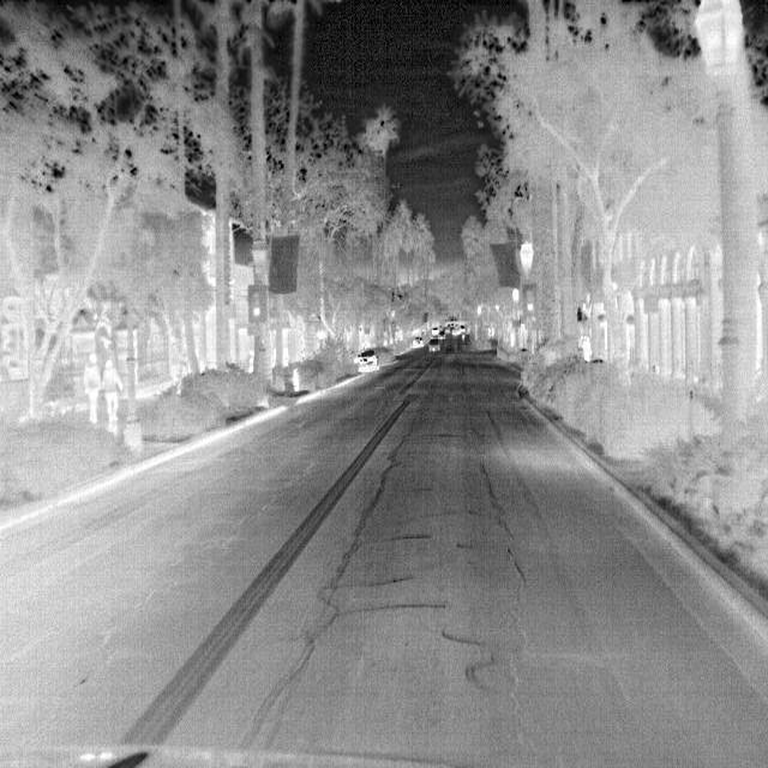} &
\includegraphics[width=0.31\columnwidth]{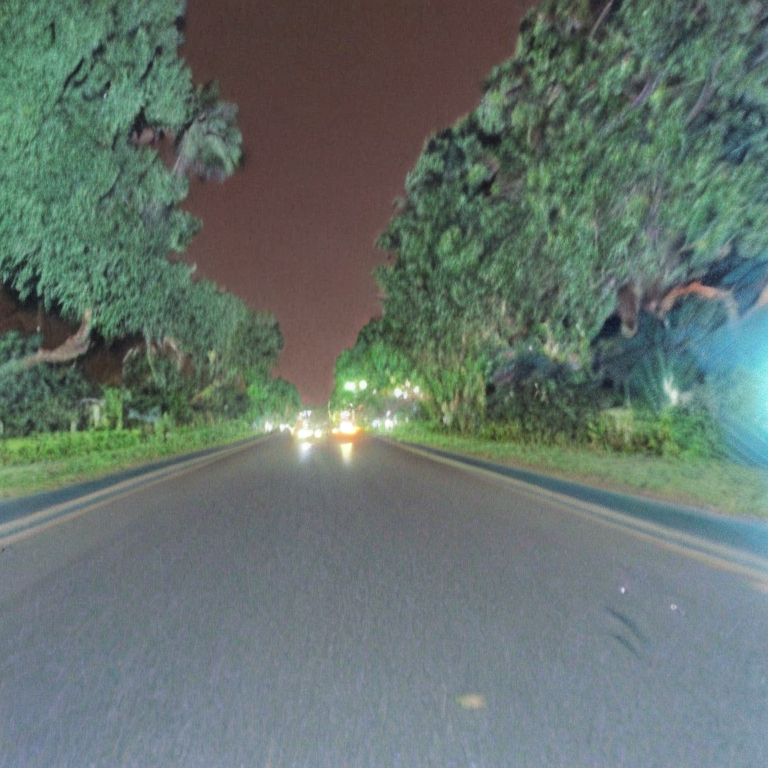} &
\includegraphics[width=0.31\columnwidth]{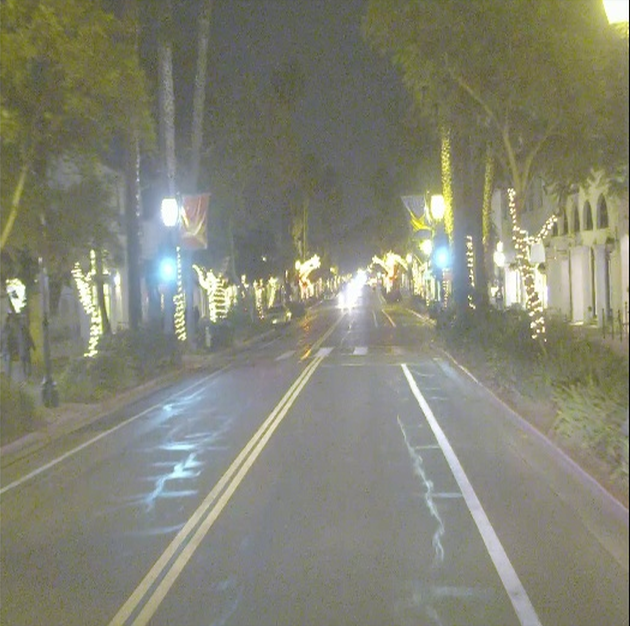}
\end{tabular}
\caption{IR500K examples of the original infrared image, the pseudo-visible image, and the paired visible image from left to right. The pseudo-visible result preserves object identity and coarse structure, but it is not expected to match the visible image pixel by pixel.}
\label{fig:ir-transfer-triptych}
\end{figure}

Figure~\ref{fig:ir-transfer-triptych} shows that the pseudo-visible images retain the object identity and coarse spatial structure of the infrared inputs while moving their appearance toward the paired visible images. These examples complement the indirect t-SNE visualization in Figure~\ref{fig:clip-tsne-transfer}: the generated images show visible-oriented appearance while retaining recognizable scene content. Pixelwise identity is not the target of this operation. DDPM-style models sample a learned reverse process rather than deterministically invert an input, and unCLIP-style decoders map a semantic image embedding to a distribution of plausible visible images~\cite{ho2020ddpm,ramesh2022unclip,rombach2022ldm}. Several visible realizations can therefore satisfy the same infrared-conditioned semantic content. The relevant visual evidence is recognizable object identity and coarse structure together with a visible-oriented change in generated appearance, rather than exact RGB reconstruction. These observations are consistent with the intended transfer, while the task and ablation results provide its quantitative evaluation.

\subsection{Sensitivity to the Transfer Weight}

We vary the fixed coefficient $\alpha$ in Eq.~\ref{eq:semantic_transfer} from 0.0 to 0.5 in steps of 0.1, keeping the direction-construction procedure and downstream evaluation protocol fixed. The zero setting corresponds to removing the shift (the fourth row of Table~\ref{tab:ablation}). This sweep tests whether the benefit depends on a narrow transfer strength or persists over neighboring values. For each setting, we evaluate the four primary classification datasets, IR-TD retrieval, and both segmentation datasets; the reported values average three prompt groups where applicable. Per-dataset and per-metric values of the sweep are reported in Section S1.6 of the Supplementary Material. 

\begin{table}[H]
\centering
\tablefontsize
\caption{Transfer-weight sensitivity. Primary classification averages four source-disjoint datasets; retrieval averages both R@20 directions; segmentation averages CART and SemanticRT mIoU. All values are percentages.}
\label{tab:alpha-sensitivity}
\begin{tabular}{cccc}
\toprule
$\alpha$ & Primary cls. $\uparrow$ & Retrieval R@20 $\uparrow$ & Seg. mIoU $\uparrow$ \\
\midrule
0.0 & 46.18 & 34.14 & 16.27 \\
0.1 & 48.49 & 37.97 & \textbf{17.96} \\
0.2 & 48.15 & 36.80 & 17.92 \\
0.3 & \textbf{48.92} & \textbf{38.28} & 17.81 \\
0.4 & 47.61 & 37.42 & 17.76 \\
0.5 & 45.41 & 34.45 & 17.23 \\
\bottomrule
\end{tabular}
\end{table}

Within $\alpha \in [0.1, 0.4]$, all three metrics vary only modestly: primary classification stays between 47.61\% and 48.92\%, retrieval R@20 between 36.80\% and 38.28\%, and segmentation between 17.76\% and 17.96\%, indicating that the benefit does not rely on a finely tuned transfer strength. Primary classification and bidirectional retrieval R@20 peak at $\alpha=0.3$ (48.92\% and 38.28\%), while segmentation peaks at $\alpha=0.1$ (17.96\%) and varies by at most 0.20 points across this range. Degradation appears mainly at the two ends: removing the shift lowers primary classification to 46.18\%, and moving to 0.5 reduces it to 45.41\%, below the zero-shift result, consistent with an overly large modality displacement. We therefore fix $\alpha=0.3$ as a single setting that is near-optimal for the two global matching tasks while remaining close to the best segmentation values.

\begin{center}
\begin{minipage}{\columnwidth}
\centering
\setlength{\tabcolsep}{1pt}
\begin{tabular}{@{}cccc@{}}
\includegraphics[width=0.24\columnwidth]{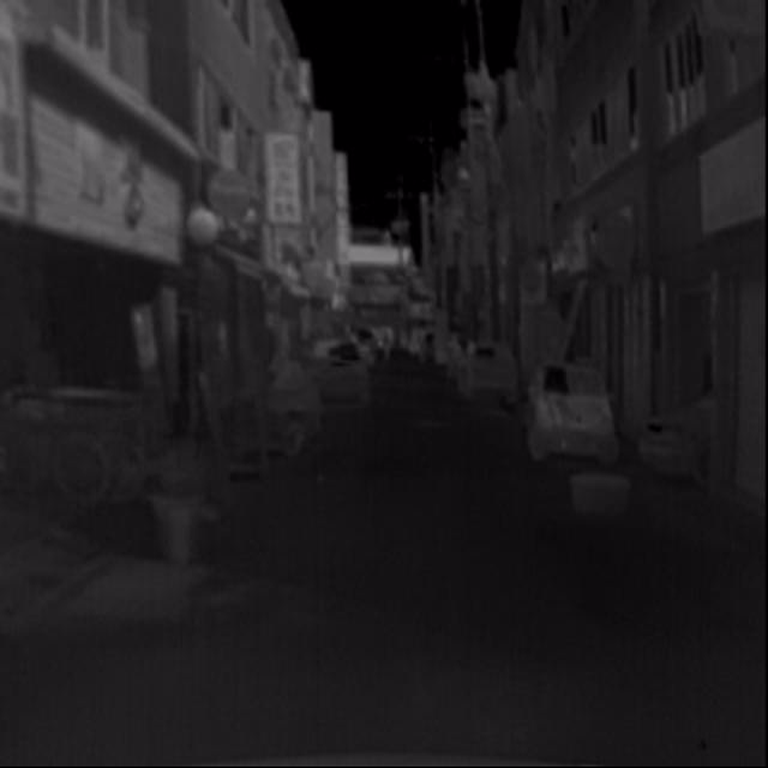} &
\includegraphics[width=0.24\columnwidth]{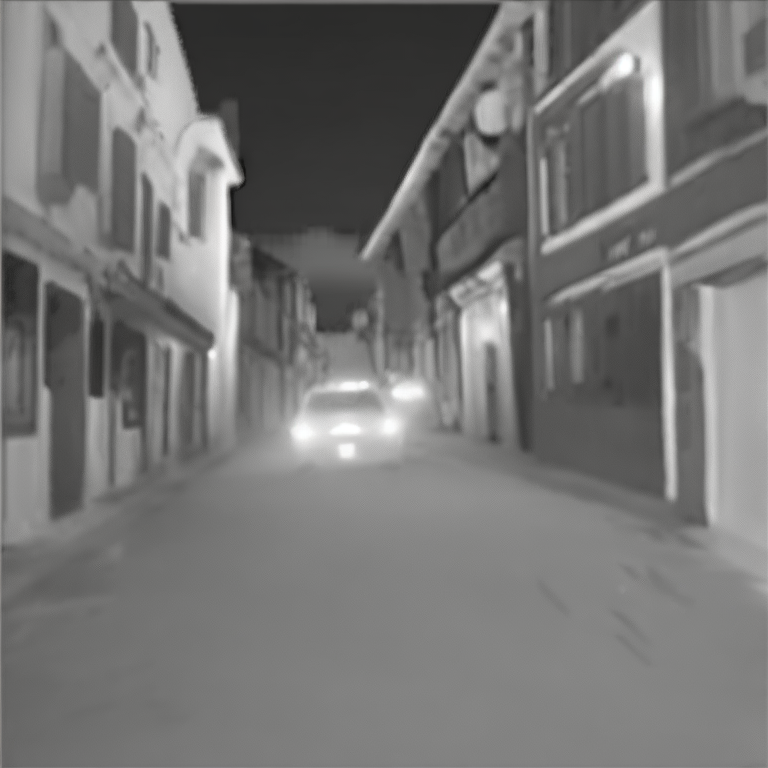} &
\includegraphics[width=0.24\columnwidth]{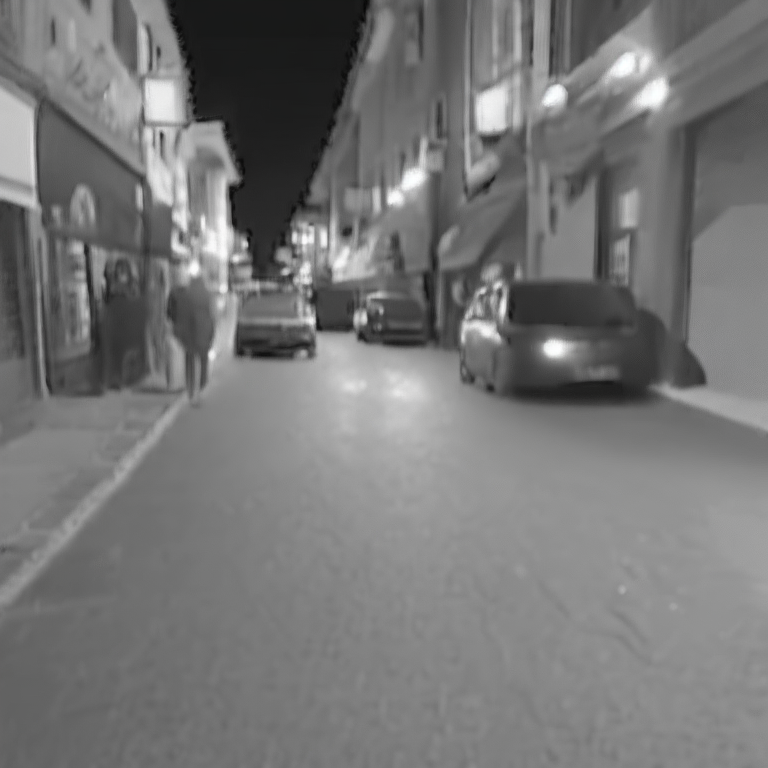} &
\includegraphics[width=0.24\columnwidth]{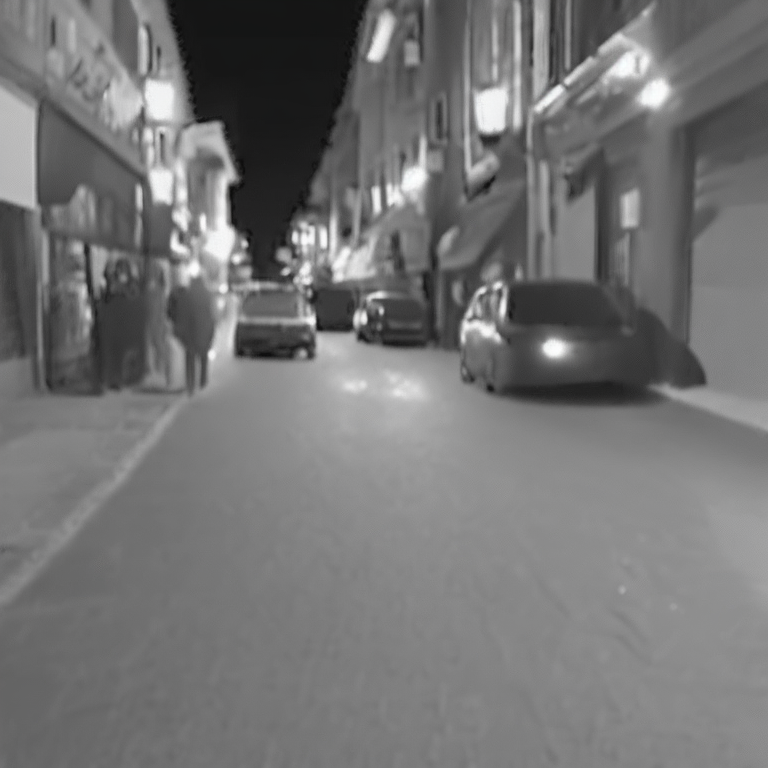} \\
{\footnotesize IR input} & {\footnotesize $\alpha=0.0$} & {\footnotesize $\alpha=0.1$} & {\footnotesize $\alpha=0.2$} \\[2pt]
\includegraphics[width=0.24\columnwidth]{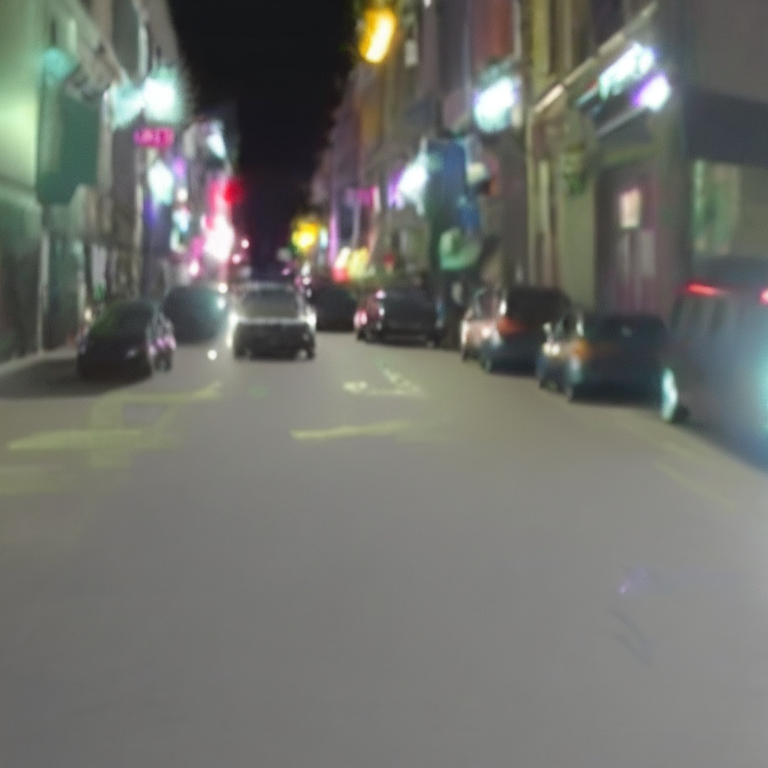} &
\includegraphics[width=0.24\columnwidth]{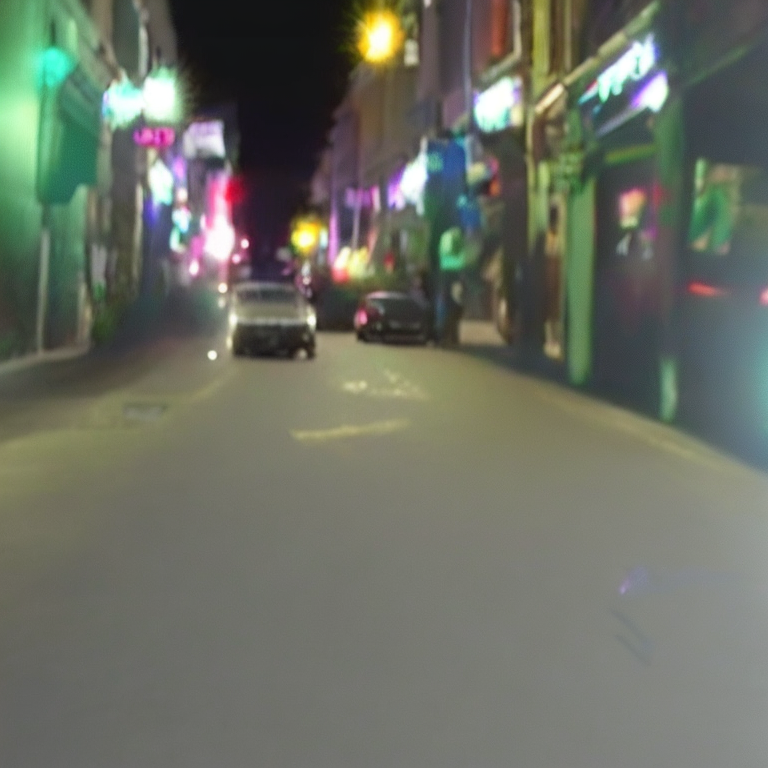} &
\includegraphics[width=0.24\columnwidth]{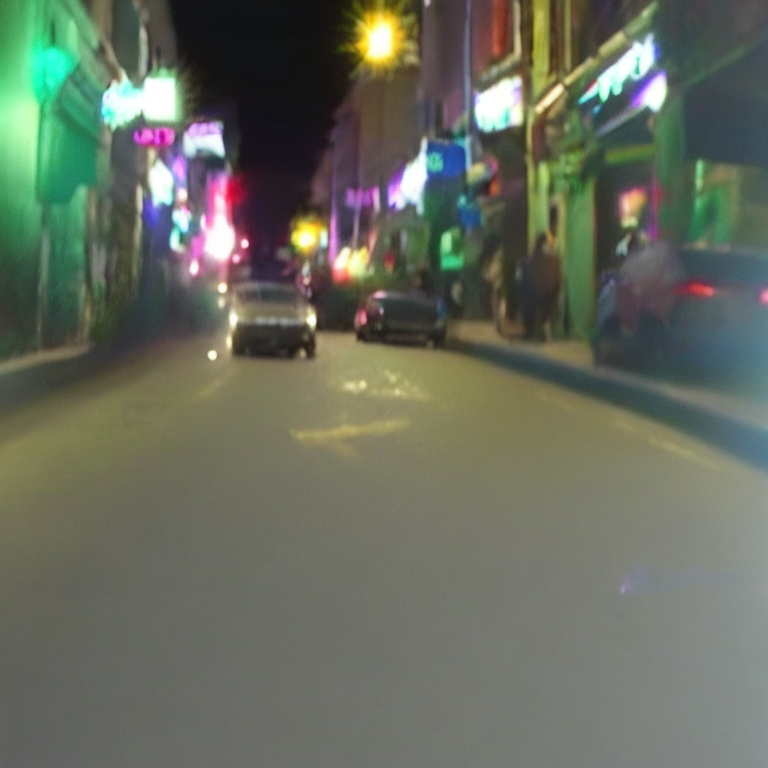} &
\includegraphics[width=0.24\columnwidth]{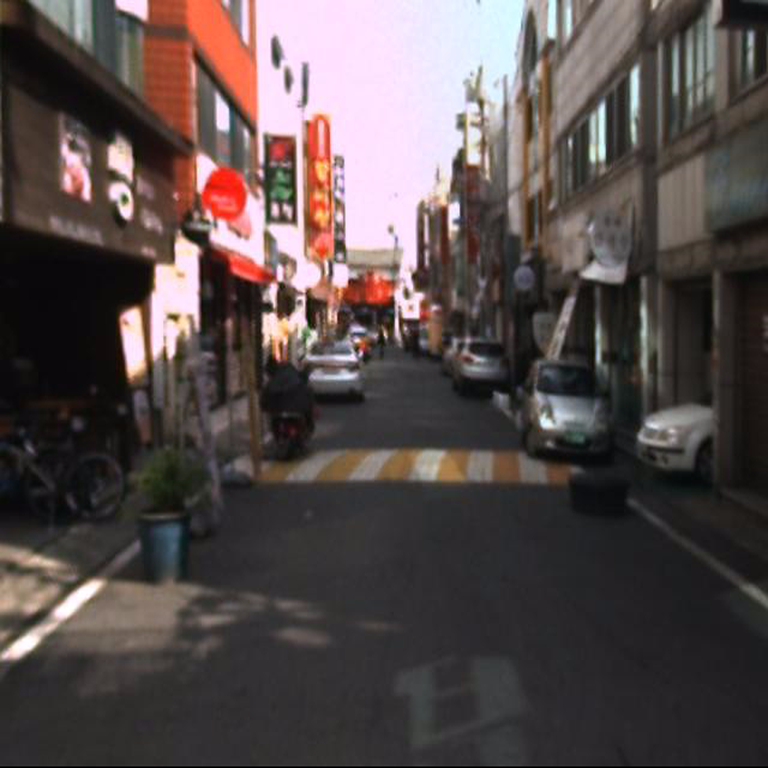} \\
{\footnotesize $\alpha=0.3$} & {\footnotesize $\alpha=0.4$} & {\footnotesize $\alpha=0.5$} & {\footnotesize Paired visible} \\
\end{tabular}
\captionof{figure}{An IR500K/KAIST example: infrared input, pseudo-visible outputs of the diffusion model at six transfer weights, and the paired visible image.}
\label{fig:alpha-sensitivity}
\end{minipage}
\end{center}

Figure~\ref{fig:alpha-sensitivity} makes the sweep visually inspectable on one paired scene. Note that the diffusion model samples a plausible image from an embedding~\cite{ramesh2022unclip}, so the visual differences are qualitative and should not be read as pixelwise fidelity rankings. 

\par Overall, the sweep shows that $ \alpha $ has a broad usable range rather than a single sensitive point; within this range, moderate transfer supports global matching, while dense prediction favors a smaller shift.

\subsection{Feature Geometry Analysis}

\begin{table*}[t]
\centering
\tablefontsize
\caption{Feature separability of infrared object embeddings. Both metrics are computed on the high-dimensional L2-normalized CLIP features before t-SNE projection. $\Delta$ reports the relative change of TMT over OpenAI CLIP, computed from unrounded values.}
\label{tab:separability}
\begin{tabular*}{\textwidth}{@{\extracolsep{\fill}}lcccccc@{}}
\toprule
\multirow{2}{*}{Dataset} & \multicolumn{3}{c}{Margin ratio $\uparrow$} & \multicolumn{3}{c}{Fisher ratio $\uparrow$} \\
\cmidrule(lr){2-4}\cmidrule(lr){5-7}
 & CLIP & TMT & $\Delta$ & CLIP & TMT & $\Delta$ \\
\midrule
M3FD & 0.102 & 0.116 & +13.52\% & 0.132 & 0.175 & +32.86\% \\
SMOD & 0.166 & 0.195 & +17.60\% & 0.124 & 0.168 & +35.95\% \\
FLIRv1 & 0.157 & 0.175 & +11.11\% & 0.117 & 0.152 & +30.08\% \\
FLIRv2 & 0.106 & 0.112 & +5.99\% & 0.149 & 0.180 & +21.08\% \\
LLVIP & 0.526 & 0.775 & +47.49\% & 0.226 & 0.349 & +54.27\% \\
JU-VNT-T & 0.654 & 0.625 & -4.41\% & 1.166 & 1.124 & -3.56\% \\
MFNet & 0.157 & 0.171 & +9.04\% & 0.243 & 0.300 & +23.17\% \\
PST900 & 0.130 & 0.167 & +28.19\% & 0.108 & 0.161 & +49.87\% \\
\midrule
Avg. & 0.250 & 0.292 & +16.94\% & 0.283 & 0.326 & +15.26\% \\
\bottomrule
\end{tabular*}
\end{table*}

\begin{figure*}[!b]
\centering
\setlength{\tabcolsep}{2pt}
\renewcommand{\arraystretch}{0.8}
\tablefontsize
\begin{tabular}{@{}c*{4}{c}@{}}
 & M3FD & FLIRv1 & FLIRv2 & LLVIP \\
\parbox[b][0.195\textwidth][c]{0.9cm}{\centering\rotatebox{90}{\scriptsize OpenAI CLIP}} &
\includegraphics[width=0.195\textwidth]{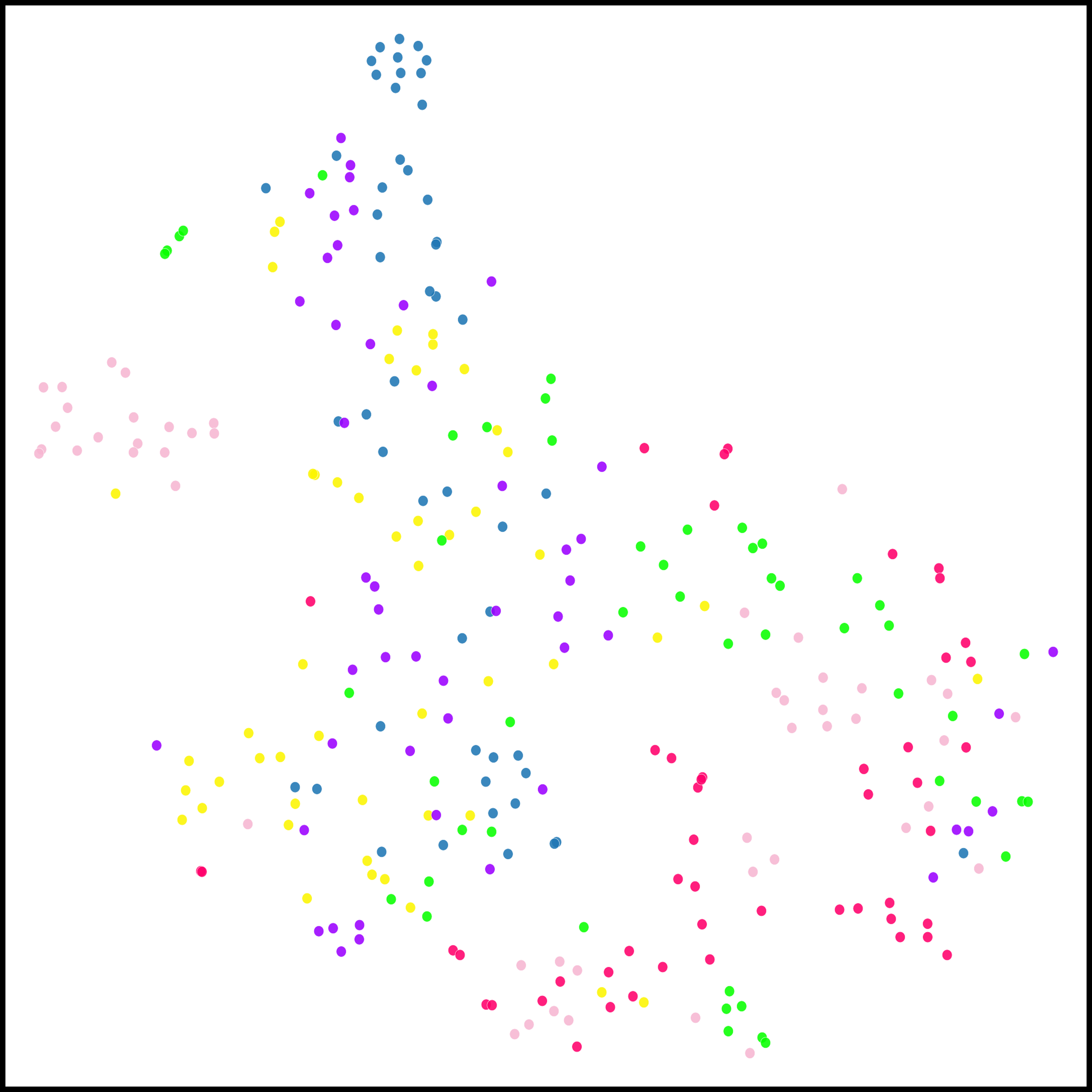} &
\includegraphics[width=0.195\textwidth]{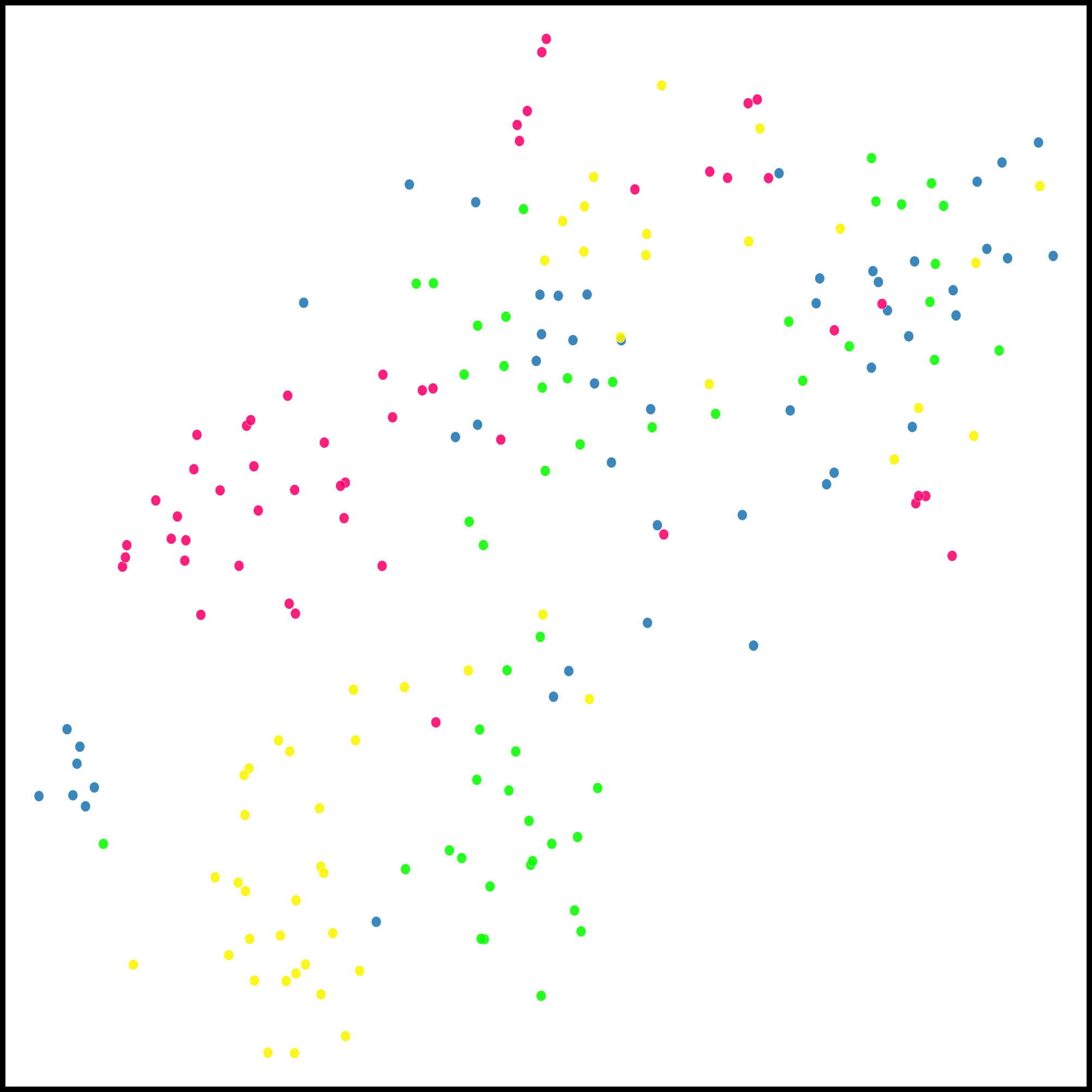} &
\includegraphics[width=0.195\textwidth]{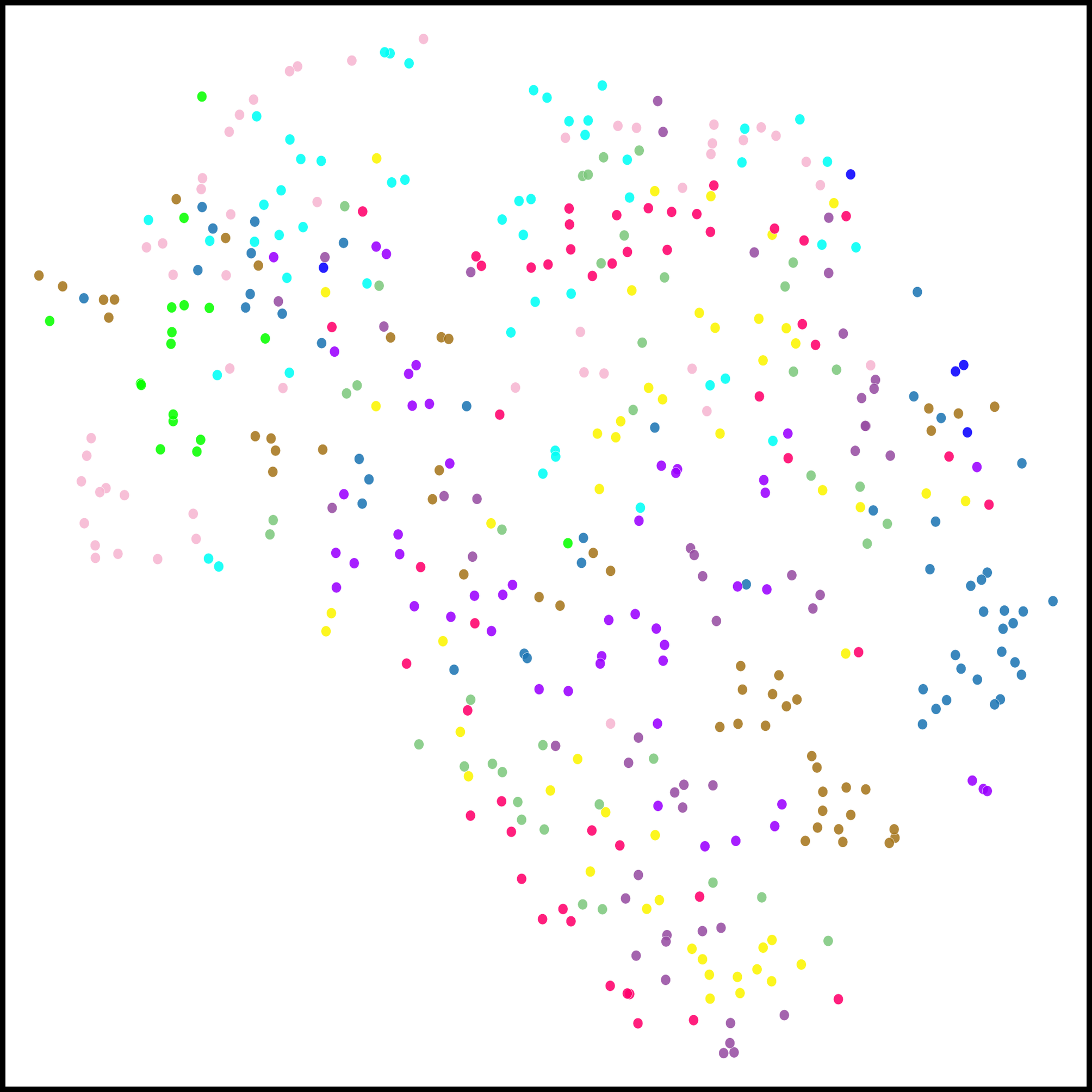} &
\includegraphics[width=0.195\textwidth]{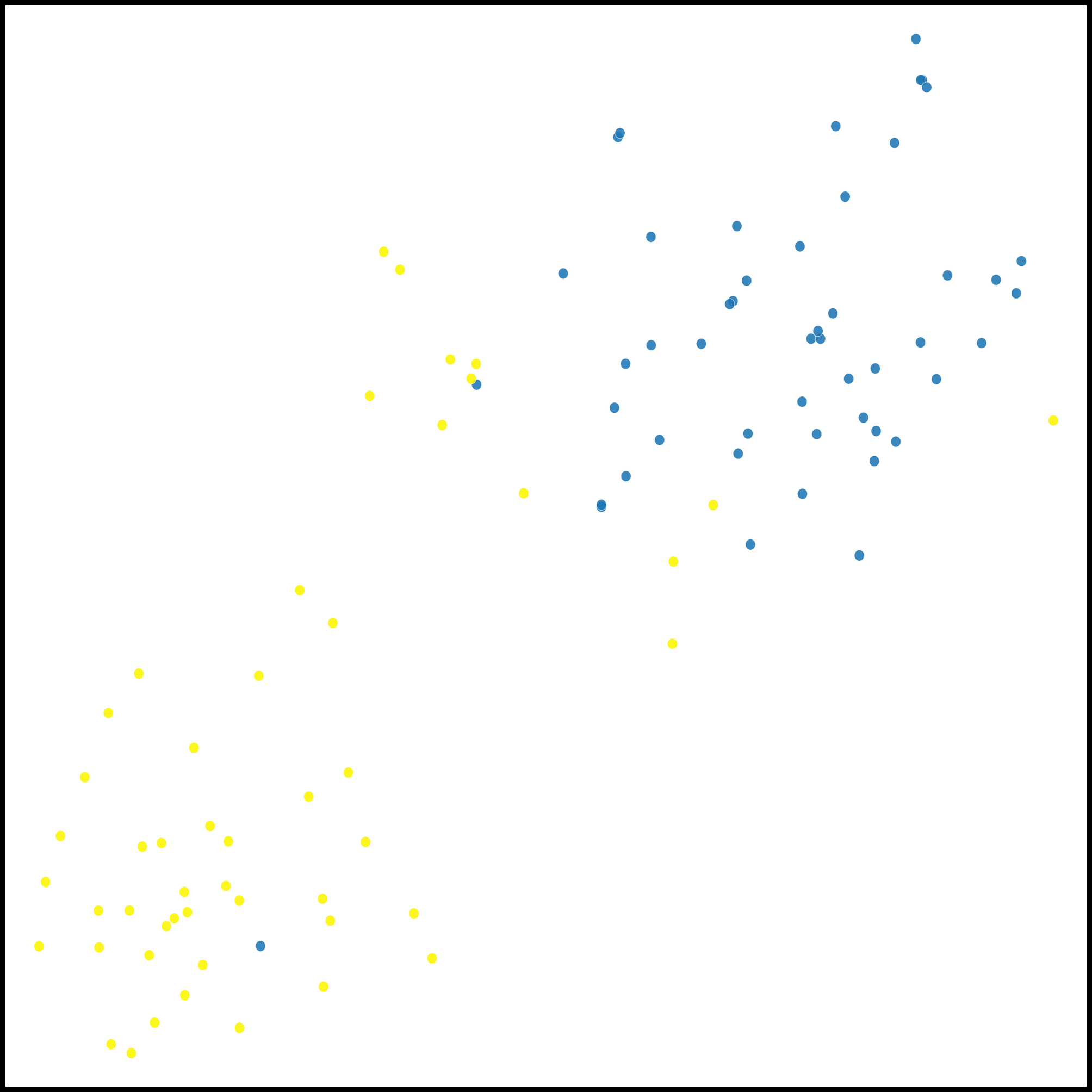} \\
\parbox[b][0.195\textwidth][c]{0.9cm}{\centering\rotatebox{90}{\scriptsize TMT}} &
\includegraphics[width=0.195\textwidth]{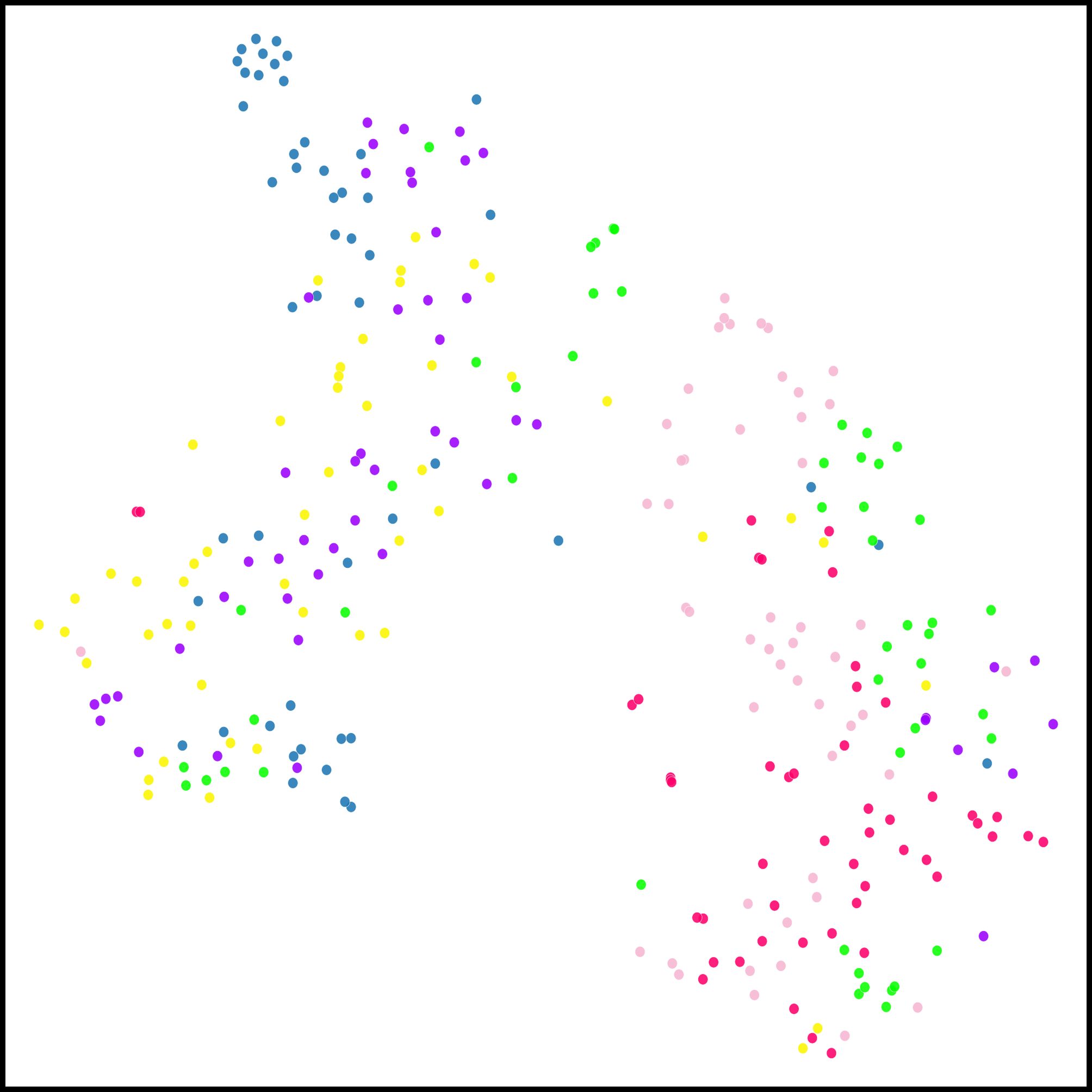} &
\includegraphics[width=0.195\textwidth]{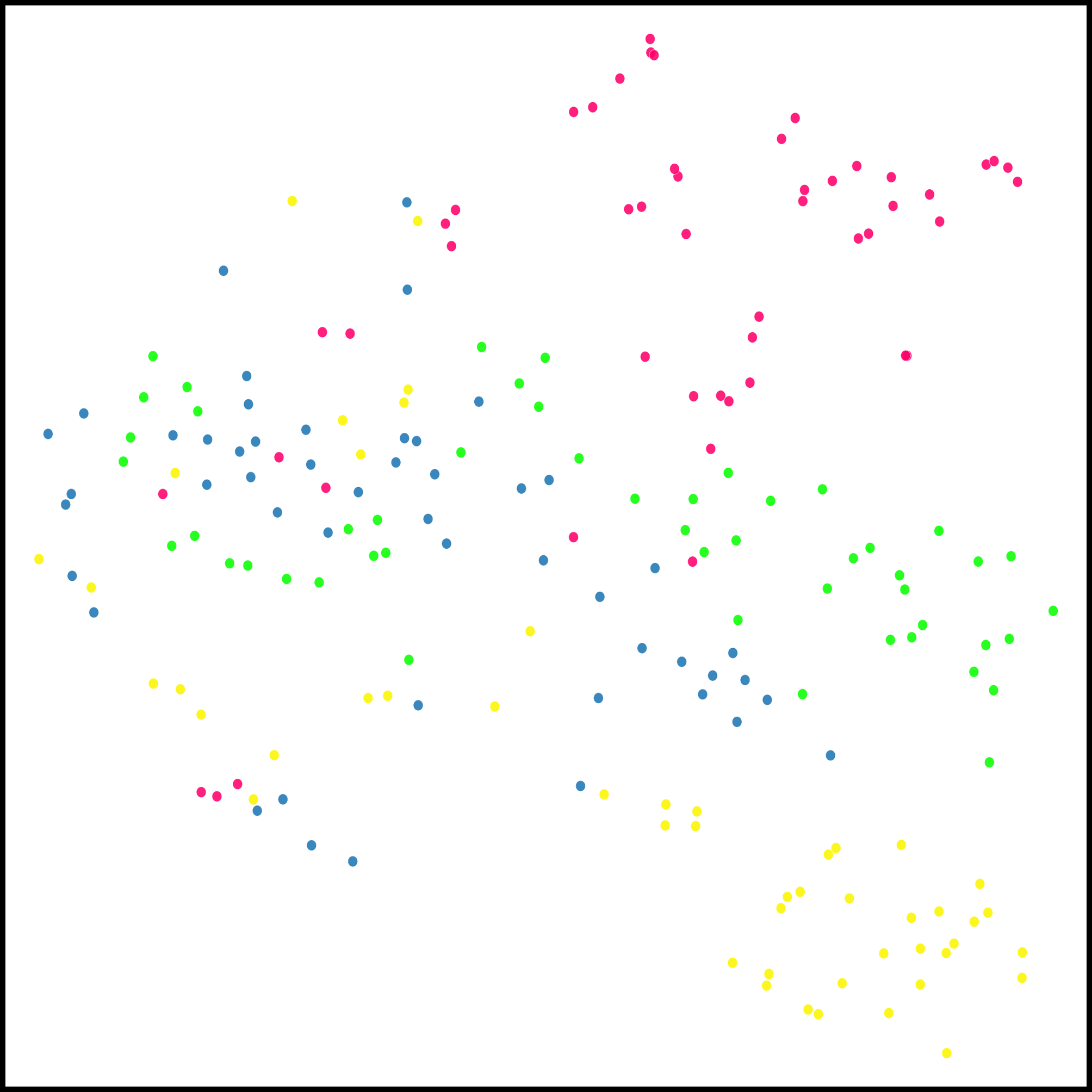} &
\includegraphics[width=0.195\textwidth]{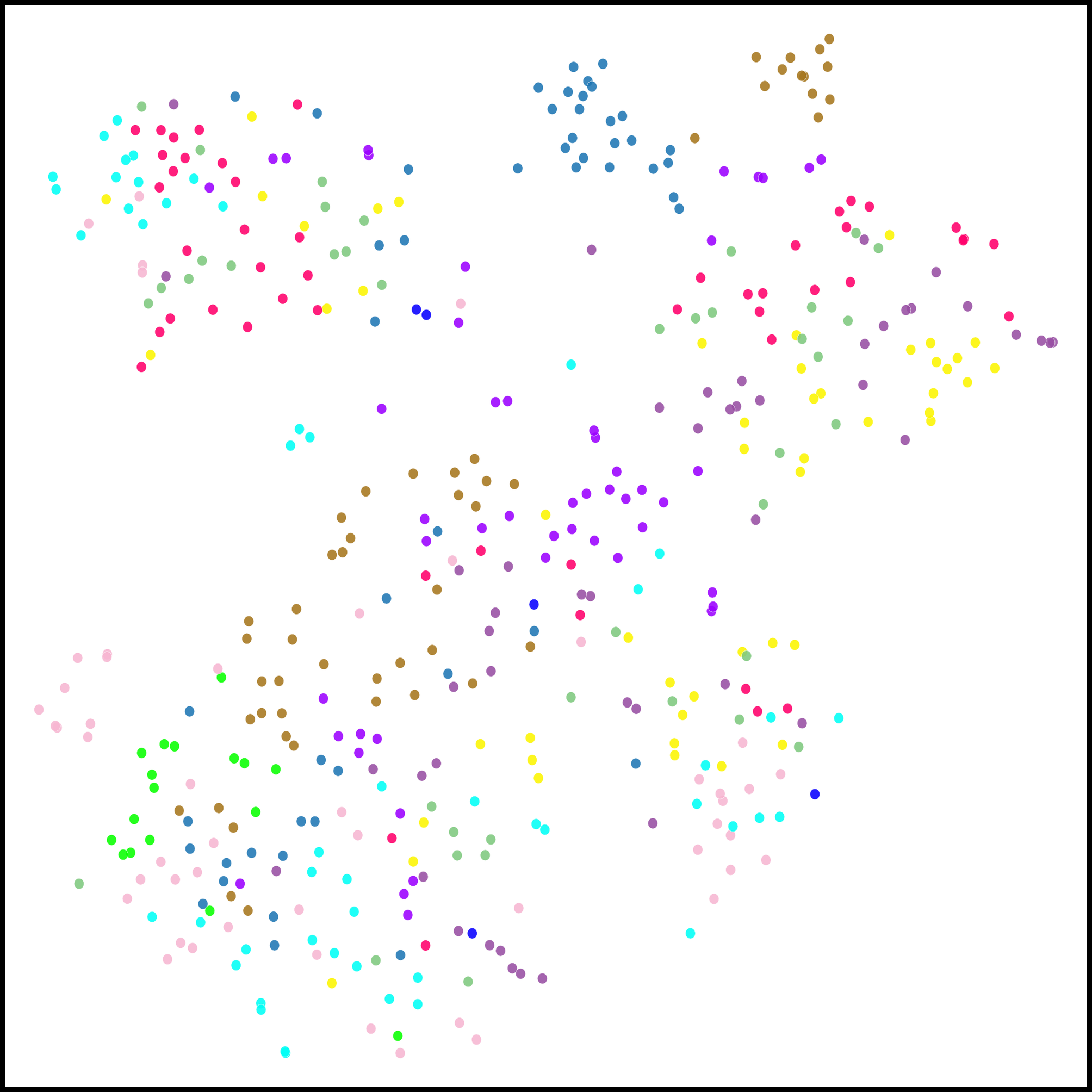} &
\includegraphics[width=0.195\textwidth]{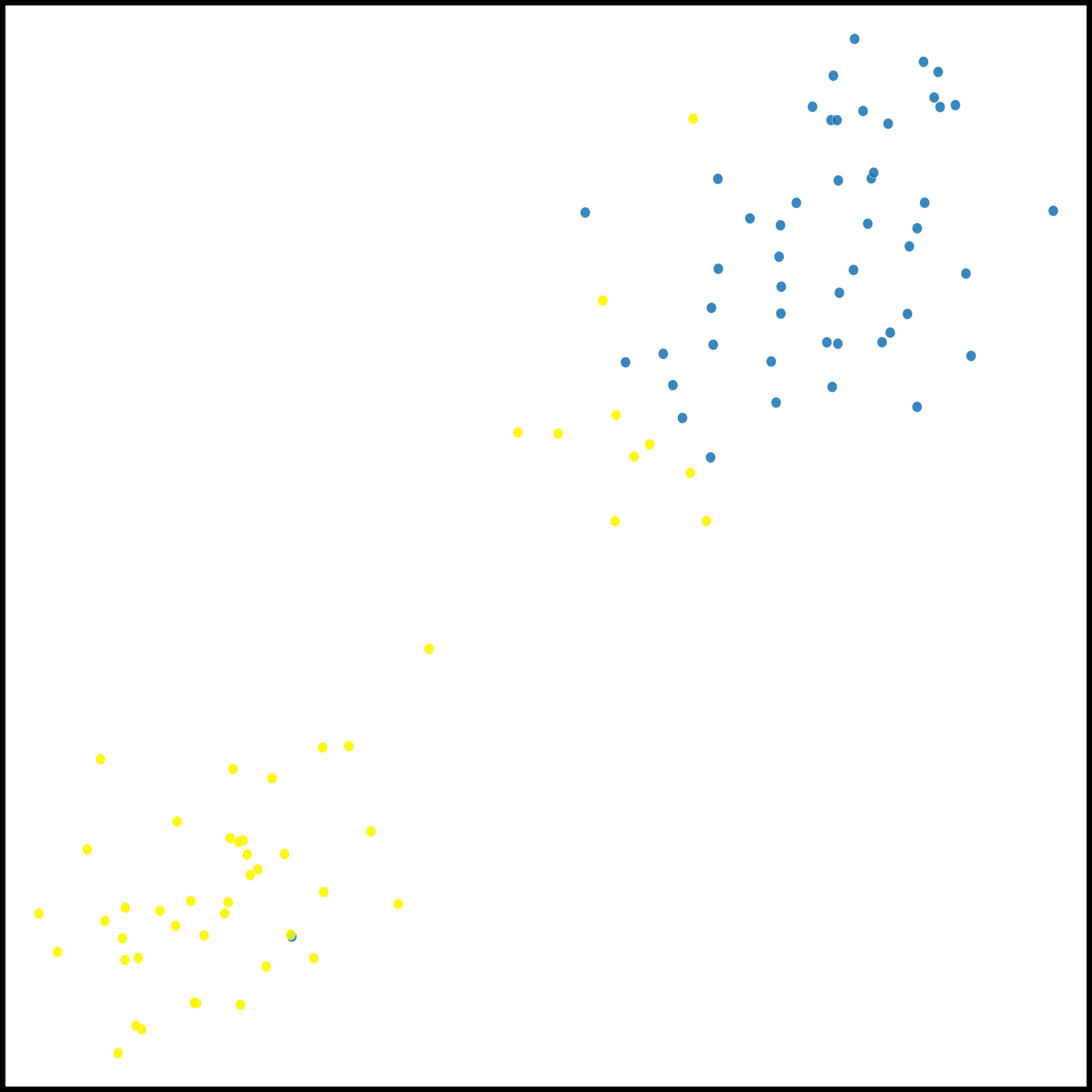}
\end{tabular}

\vspace{1.5mm}

\begin{tabular}{@{}c*{4}{c}@{}}
 & JU-VNT-T & MFNet & PST900 & SMOD \\
\parbox[b][0.195\textwidth][c]{0.9cm}{\centering\rotatebox{90}{\scriptsize OpenAI CLIP}} &
\includegraphics[width=0.195\textwidth]{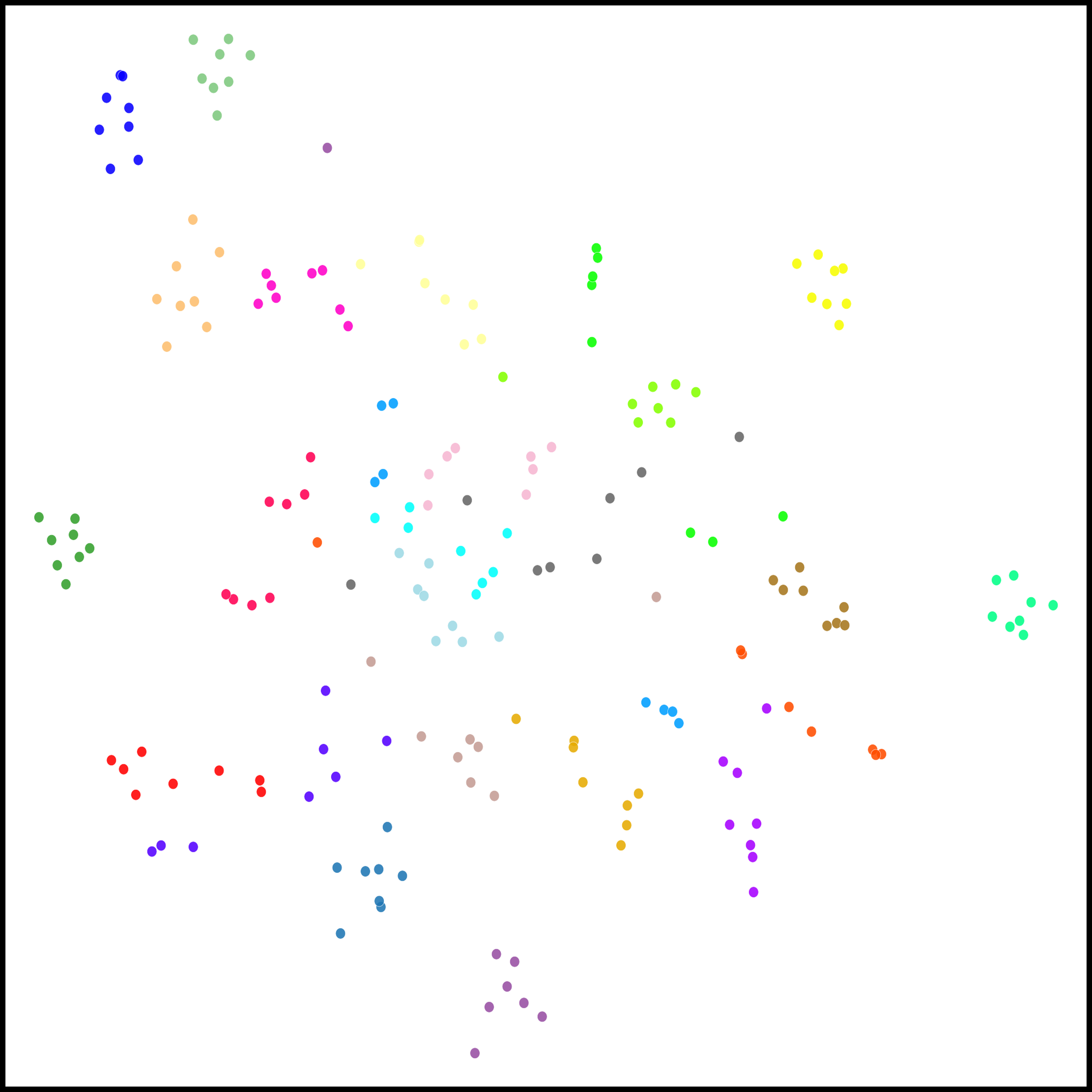} &
\includegraphics[width=0.195\textwidth]{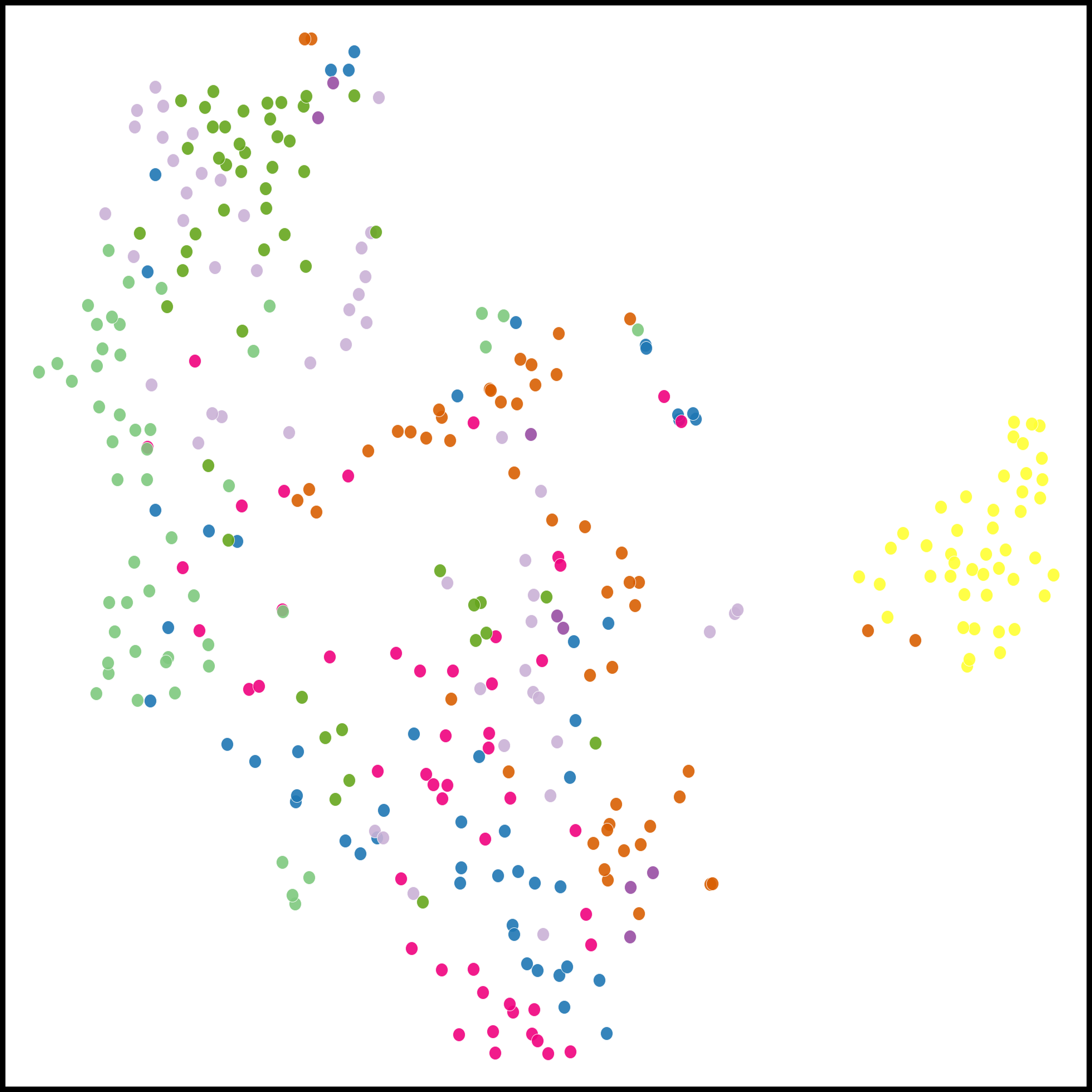} &
\includegraphics[width=0.195\textwidth]{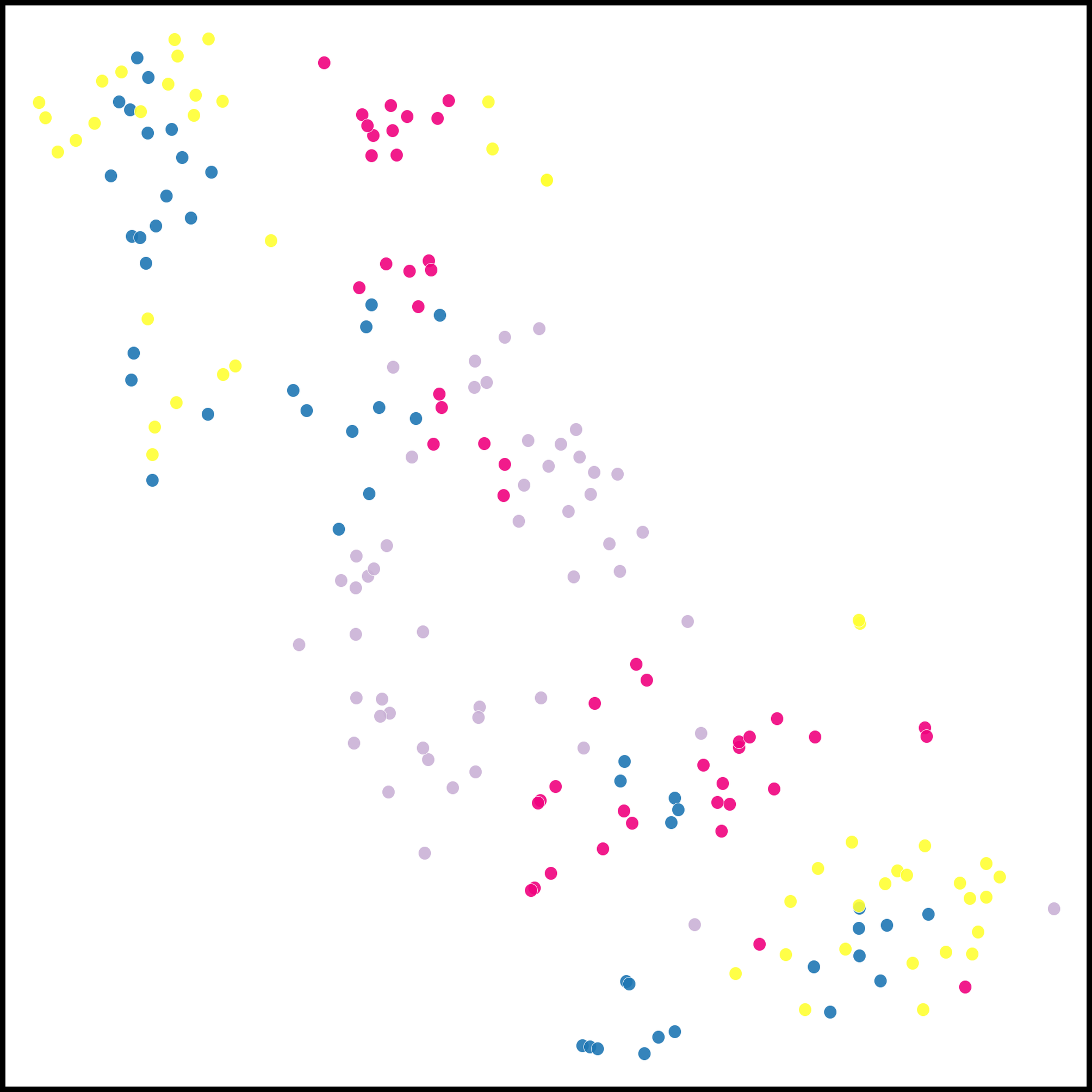} &
\includegraphics[width=0.195\textwidth]{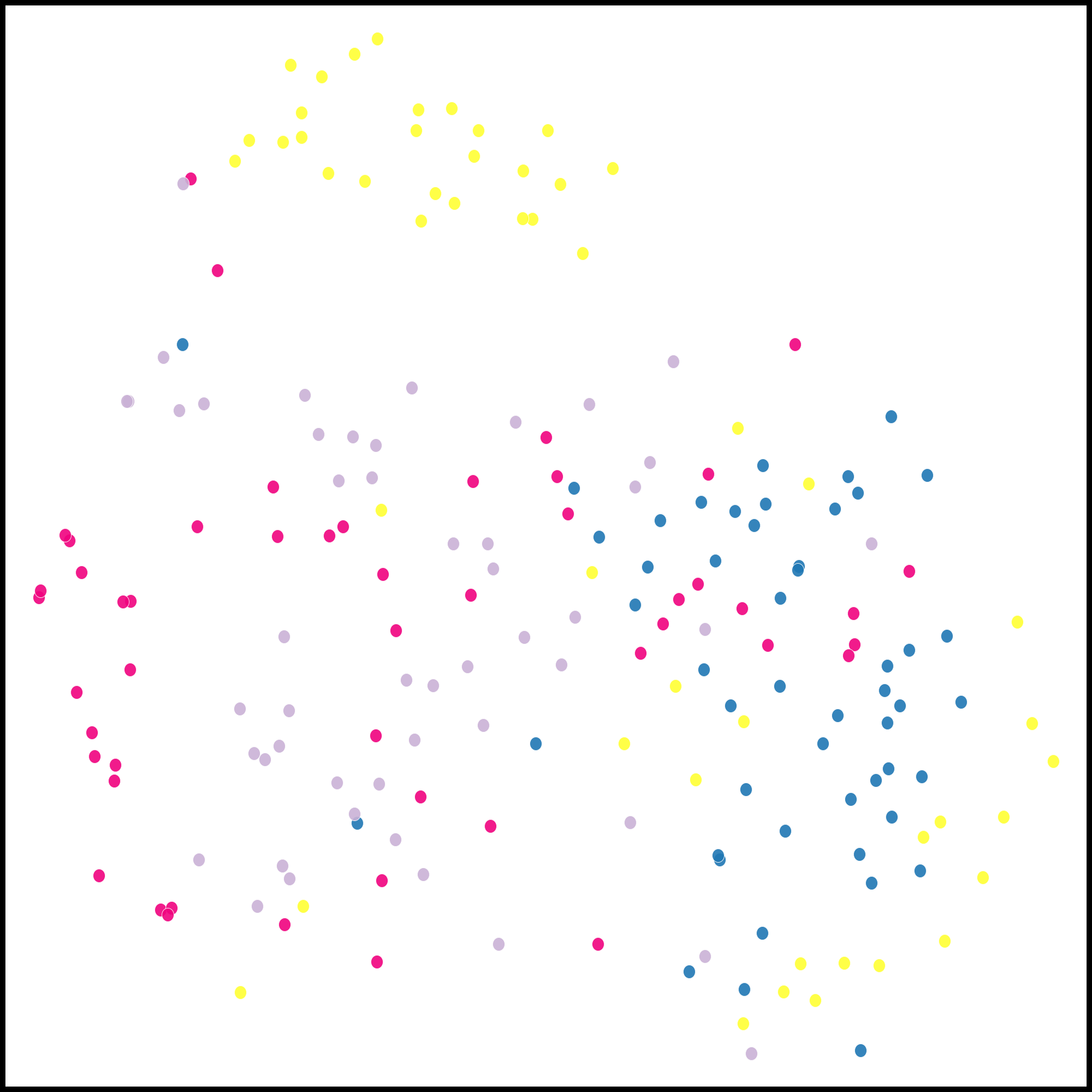} \\
\parbox[b][0.195\textwidth][c]{0.9cm}{\centering\rotatebox{90}{\scriptsize TMT}} &
\includegraphics[width=0.195\textwidth]{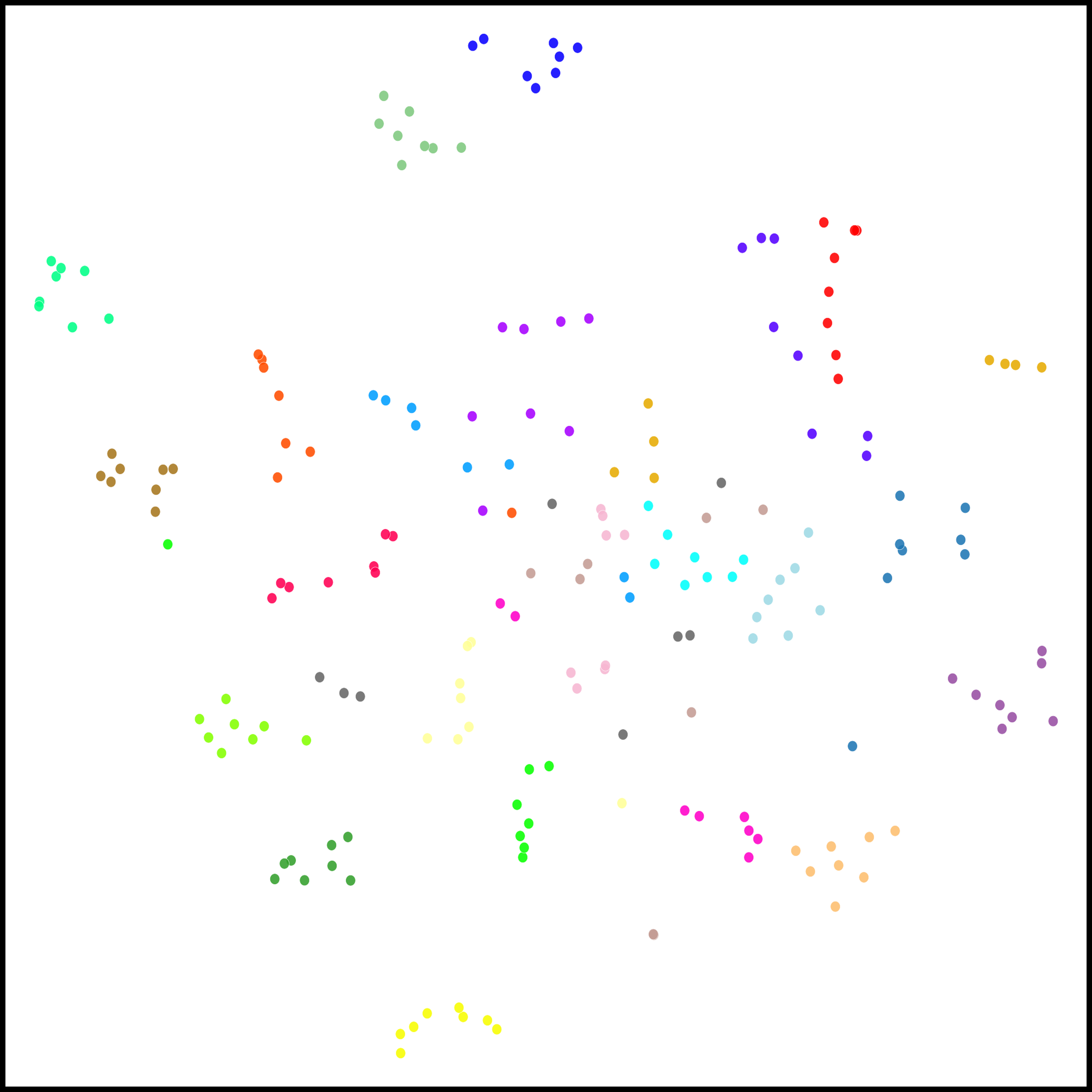} &
\includegraphics[width=0.195\textwidth]{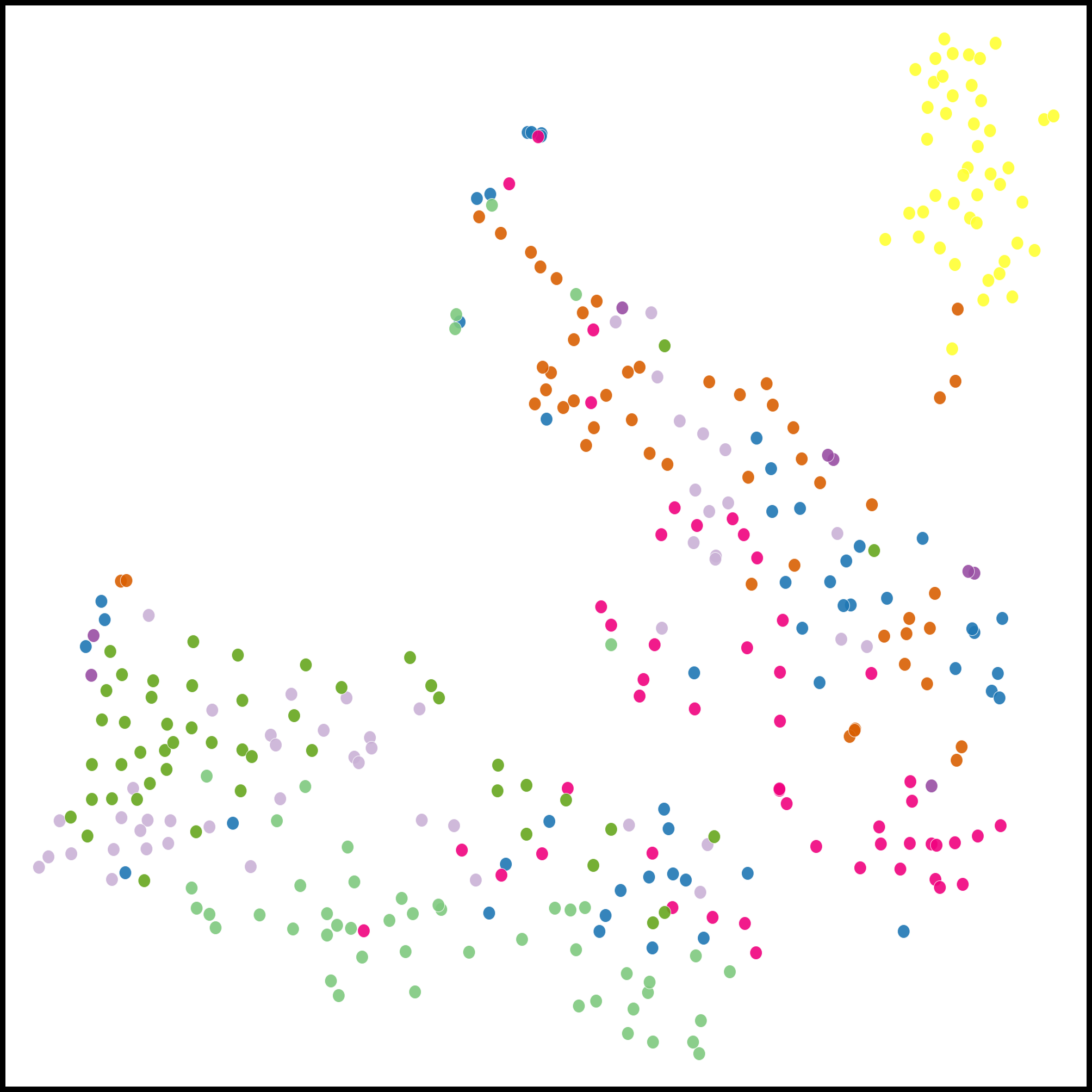} &
\includegraphics[width=0.195\textwidth]{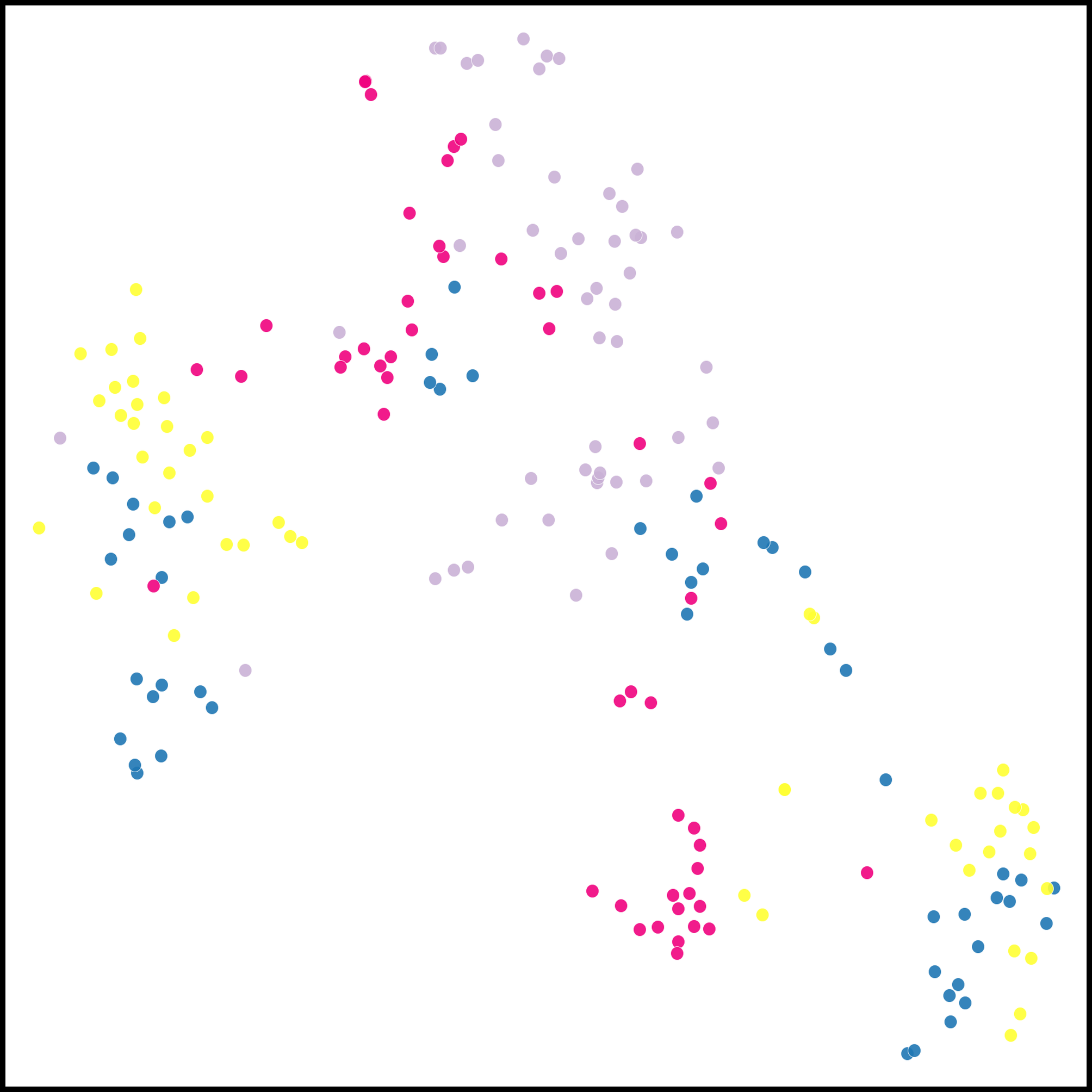} &
\includegraphics[width=0.195\textwidth]{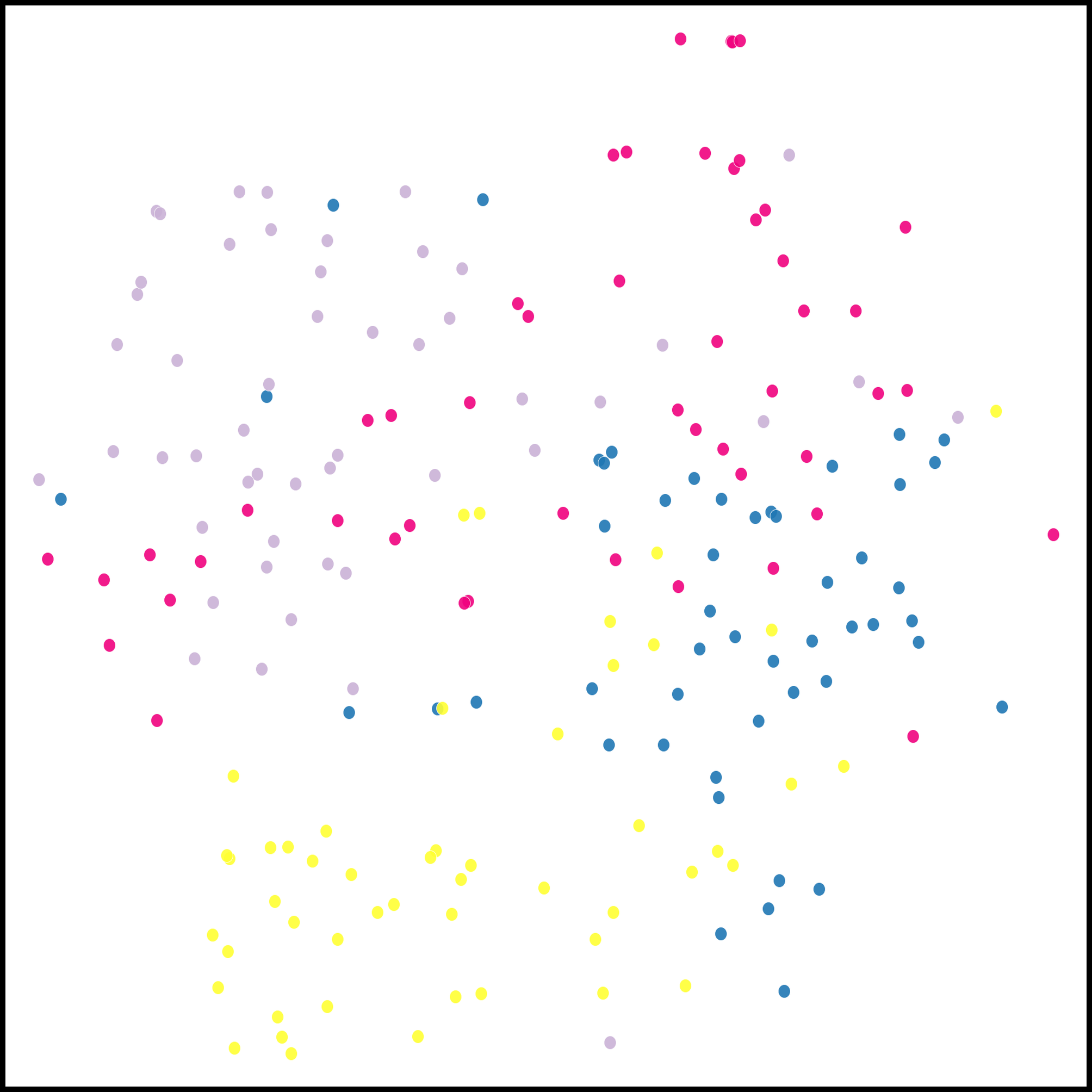}
\end{tabular}
\caption{t-SNE visualization of infrared object features. The upper block shows M3FD, FLIRv1, FLIRv2, and LLVIP; the lower block shows JU-VNT-T, MFNet, PST900, and SMOD. Separability values in Table~\ref{tab:separability} are computed in the original embedding space.}
\label{fig:tsne-separability}
\end{figure*}

To determine whether the classification gains reflect a change in feature geometry rather than a narrow improvement on a particular benchmark, we measure inter-class margins and within-class compactness on the L2-normalized embeddings produced by each model for the evaluation images of every dataset. 
The margin ratio is defined as
\begin{equation}
    \mathrm{MR} = \frac{\frac{1}{C} \sum_{c=1}^{C} \min_{c' \neq c} \|\mu_c - \mu_{c'}\|_2}{\frac{1}{C} \sum_{c=1}^{C} s_c},
\end{equation}
and the Fisher ratio is defined as
\begin{equation}
    \mathrm{FR} = \frac{\sum_{c=1}^{C} N_c \|\mu_c - \mu\|_2^2}{\sum_{c=1}^{C} \sum_{i:y_i=c} \|x_i - \mu_c\|_2^2}.
\end{equation}
Both metrics are computed in the high-dimensional embedding space; t-SNE projections of the same embeddings (Figure~\ref{fig:tsne-separability}) serve only as a qualitative visualization.

Across eight datasets, mean MR rises from 0.250 to 0.292 (16.94\% relative, computed from unrounded values), and mean FR rises from 0.283 to 0.326 (15.26\% relative). Both new source-disjoint datasets improve on both measures: MFNet gains 9.04\% in MR and 23.17\% in FR, while PST900 gains 28.19\% and 49.87\%. This supports transfer of the feature-geometry change beyond the IR500K source families. The exception is JU-VNT-T, whose MR and FR fall by 4.41\% and 3.56\% from already high baseline values. 
Its classification accuracy still improves over the unadapted encoder, so a lower aggregate separability ratio does not imply uniformly worse category matching. 
For qualitative analysis, the plots in Figure~\ref{fig:tsne-separability} provide a complementary visual check of these gains on MFNet and PST900 as well as the earlier benchmarks. Their two-dimensional layout is not a quantitative measure of class separation: the high-dimensional MR and FR values are the relevant evidence, and the JU-VNT-T discrepancy shows why both views are reported. 
These metrics summarize aggregate geometric properties and cannot fully identify which individual classes remain confusable; nevertheless, their improvement on seven of the eight datasets provides converging evidence that the classification gains reflect a genuine change in feature geometry. 

\section{Conclusion}

We present TMT, a text-derived modality-transfer framework that adapts CLIP to infrared imagery with paired semantic alignment and diffusion noise-prediction supervision.
Training uses paired infrared--visible images and generic modality descriptions without instance-level infrared--text annotations.
Text-Derived Modality Transfer derives a reusable infrared-to-visible modality direction from generic modality descriptions in CLIP's joint embedding space and applies it to the infrared image representation.
Generative--Perceptual Co-Supervision then constrains the transferred representation from two complementary perspectives: a frozen CLIP image encoder preserves correspondence with the paired visible image, while a frozen diffusion model provides feedback related to visible-image detail. 

At inference, the TMT-adapted image encoder and frozen CLIP text encoder perform image--text matching; the modality direction and auxiliary supervision models are used only during training.
Across four source-disjoint classification benchmarks, TMT reaches 48.92\% mean zero-shot accuracy, leads outright on two, ties on SMOD at two-decimal precision, and exceeds LanguageBind by 3.98 points under our protocol. The four source-overlapping classification benchmarks provide supplementary evaluation. On IR-TD retrieval, TMT is the strongest among the methods that do not use infrared--text supervision during adaptation. The segmentation results show gains on both evaluated datasets. Ablations and feature-geometry analyzes provide complementary evidence for improved semantic organization, and a linear-probe analysis shows that the diffusion branch increases the linear readability of infrared texture detail while suppressing global brightness variation. 

\paragraph{Limitations and Future Work.}
We evaluate TMT only with a CLIP image encoder. Future work will investigate thermodynamic priors for infrared imaging, test whether the adaptation strategy extends to other visual foundation models, and examine how synthesized infrared captions~\cite{qazi2026tclip} can complement the present caption-free supervision.

\appendix
\counterwithin{equation}{section}
\counterwithin{figure}{section}
\counterwithin{table}{section}
\section{Analysis of the Diffusion Noise-Prediction Supervision} 
\label{app:noise-free}

\subsection{Motivation}
The empirical comparison below is conducted \emph{solely with the generative branch}
(i.e., the diffusion noise-prediction loss $\mathcal{L}_{\text{diff}}$)
in order to isolate the effect of the training strategy.
The paired semantic alignment branch ($\mathcal{L}_{\text{semantic}}$) is omitted so that
any observed difference in stability can be attributed unambiguously to the diffusion-supervision design.

For diffusion noise-prediction supervision,
two strategies exist for using the frozen diffusion model as a supervisory signal.
\textbf{Strategy~A} (noise-free) feeds the \emph{clean} visible image latent directly into the diffusion model,
which predicts noise conditioned on $\mathbf{v}_{\text{pseudo vis}}$;
the prediction is compared against a sampled Gaussian noise target.
\textbf{Strategy~B} (DDPM-style) follows the standard forward-noising pipeline:
the visible image is first corrupted with noise at timestep $t$ to obtain $\mathbf{x}_t$,
then the diffusion model predicts the noise from $\mathbf{x}_t$.
Empirically, Strategy~A improves CLIP's infrared-image performance,
whereas Strategy~B causes training instability and eventual collapse.
Figure~\ref{fig:training_curves} compares the training curves.

\begin{center}
\begin{minipage}{\columnwidth}
    \centering
    \includegraphics[width=0.48\columnwidth]{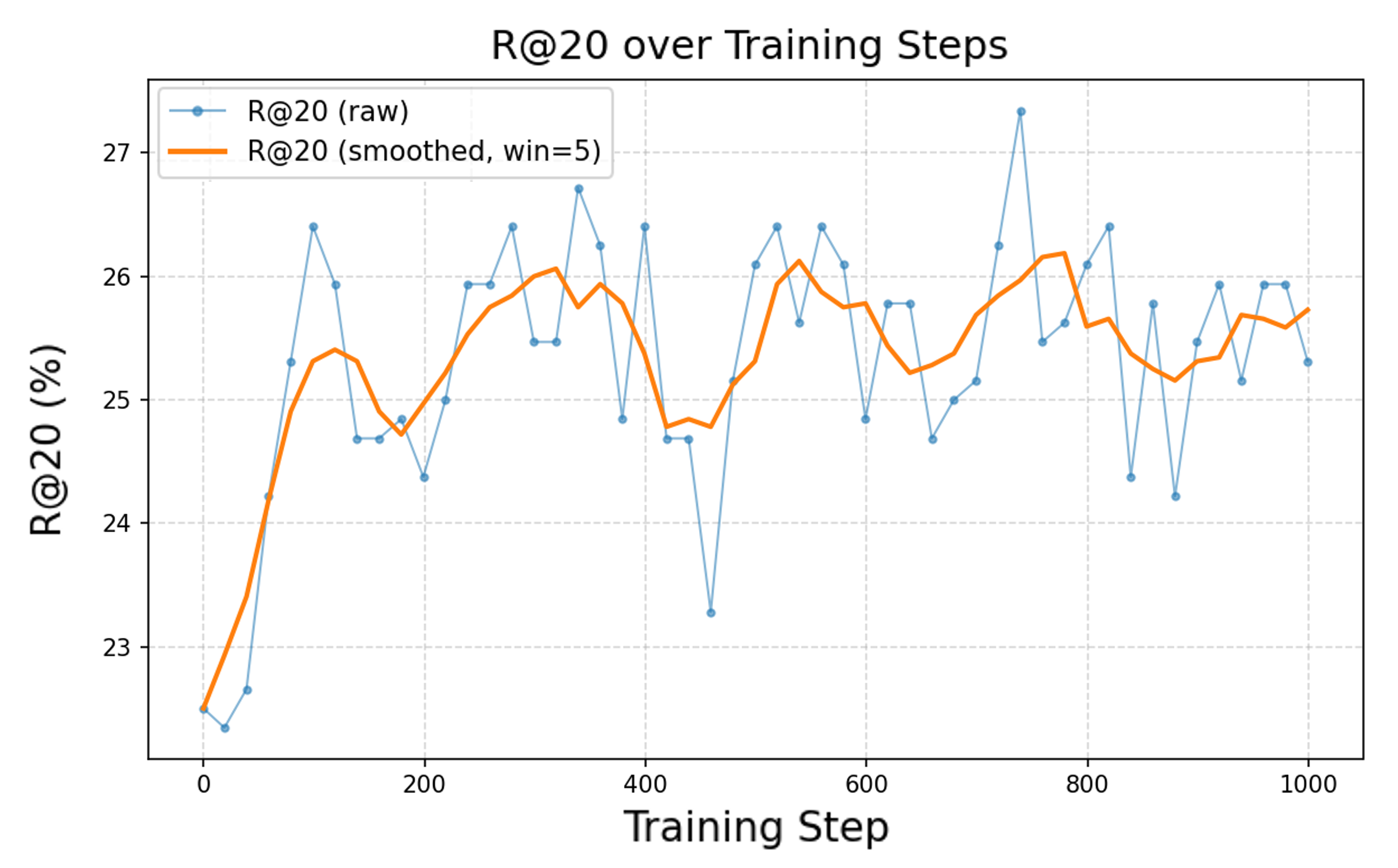}
    \hfill
    \includegraphics[width=0.48\columnwidth]{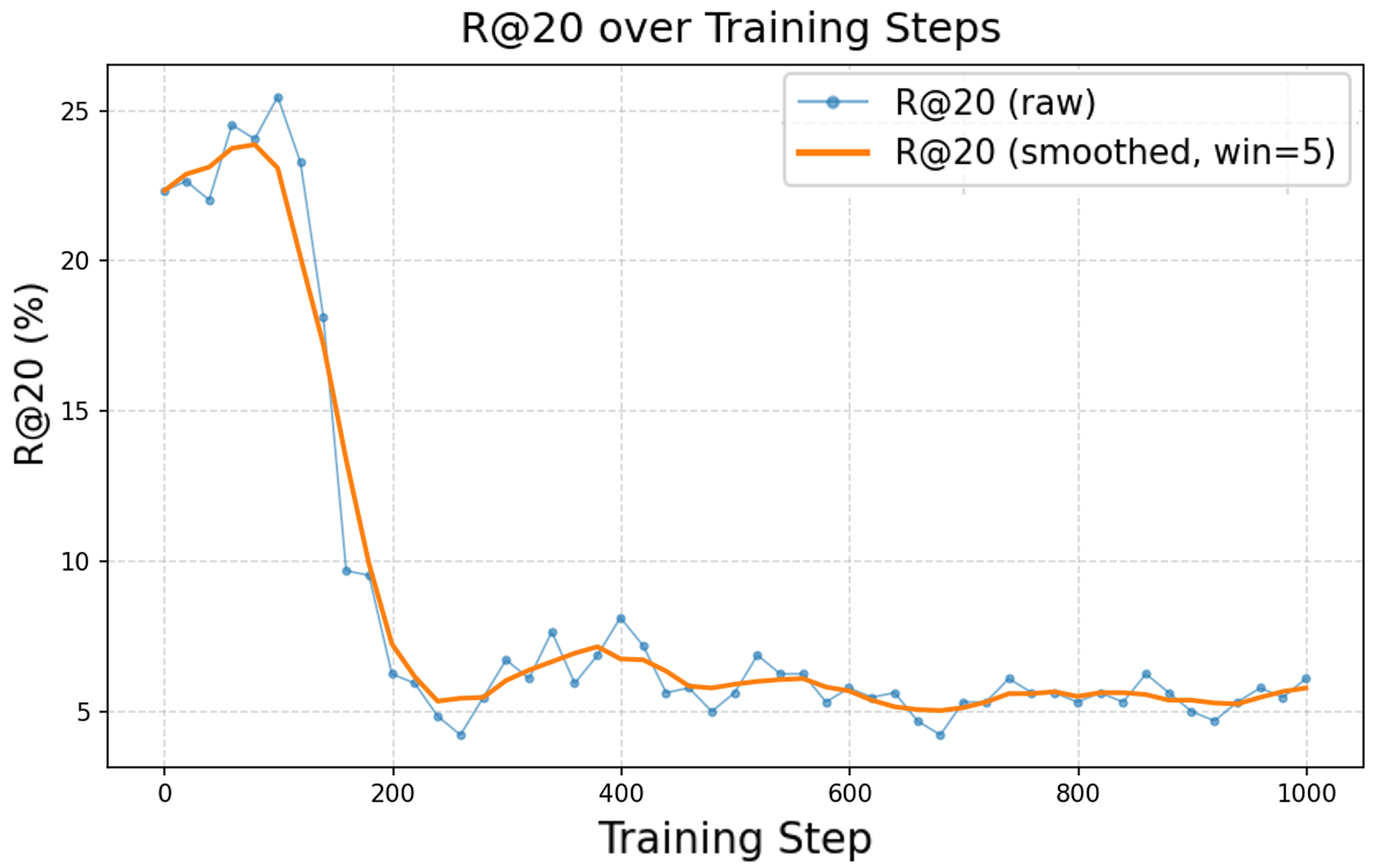}
    \\[0.5em]
    {\small (left) Strategy~A (noise-free) \quad (right) Strategy~B (DDPM-style)}
    \captionof{figure}{Training curves of the two diffusion supervision strategies.}
    \label{fig:training_curves}
\end{minipage}
\end{center}

\subsection{Theoretical Analysis}

Let $f(\mathbf{x}, t, \mathbf{c})$ denote the frozen diffusion model,
where $\mathbf{x}$ is the diffusion input,
$t$ is the timestep,
and $\mathbf{c}$ is the condition vector (here, $\mathbf{c} = \mathbf{v}_{\text{pseudo vis}}$).
The standard diffusion training objective is:
\begin{equation}
    \mathcal{L} = \mathbb{E}\left[\left\| f(\mathbf{x}, t, \mathbf{c}) - \boldsymbol{\varepsilon} \right\|_2^2\right],
\end{equation}
where $\boldsymbol{\varepsilon} \sim \mathcal{N}(\mathbf{0}, \mathbf{I})$ is the Gaussian noise target.
The gradient w.r.t.~the condition vector is:
\begin{equation}
    \begin{aligned}
    \nabla_{\mathbf{c}} \mathcal{L}
    &= \nabla_{\mathbf{c}} \mathbb{E}\left[\left\| f(\mathbf{x}, t, \mathbf{c}) - \boldsymbol{\varepsilon} \right\|_2^2\right] \\
    &= \mathbb{E}\left[ \nabla_{\mathbf{c}} \left\| f(\mathbf{x}, t, \mathbf{c}) - \boldsymbol{\varepsilon} \right\|_2^2 \right] \\
    &= \mathbb{E}\left[ 2\left(f(\mathbf{x}, t, \mathbf{c}) - \boldsymbol{\varepsilon}\right)^{\!\top} \nabla_{\mathbf{c}} f(\mathbf{x}, t, \mathbf{c}) \right].
    \end{aligned}
\end{equation}

\paragraph{Strategy~A (noise-free).}
Since the input $\mathbf{x}$ is computed from the clean visible image without adding noise,
$\mathbf{x}$ and $\boldsymbol{\varepsilon}$ are statistically independent.
For fixed $(\mathbf{x}, t, \mathbf{c})$,
the term $\nabla_{\mathbf{c}} f(\mathbf{x}, t, \mathbf{c})$ is determined by the clean latent, timestep, and condition vector, and is therefore independent of the sampled noise $\boldsymbol{\varepsilon}$.
This independence allows the gradient term to be taken outside the expectation over $\boldsymbol{\varepsilon}$.
Writing the factorization with an explicit conditional expectation gives:
\begin{equation}
    \begin{aligned}
    \nabla_{\mathbf{c}} \mathcal{L}
    &= \mathbb{E}_{\mathbf{x},t}\!\left[
    \mathbb{E}_{\boldsymbol{\varepsilon}}\!\left[
    2\left(f(\mathbf{x}, t, \mathbf{c}) - \boldsymbol{\varepsilon}\right)^{\!\top}
    \nabla_{\mathbf{c}} f(\mathbf{x}, t, \mathbf{c})
    \,\middle|\, \mathbf{x}, t
    \right]\right] \\
    &= \mathbb{E}_{\mathbf{x},t}\!\left[
    2\left(f(\mathbf{x}, t, \mathbf{c})^{\!\top}
    - \mathbb{E}_{\boldsymbol{\varepsilon}}[\boldsymbol{\varepsilon}]^{\!\top}\right)
    \nabla_{\mathbf{c}} f(\mathbf{x}, t, \mathbf{c})
    \right] \\
    &= \mathbb{E}_{\mathbf{x},t}\!\left[
    2 f(\mathbf{x}, t, \mathbf{c})^{\!\top}
    \nabla_{\mathbf{c}} f(\mathbf{x}, t, \mathbf{c})
    \right],
    \end{aligned}
\end{equation}
where the last step uses $\mathbb{E}[\boldsymbol{\varepsilon}] = \mathbf{0}$.
Thus, Strategy~A drives the \emph{expected} diffusion output toward zero (the mean of the noise distribution).
This derivation does not make Strategy~A equivalent to the full DDPM objective.
It only shows that removing the dependence between the diffusion input and the sampled noise eliminates the noise-coupled gradient term, yielding a lower-variance supervisory signal in expectation.

\paragraph{Strategy~B (DDPM-style).}
When the input is the noised latent $\mathbf{x}_t$,
$\mathbf{x}_t$ and $\boldsymbol{\varepsilon}$ are correlated (both depend on the same noise sample),
so the expectation does \emph{not} factorize.
The gradient contains noise-coupled terms that can increase the variance and magnitude of backpropagated updates,
which is consistent with the observed training instability.

\subsection{Empirical Validation}

\subsubsection{Gradient Statistics}

To verify the theoretical prediction that Strategy~A yields smaller gradients,
we visualize the gradient distributions of the trainable CLIP image encoder under both strategies.
As shown in Figure~\ref{fig:gradients},
Strategy~A produces gradients with substantially smaller magnitudes in this diagnostic,
supporting the interpretation that the noise-free approach is a milder form of supervision.

\begin{center}
\begin{minipage}{\columnwidth}
    \centering
    \includegraphics[width=0.48\columnwidth]{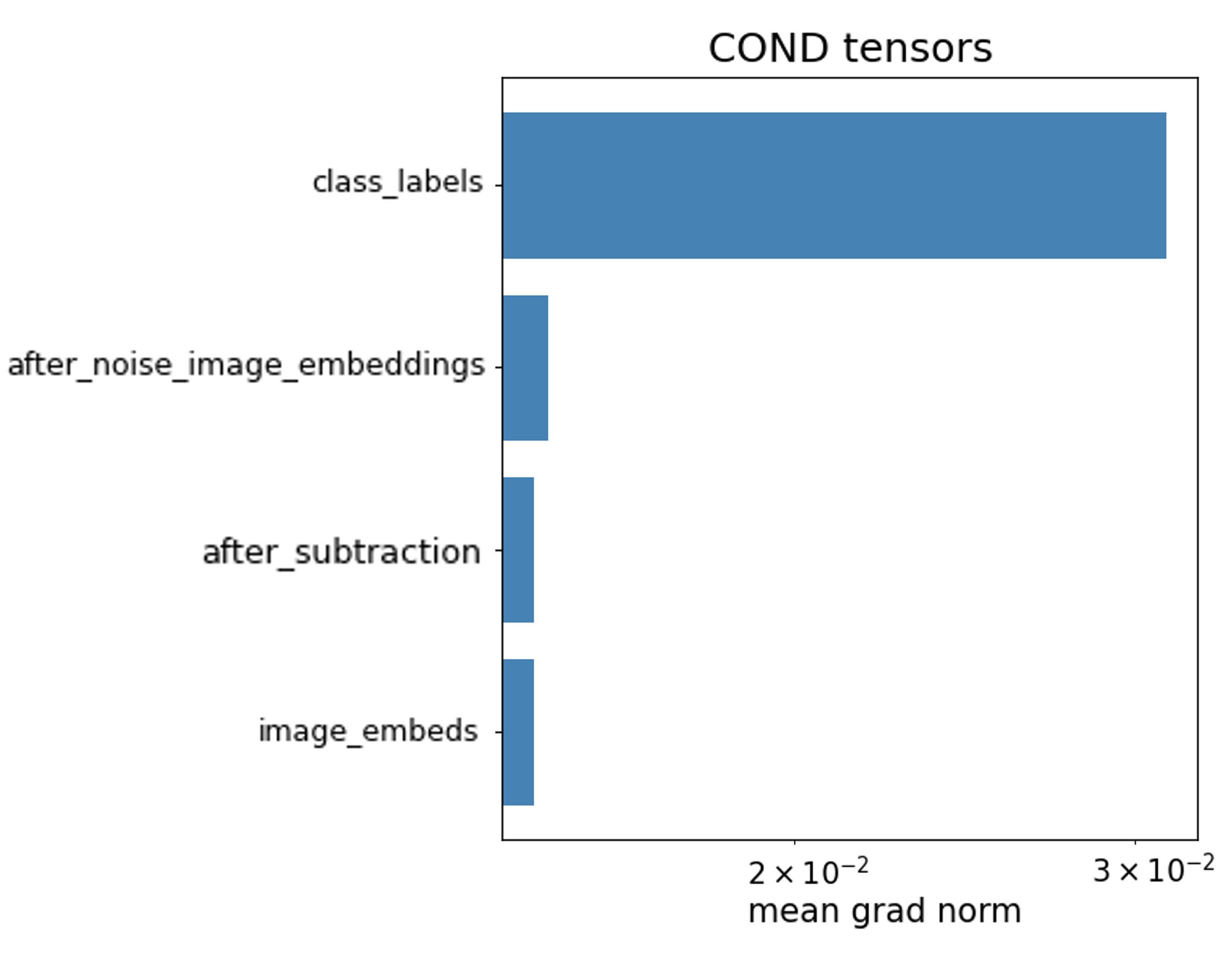}
    \hfill
    \includegraphics[width=0.48\columnwidth]{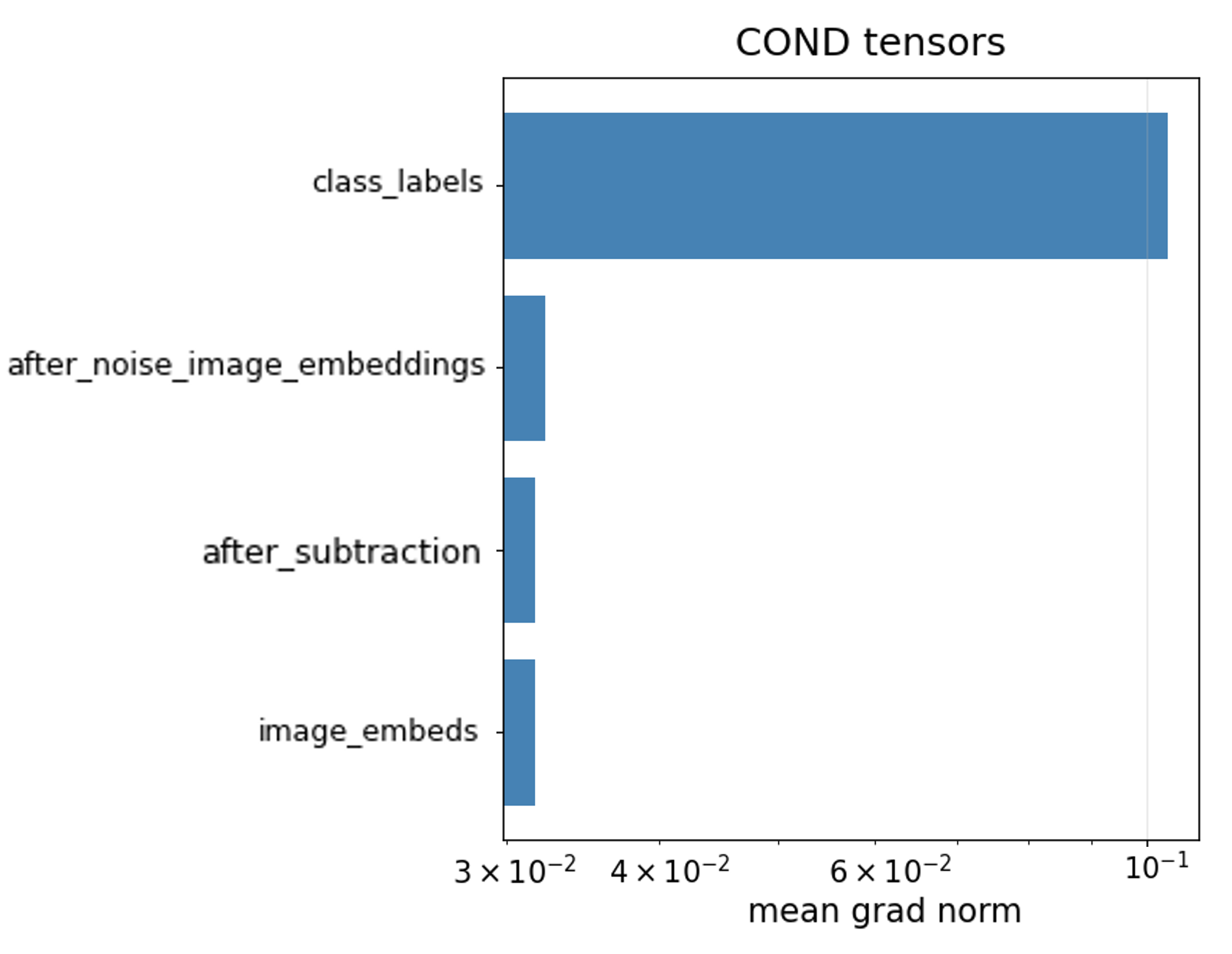}
    \\[0.5em]
    {\small (left) Strategy~A (noise-free) \quad (right) Strategy~B (DDPM-style)}
    \captionof{figure}{Gradient distributions in the trainable CLIP image encoder.}
    \label{fig:gradients}
\end{minipage}
\end{center}

\subsubsection{Learning Rate Sensitivity}

We further investigate whether reducing the learning rate can stabilize Strategy~B.
Figure~\ref{fig:lr_sensitivity} shows the training behavior under learning rates $10^{-4}$, $10^{-5}$, and $10^{-6}$.
While lowering the learning rate does suppress the collapse phenomenon,
the training signal becomes too weak to produce meaningful improvements.
This trade-off suggests that the instability of Strategy~B is closely tied to its larger gradient magnitudes rather than only to an extraneous hyperparameter issue.

\begin{center}
\begin{minipage}{\columnwidth}
    \centering
    \includegraphics[width=0.32\columnwidth]{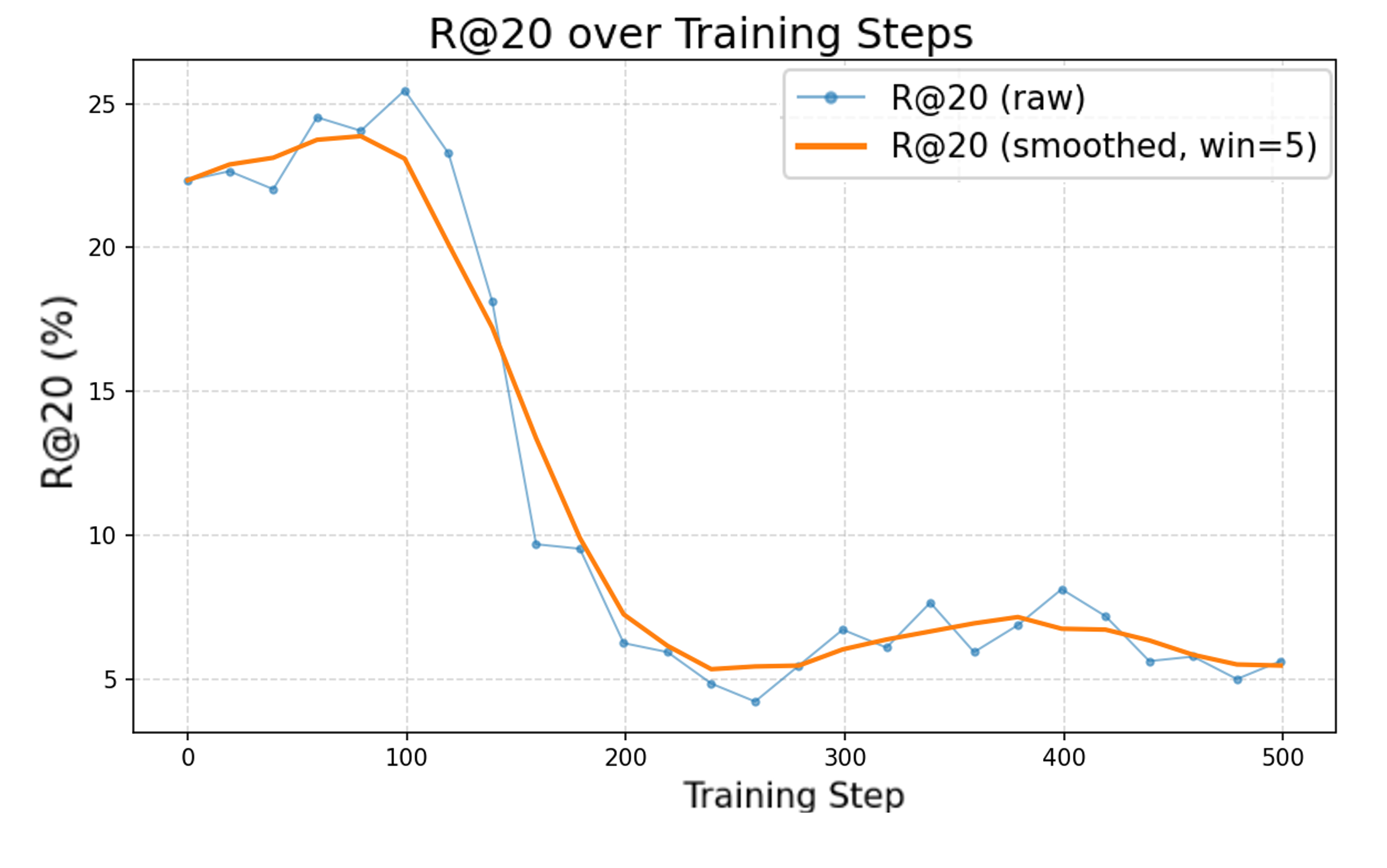}
    \hfill
    \includegraphics[width=0.32\columnwidth]{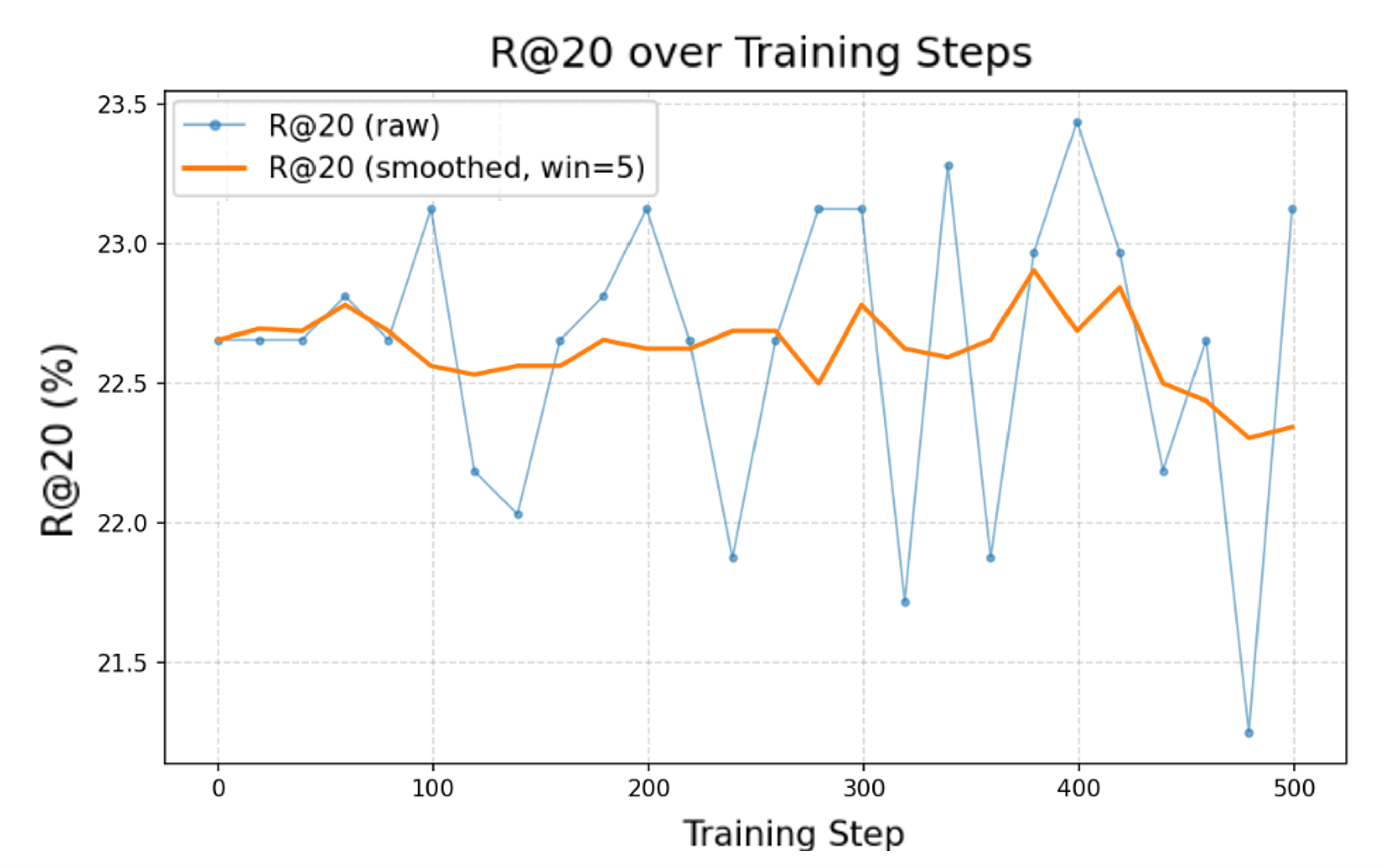}
    \hfill
    \includegraphics[width=0.32\columnwidth]{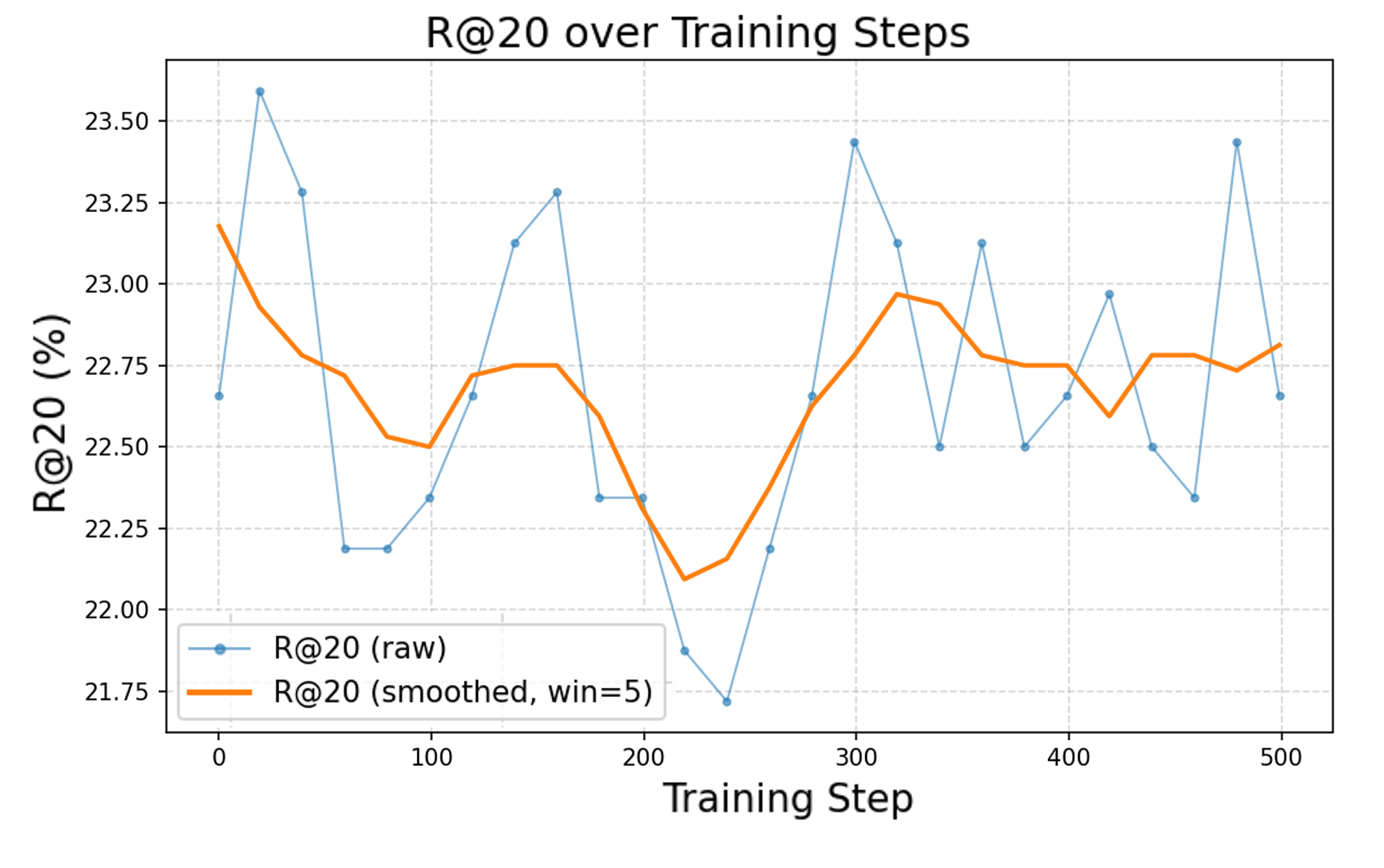}
    \\[0.5em]
    {\small $10^{-4}$ \hfill $10^{-5}$ \hfill $10^{-6}$}
    \captionof{figure}{Training curves of Strategy~B under different learning rates.}
    \label{fig:lr_sensitivity}
\end{minipage}
\end{center}

Based on the theoretical analysis and empirical evidence above,
we adopt Strategy~A (noise-free direct noise prediction) as the diffusion noise-prediction supervision in our framework.

\subsection{Linear-Probe Analysis of the Diffusion Branch}
\label{app:diffusion-probe}

The analyses above justify how the diffusion supervision is applied; a separate question is what the branch changes in the representation. Concretely, we ask whether training with $\mathcal{L}_{\text{diff}}$ makes texture information more linearly readable from the CLIP embedding, and whether such a change is specific to detail or merely reflects a generic sensitivity to any pixel-level variation. To separate the two cases, we probe two controlled perturbation families: a \emph{texture} family that scales high-frequency energy and therefore acts as a detail metric, and a \emph{gain} family that only changes global brightness and contrast without moving edges or altering texture structure and therefore acts as a non-detail control. Improved texture readability together with unchanged or reduced gain readability indicates detail-specific enhancement, whereas a uniform increase across both families would indicate a generic gain that cannot be attributed to detail.

\paragraph{Perturbation families.}
For the texture family, a single-channel image $\mathbf{x}$ is decomposed into a low-pass component $\mathbf{L}=\mathrm{GaussianBlur}(\mathbf{x},\sigma{=}3.0)$ and its high-frequency residual. The residual is scaled by a retention factor $r$, and the result is mean-aligned to the original image so that brightness changes cannot masquerade as texture changes:
\begin{equation}
\mathbf{x}' = \mathbf{L} + r\cdot(\mathbf{x}-\mathbf{L}),
\qquad
\mathbf{x}' \leftarrow \mathbf{x}' + \left(\overline{\mathbf{x}}-\overline{\mathbf{x}'}\right).
\end{equation}
We use six retention levels from $200\%$ (high-frequency enhancement) to $0\%$ (low-pass only), with $100\%$ being the unmodified reference. For the gain family, $\mathbf{x}'=\mathrm{clip}(g\cdot\mathbf{x}+b\cdot R,\,0,\,255)$, where $R$ is the dynamic range of $\mathbf{x}$; we use five multiplicative levels $g\in\{0.8,0.9,1.0,1.1,1.2\}$ and two additive levels $b=\pm5\%$ of the dynamic range. Neither family moves edges, changes object shapes, or alters scene geometry.

\paragraph{Probe.}
Each perturbed image is encoded by the frozen encoder under test, giving an L2-normalized 768-dimensional CLIP embedding that is kept fixed during probe fitting. The probe is a ridge regressor from the embedding to the perturbation target: the degradation level for the texture family, and the signed brightness change ($g-1$ or $b$) for the gain family, whose seven levels do not form a monotone sequence. We use an outer 5-fold GroupKFold grouped by anchor image, so that all perturbation levels of one image fall in the same fold and every test fold contains only unseen images; within each outer training fold, the regularization strength is selected by an inner 3-fold GroupKFold over 13 candidates in $[10^{-3},10^{3}]$. We report the coefficient of determination $R^2$ on the held-out folds, where $R^2=1$ denotes perfect linear readout and $R^2=0$ denotes predicting the target mean.

\paragraph{Data and models.}
We compare the unadapted OpenAI CLIP encoder with the diffusion-branch variant trained with $\mathcal{L}_{\text{diff}}$ only, on 48 infrared anchors: 8 from LLVIP, 20 from M3FD, and 20 from JU-VNT-T. No text encoder, class prompt, or category-matching score is used in this analysis.

\paragraph{Results.}
Table~\ref{tab:diffusion-probe} reports the probe $R^2$ for both families. Texture readability improves under the diffusion branch ($0.654 \rightarrow 0.677$), while the readability of the signed global brightness change drops substantially ($0.263 \rightarrow 0.067$). This pattern matches the detail-specific case of the reading framework above: rather than producing a uniform sensitivity gain, the diffusion branch increases the linear accessibility of high-frequency texture detail and simultaneously suppresses content-irrelevant global brightness variation in the representation.

\begin{center}
\begin{minipage}{\columnwidth}
    \centering
    \tablefontsize
    \begin{tabular}{lccc}
    \toprule
    Probe target ($R^2$) & OpenAI CLIP & diffusion branch & $\Delta$ \\
    \midrule
    Texture level & 0.654 & 0.677 & $+0.023$ \\
    Gain (signed brightness) & 0.263 & 0.067 & $-0.196$ \\
    \bottomrule
    \end{tabular}
    \captionof{table}{Linear-probe $R^2$ for reading the perturbation target from frozen CLIP embeddings. Higher texture $R^2$ with lower gain $R^2$ indicates detail-specific enhancement rather than a generic sensitivity gain.}
    \label{tab:diffusion-probe}
\end{minipage}
\end{center}

\FloatBarrier

\printcredits

\section*{Declaration of generative AI and AI-assisted technologies in the manuscript preparation process}
During the preparation of this work, the authors used KIMI Code in order to assist with language editing, LaTeX preparation, and manuscript organization. After using this tool, the authors reviewed and edited the content as needed and take full responsibility for the content of the published article. Kimi-K2.5 was additionally used to draft the generic modality descriptions and the evaluation class prompts; all such materials were reviewed and finalized by the authors. 

\section*{Declaration of competing interest}
The authors declare that they have no known competing financial interests or personal relationships that could have appeared to influence the work reported in this paper.

\section*{Data availability}
The evaluation prompt files are publicly available at \url{https://github.com/wanginsten/TMT}. The training and evaluation code will be released in the same repository.

\section*{Funding}
This work is supported by the National Natural Science Foundation of China (No. 42501493).

\bibliographystyle{cas-model2-names}
\bibliography{references}

\begin{thebibliography}{35}
\expandafter\ifx\csname natexlab\endcsname\relax\def\natexlab#1{#1}\fi
\providecommand{\url}[1]{\texttt{#1}}
\providecommand{\href}[2]{#2}
\providecommand{\path}[1]{#1}
\providecommand{\DOIprefix}{doi:}
\providecommand{\ArXivprefix}{arXiv:}
\providecommand{\URLprefix}{URL: }
\providecommand{\Pubmedprefix}{pmid:}
\providecommand{\doi}[1]{\href{http://dx.doi.org/#1}{\path{#1}}}
\providecommand{\Pubmed}[1]{\href{pmid:#1}{\path{#1}}}
\providecommand{\bibinfo}[2]{#2}
\ifx\xfnm\relax \def\xfnm[#1]{\unskip,\space#1}\fi
\bibitem[{Cao et~al.(2025)Cao, Zhang and Zhang}]{cao2025irgpt}
\bibinfo{author}{Cao, Z.}, \bibinfo{author}{Zhang, J.}, \bibinfo{author}{Zhang,
  R.}, \bibinfo{year}{2025}.
\newblock \bibinfo{title}{{IRGPT}: Understanding real-world infrared image with
  bi-cross-modal curriculum on large-scale benchmark}, in:
  \bibinfo{booktitle}{Proceedings of the IEEE/CVF International Conference on
  Computer Vision}, pp. \bibinfo{pages}{166--176}.
\bibitem[{Chen et~al.(2025)Chen, Qian, Yang, Wang and Yang}]{chen2024amfd}
\bibinfo{author}{Chen, Z.}, \bibinfo{author}{Qian, Y.}, \bibinfo{author}{Yang,
  X.}, \bibinfo{author}{Wang, C.}, \bibinfo{author}{Yang, M.},
  \bibinfo{year}{2025}.
\newblock \bibinfo{title}{{AMFD}: Distillation via adaptive multimodal fusion
  for multispectral pedestrian detection}.
\newblock \bibinfo{journal}{IEEE Transactions on Multimedia}
  \bibinfo{volume}{27}, \bibinfo{pages}{8298--8310}.
\newblock \DOIprefix\doi{10.1109/TMM.2025.3604937}.
\bibitem[{{FLIR Systems}(2018)}]{flirv1}
\bibinfo{author}{{FLIR Systems}}, \bibinfo{year}{2018}.
\newblock \bibinfo{title}{{FLIR} releases starter thermal imaging dataset for
  machine learning advanced driver assistance development}.
\newblock
  \bibinfo{howpublished}{\url{https://www.flir.com/news-center/press-releases/flir-releases-starter-thermal-imaging-dataset-for-machine-learning-advanced-driver-assistance-development/}}.
\bibitem[{Gao et~al.(2024)Gao, Geng, Zhang, Ma, Fang, Zhang, Li and
  Qiao}]{gao2023clipadapter}
\bibinfo{author}{Gao, P.}, \bibinfo{author}{Geng, S.}, \bibinfo{author}{Zhang,
  R.}, \bibinfo{author}{Ma, T.}, \bibinfo{author}{Fang, R.},
  \bibinfo{author}{Zhang, Y.}, \bibinfo{author}{Li, H.}, \bibinfo{author}{Qiao,
  Y.}, \bibinfo{year}{2024}.
\newblock \bibinfo{title}{{CLIP-Adapter}: Better vision-language models with
  feature adapters}.
\newblock \bibinfo{journal}{International Journal of Computer Vision}
  \bibinfo{volume}{132}, \bibinfo{pages}{581--595}.
\newblock \DOIprefix\doi{10.1007/s11263-023-01891-x}.
\bibitem[{Ghosh et~al.(2021)Ghosh, Das, Sarkar and Nasipuri}]{ghosh2021juvnt}
\bibinfo{author}{Ghosh, S.}, \bibinfo{author}{Das, N.},
  \bibinfo{author}{Sarkar, P.}, \bibinfo{author}{Nasipuri, M.},
  \bibinfo{year}{2021}.
\newblock \bibinfo{title}{{JU-VNT}: A multi-spectral dataset of indoor object
  recognition using visible, near-infrared and thermal spectrum}.
\newblock \bibinfo{journal}{Multimedia Tools and Applications}
  \bibinfo{volume}{80}, \bibinfo{pages}{17807--17826}.
\newblock \DOIprefix\doi{10.1007/s11042-020-10302-z}.
\bibitem[{Girdhar et~al.(2023)Girdhar, El-Nouby, Liu, Singh, Alwala, Joulin and
  Misra}]{girdhar2023imagebind}
\bibinfo{author}{Girdhar, R.}, \bibinfo{author}{El-Nouby, A.},
  \bibinfo{author}{Liu, Z.}, \bibinfo{author}{Singh, M.},
  \bibinfo{author}{Alwala, K.V.}, \bibinfo{author}{Joulin, A.},
  \bibinfo{author}{Misra, I.}, \bibinfo{year}{2023}.
\newblock \bibinfo{title}{Imagebind: One embedding space to bind them all}, in:
  \bibinfo{booktitle}{Proceedings of the IEEE/CVF Conference on Computer Vision
  and Pattern Recognition}, pp. \bibinfo{pages}{15180--15190}.
\bibitem[{Ha et~al.(2017)Ha, Watanabe, Karasawa, Ushiku and
  Harada}]{ha2017mfnet}
\bibinfo{author}{Ha, Q.}, \bibinfo{author}{Watanabe, K.},
  \bibinfo{author}{Karasawa, T.}, \bibinfo{author}{Ushiku, Y.},
  \bibinfo{author}{Harada, T.}, \bibinfo{year}{2017}.
\newblock \bibinfo{title}{{MFNet}: Towards real-time semantic segmentation for
  autonomous vehicles with multi-spectral scenes}, in: \bibinfo{booktitle}{2017
  IEEE/RSJ International Conference on Intelligent Robots and Systems}, pp.
  \bibinfo{pages}{5108--5115}.
\newblock \DOIprefix\doi{10.1109/IROS.2017.8206396}.
\bibitem[{Hajimiri et~al.(2025)Hajimiri, Ben~Ayed and
  Dolz}]{hajimiri2025naclip}
\bibinfo{author}{Hajimiri, S.}, \bibinfo{author}{Ben~Ayed, I.},
  \bibinfo{author}{Dolz, J.}, \bibinfo{year}{2025}.
\newblock \bibinfo{title}{Pay attention to your neighbours: Training-free
  open-vocabulary semantic segmentation}, in: \bibinfo{booktitle}{Proceedings
  of the IEEE/CVF Winter Conference on Applications of Computer Vision}, pp.
  \bibinfo{pages}{5061--5071}.
\bibitem[{Ho et~al.(2020)Ho, Jain and Abbeel}]{ho2020ddpm}
\bibinfo{author}{Ho, J.}, \bibinfo{author}{Jain, A.}, \bibinfo{author}{Abbeel,
  P.}, \bibinfo{year}{2020}.
\newblock \bibinfo{title}{Denoising diffusion probabilistic models}, in:
  \bibinfo{booktitle}{Advances in Neural Information Processing Systems}, pp.
  \bibinfo{pages}{6840--6851}.
\bibitem[{Ji et~al.(2023)Ji, Li, Bian, Zhang and Cheng}]{ji2023semanticrt}
\bibinfo{author}{Ji, W.}, \bibinfo{author}{Li, J.}, \bibinfo{author}{Bian, C.},
  \bibinfo{author}{Zhang, Z.}, \bibinfo{author}{Cheng, L.},
  \bibinfo{year}{2023}.
\newblock \bibinfo{title}{{SemanticRT}: A large-scale dataset and method for
  robust semantic segmentation in multispectral images}, in:
  \bibinfo{booktitle}{Proceedings of the ACM International Conference on
  Multimedia}, pp. \bibinfo{pages}{3307--3316}.
\bibitem[{Jia et~al.(2021a)Jia, Yang, Xia, Chen, Parekh, Pham, Le, Sung, Li and
  Duerig}]{jia2021align}
\bibinfo{author}{Jia, C.}, \bibinfo{author}{Yang, Y.}, \bibinfo{author}{Xia,
  Y.}, \bibinfo{author}{Chen, Y.T.}, \bibinfo{author}{Parekh, Z.},
  \bibinfo{author}{Pham, H.}, \bibinfo{author}{Le, Q.}, \bibinfo{author}{Sung,
  Y.H.}, \bibinfo{author}{Li, Z.}, \bibinfo{author}{Duerig, T.},
  \bibinfo{year}{2021}a.
\newblock \bibinfo{title}{Scaling up visual and vision-language representation
  learning with noisy text supervision}, in: \bibinfo{booktitle}{International
  Conference on Machine Learning}, pp. \bibinfo{pages}{4904--4916}.
\bibitem[{Jia et~al.(2021b)Jia, Zhu, Li, Tang and Zhou}]{jia2021llvip}
\bibinfo{author}{Jia, X.}, \bibinfo{author}{Zhu, C.}, \bibinfo{author}{Li, M.},
  \bibinfo{author}{Tang, W.}, \bibinfo{author}{Zhou, W.},
  \bibinfo{year}{2021}b.
\newblock \bibinfo{title}{{LLVIP}: A visible-infrared paired dataset for
  low-light vision}, in: \bibinfo{booktitle}{Proceedings of the IEEE/CVF
  International Conference on Computer Vision Workshops}, pp.
  \bibinfo{pages}{3496--3504}.
\newblock \DOIprefix\doi{10.1109/ICCVW54120.2021.00389}.
\bibitem[{{Kimi Team}(2026)}]{kimi2026k25}
\bibinfo{author}{{Kimi Team}}, \bibinfo{year}{2026}.
\newblock \bibinfo{title}{Kimi k2.5: Visual agentic intelligence}.
\newblock \bibinfo{journal}{arXiv preprint arXiv:2602.02276} .
\bibitem[{Lee et~al.(2024)Lee, Anderson, Ranganathan, Zuo, Do, Gkioxari and
  Chung}]{lee2024cart}
\bibinfo{author}{Lee, C.}, \bibinfo{author}{Anderson, M.},
  \bibinfo{author}{Ranganathan, N.}, \bibinfo{author}{Zuo, X.},
  \bibinfo{author}{Do, K.}, \bibinfo{author}{Gkioxari, G.},
  \bibinfo{author}{Chung, S.J.}, \bibinfo{year}{2024}.
\newblock \bibinfo{title}{{CART}: Caltech aerial {RGB}-thermal dataset in the
  wild}, in: \bibinfo{booktitle}{Proceedings of the European Conference on
  Computer Vision}, pp. \bibinfo{pages}{236--256}.
\newblock \DOIprefix\doi{10.1007/978-3-031-73036-8_14}.
\bibitem[{Li et~al.(2025)Li, Zhao, Chang, Hou, Shan and Chen}]{li2025un2clip}
\bibinfo{author}{Li, Y.}, \bibinfo{author}{Zhao, J.}, \bibinfo{author}{Chang,
  H.}, \bibinfo{author}{Hou, R.}, \bibinfo{author}{Shan, S.},
  \bibinfo{author}{Chen, X.}, \bibinfo{year}{2025}.
\newblock \bibinfo{title}{un$^2$clip: Improving clip's visual detail capturing
  ability via inverting unclip}.
\newblock \bibinfo{journal}{arXiv preprint arXiv:2505.24517} .
\bibitem[{Liu et~al.(2022)Liu, Fan, Huang, Wu, Liu, Zhong and Luo}]{m3fd}
\bibinfo{author}{Liu, J.}, \bibinfo{author}{Fan, X.}, \bibinfo{author}{Huang,
  Z.}, \bibinfo{author}{Wu, G.}, \bibinfo{author}{Liu, R.},
  \bibinfo{author}{Zhong, W.}, \bibinfo{author}{Luo, Z.}, \bibinfo{year}{2022}.
\newblock \bibinfo{title}{Target-aware dual adversarial learning and a
  multi-scenario multi-modality benchmark to fuse infrared and visible for
  object detection}, in: \bibinfo{booktitle}{Proceedings of the IEEE/CVF
  Conference on Computer Vision and Pattern Recognition}, pp.
  \bibinfo{pages}{5802--5811}.
\bibitem[{Mao et~al.(2026)Mao, Mei, Lu, Liu, Chen, Zhao and Hu}]{mao2026pid}
\bibinfo{author}{Mao, F.}, \bibinfo{author}{Mei, J.}, \bibinfo{author}{Lu, S.},
  \bibinfo{author}{Liu, F.}, \bibinfo{author}{Chen, L.}, \bibinfo{author}{Zhao,
  F.}, \bibinfo{author}{Hu, Y.}, \bibinfo{year}{2026}.
\newblock \bibinfo{title}{{PID}: Physics-informed diffusion model for infrared
  image generation}.
\newblock \bibinfo{journal}{Pattern Recognition} \bibinfo{volume}{169},
  \bibinfo{pages}{111816}.
\newblock \DOIprefix\doi{10.1016/j.patcog.2025.111816}.
\bibitem[{Mikolov et~al.(2013a)Mikolov, Chen, Corrado and
  Dean}]{mikolov2013efficient}
\bibinfo{author}{Mikolov, T.}, \bibinfo{author}{Chen, K.},
  \bibinfo{author}{Corrado, G.}, \bibinfo{author}{Dean, J.},
  \bibinfo{year}{2013}a.
\newblock \bibinfo{title}{Efficient estimation of word representations in
  vector space}.
\newblock \bibinfo{journal}{arXiv preprint arXiv:1301.3781} .
\bibitem[{Mikolov et~al.(2013b)Mikolov, Sutskever, Chen, Corrado and
  Dean}]{mikolov2013distributed}
\bibinfo{author}{Mikolov, T.}, \bibinfo{author}{Sutskever, I.},
  \bibinfo{author}{Chen, K.}, \bibinfo{author}{Corrado, G.S.},
  \bibinfo{author}{Dean, J.}, \bibinfo{year}{2013}b.
\newblock \bibinfo{title}{Distributed representations of words and phrases and
  their compositionality}, in: \bibinfo{booktitle}{Advances in Neural
  Information Processing Systems}.
\bibitem[{Qazi et~al.(2026)Qazi, Maheshwari, Mukherjee and
  Lall}]{qazi2026tclip}
\bibinfo{author}{Qazi, T.}, \bibinfo{author}{Maheshwari, A.},
  \bibinfo{author}{Mukherjee, P.}, \bibinfo{author}{Lall, B.},
  \bibinfo{year}{2026}.
\newblock \bibinfo{title}{{T-CLIP}: Enabling thermal perception for contrastive
  language-image pretraining}.
\newblock \bibinfo{journal}{arXiv preprint arXiv:2606.00673} .
\bibitem[{Radford et~al.(2021)Radford, Kim, Hallacy, Ramesh, Goh, Agarwal,
  Sastry, Askell, Mishkin, Clark et~al.}]{radford2021clip}
\bibinfo{author}{Radford, A.}, \bibinfo{author}{Kim, J.W.},
  \bibinfo{author}{Hallacy, C.}, \bibinfo{author}{Ramesh, A.},
  \bibinfo{author}{Goh, G.}, \bibinfo{author}{Agarwal, S.},
  \bibinfo{author}{Sastry, G.}, \bibinfo{author}{Askell, A.},
  \bibinfo{author}{Mishkin, P.}, \bibinfo{author}{Clark, J.}, et~al.,
  \bibinfo{year}{2021}.
\newblock \bibinfo{title}{Learning transferable visual models from natural
  language supervision}, in: \bibinfo{booktitle}{International Conference on
  Machine Learning}, pp. \bibinfo{pages}{8748--8763}.
\bibitem[{Ramesh et~al.(2022)Ramesh, Dhariwal, Nichol, Chu and
  Chen}]{ramesh2022unclip}
\bibinfo{author}{Ramesh, A.}, \bibinfo{author}{Dhariwal, P.},
  \bibinfo{author}{Nichol, A.}, \bibinfo{author}{Chu, C.},
  \bibinfo{author}{Chen, M.}, \bibinfo{year}{2022}.
\newblock \bibinfo{title}{Hierarchical text-conditional image generation with
  clip latents}.
\newblock \bibinfo{journal}{arXiv preprint arXiv:2204.06125} .
\bibitem[{Ran et~al.(2025)Ran, Wang, Wang, Wang and Zhang}]{ran2025diffv2ir}
\bibinfo{author}{Ran, L.}, \bibinfo{author}{Wang, L.}, \bibinfo{author}{Wang,
  G.}, \bibinfo{author}{Wang, P.}, \bibinfo{author}{Zhang, Y.},
  \bibinfo{year}{2025}.
\newblock \bibinfo{title}{Diffv2ir: Visible-to-infrared diffusion model via
  vision-language understanding}.
\newblock \bibinfo{journal}{arXiv preprint arXiv:2503.19012} .
\bibitem[{Ranasinghe et~al.(2026)Ranasinghe, Schenck, Yellin, Hu, Funk and
  Patel}]{ranasinghe2026thermaldet}
\bibinfo{author}{Ranasinghe, Y.}, \bibinfo{author}{Schenck, E.},
  \bibinfo{author}{Yellin, F.}, \bibinfo{author}{Hu, S.},
  \bibinfo{author}{Funk, C.}, \bibinfo{author}{Patel, V.M.},
  \bibinfo{year}{2026}.
\newblock \bibinfo{title}{{Thermal-Det}: Language-guided cross-modal
  distillation for open-vocabulary thermal object detection}, in:
  \bibinfo{booktitle}{Proceedings of the IEEE/CVF Conference on Computer Vision
  and Pattern Recognition}, pp. \bibinfo{pages}{34628--34637}.
\newblock \URLprefix
  \url{https://openaccess.thecvf.com/content/CVPR2026/html/Ranasinghe_Thermal-Det_Language-Guided_Cross-Modal_Distillation_for_Open-Vocabulary_Thermal_Object_Detection_CVPR_2026_paper.html}.
\bibitem[{Rombach et~al.(2022)Rombach, Blattmann, Lorenz, Esser and
  Ommer}]{rombach2022ldm}
\bibinfo{author}{Rombach, R.}, \bibinfo{author}{Blattmann, A.},
  \bibinfo{author}{Lorenz, D.}, \bibinfo{author}{Esser, P.},
  \bibinfo{author}{Ommer, B.}, \bibinfo{year}{2022}.
\newblock \bibinfo{title}{High-resolution image synthesis with latent diffusion
  models}, in: \bibinfo{booktitle}{Proceedings of the IEEE/CVF Conference on
  Computer Vision and Pattern Recognition}, pp. \bibinfo{pages}{10684--10695}.
\bibitem[{Shivakumar et~al.(2020)Shivakumar, Rodrigues, Zhou, Miller, Kumar and
  Taylor}]{shivakumar2020pst900}
\bibinfo{author}{Shivakumar, S.S.}, \bibinfo{author}{Rodrigues, N.},
  \bibinfo{author}{Zhou, A.}, \bibinfo{author}{Miller, I.D.},
  \bibinfo{author}{Kumar, V.}, \bibinfo{author}{Taylor, C.J.},
  \bibinfo{year}{2020}.
\newblock \bibinfo{title}{{PST900}: {RGB}-thermal calibration, dataset and
  segmentation network}, in: \bibinfo{booktitle}{2020 IEEE International
  Conference on Robotics and Automation}, pp. \bibinfo{pages}{9441--9447}.
\newblock \DOIprefix\doi{10.1109/ICRA40945.2020.9196831}.
\bibitem[{{Teledyne FLIR}(2022)}]{flirv2}
\bibinfo{author}{{Teledyne FLIR}}, \bibinfo{year}{2022}.
\newblock \bibinfo{title}{{Teledyne FLIR} releases free expanded starter
  thermal dataset for {ADAS} and autonomous vehicle testing}.
\newblock
  \bibinfo{howpublished}{\url{https://oem.flir.com/en-150/about/news/expanded-teledyne-flir-starter-thermal-dataset-for-adas-and-autonomous-vehicle-testing/}}.
\bibitem[{Vidit et~al.(2023)Vidit, Engilberge and Salzmann}]{vidit2023clipgap}
\bibinfo{author}{Vidit, V.}, \bibinfo{author}{Engilberge, M.},
  \bibinfo{author}{Salzmann, M.}, \bibinfo{year}{2023}.
\newblock \bibinfo{title}{{CLIP} the gap: A single domain generalization
  approach for object detection}, in: \bibinfo{booktitle}{Proceedings of the
  IEEE/CVF Conference on Computer Vision and Pattern Recognition}, pp.
  \bibinfo{pages}{3219--3229}.
\newblock \DOIprefix\doi{10.1109/CVPR52729.2023.00314}.
\bibitem[{Wang et~al.(2025)Wang, Sun, Zhang, Tang, Liu and
  Wang}]{wang2024diffusion}
\bibinfo{author}{Wang, W.}, \bibinfo{author}{Sun, Q.}, \bibinfo{author}{Zhang,
  F.}, \bibinfo{author}{Tang, Y.}, \bibinfo{author}{Liu, J.},
  \bibinfo{author}{Wang, X.}, \bibinfo{year}{2025}.
\newblock \bibinfo{title}{Diffusion feedback helps {CLIP} see better}, in:
  \bibinfo{booktitle}{International Conference on Learning Representations}.
\bibitem[{Wei et~al.(2024)Wei, Liu, Qiao, Zhang, Yuille and
  Yu}]{wei2024dediffusion}
\bibinfo{author}{Wei, C.}, \bibinfo{author}{Liu, C.}, \bibinfo{author}{Qiao,
  S.}, \bibinfo{author}{Zhang, Z.}, \bibinfo{author}{Yuille, A.},
  \bibinfo{author}{Yu, J.}, \bibinfo{year}{2024}.
\newblock \bibinfo{title}{De-diffusion makes text a strong cross-modal
  interface}, in: \bibinfo{booktitle}{Proceedings of the IEEE/CVF Conference on
  Computer Vision and Pattern Recognition}, pp. \bibinfo{pages}{13492--13503}.
\bibitem[{Xie et~al.(2026)Xie, Zhang, Hou, Wang and Zhang}]{xie2026madf}
\bibinfo{author}{Xie, Y.}, \bibinfo{author}{Zhang, Z.}, \bibinfo{author}{Hou,
  Y.}, \bibinfo{author}{Wang, P.}, \bibinfo{author}{Zhang, H.},
  \bibinfo{year}{2026}.
\newblock \bibinfo{title}{Modality-aware dynamic fusion for weakly aligned
  {RGB-T} tiny object detection}.
\newblock \bibinfo{journal}{IEEE Signal Processing Letters}
  \bibinfo{volume}{33}, \bibinfo{pages}{1023--1027}.
\newblock \DOIprefix\doi{10.1109/LSP.2026.3659814}.
\bibitem[{Zhang et~al.(2025)Zhang, Li, Chou, Chen, Yuille, Wei and
  Xiao}]{zhang2025vlv}
\bibinfo{author}{Zhang, T.}, \bibinfo{author}{Li, Y.}, \bibinfo{author}{Chou,
  Y.C.}, \bibinfo{author}{Chen, J.}, \bibinfo{author}{Yuille, A.},
  \bibinfo{author}{Wei, C.}, \bibinfo{author}{Xiao, J.}, \bibinfo{year}{2025}.
\newblock \bibinfo{title}{Vision-language-vision auto-encoder: Scalable
  knowledge distillation from diffusion models}.
\newblock \bibinfo{journal}{arXiv preprint arXiv:2507.07104} .
\bibitem[{Zhou et~al.(2022a)Zhou, Yang, Loy and Liu}]{zhou2022cocoop}
\bibinfo{author}{Zhou, K.}, \bibinfo{author}{Yang, J.}, \bibinfo{author}{Loy,
  C.C.}, \bibinfo{author}{Liu, Z.}, \bibinfo{year}{2022}a.
\newblock \bibinfo{title}{Conditional prompt learning for vision-language
  models}, in: \bibinfo{booktitle}{Proceedings of the IEEE/CVF Conference on
  Computer Vision and Pattern Recognition}, pp. \bibinfo{pages}{16816--16825}.
\bibitem[{Zhou et~al.(2022b)Zhou, Yang, Loy and Liu}]{zhou2022coop}
\bibinfo{author}{Zhou, K.}, \bibinfo{author}{Yang, J.}, \bibinfo{author}{Loy,
  C.C.}, \bibinfo{author}{Liu, Z.}, \bibinfo{year}{2022}b.
\newblock \bibinfo{title}{Learning to prompt for vision-language models}.
\newblock \bibinfo{journal}{International Journal of Computer Vision}
  \bibinfo{volume}{130}, \bibinfo{pages}{2337--2348}.
\newblock \DOIprefix\doi{10.1007/s11263-022-01653-1}.
\bibitem[{Zhu et~al.(2024)Zhu, Lin, Ning et~al.}]{zhu2023languagebind}
\bibinfo{author}{Zhu, B.}, \bibinfo{author}{Lin, B.}, \bibinfo{author}{Ning,
  M.}, et~al., \bibinfo{year}{2024}.
\newblock \bibinfo{title}{Languagebind: Extending video-language pretraining to
  n-modality by language-based semantic alignment}, in:
  \bibinfo{booktitle}{International Conference on Learning Representations}.

\end{thebibliography}

\end{document}